\documentclass[10pt, a4paper, logo, copyright, numbering]{googledeepmind}

\usepackage[utf8]{inputenc}

\usepackage[authoryear, sort&compress, round]{natbib}
\usepackage{graphicx} 
\usepackage{amsmath, amsthm, amssymb}
\usepackage{tabularx}
\usepackage{booktabs}
\usepackage{bm}
\usepackage{multirow}
\usepackage{xcolor}
\usepackage{caption, subcaption}
\usepackage{xspace}
\usepackage{xurl}
\usepackage{hyperref}
\usepackage{algorithm,algorithmic}
\usepackage[export]{adjustbox}
\usepackage{placeins}
\allowdisplaybreaks

\newcommand{\cond}{\,|\,}
\newcommand{\acro}[1]{\textsc{#1}\xspace}
\newcommand{\coreset}{\mathcal{C}}
\newcommand{\sr}{s}
\newcommand{\mut}{\mu^{\theta}}
\newcommand{\Sigmat}{\Sigma^{\theta}}
\renewcommand{\eqref}[1]{Eq. (\ref{#1})}
\newcommand{\secref}[1]{Section~\ref{#1}}

\newcommand{\algcomment}[1]{\hfill\makebox[3.1cm][l]{$\triangleright$ #1}}

\newcommand{\AIPW}{\acro{AIPW}}
\newcommand{\BayesAME}{\acro{BayesAME}}
\newcommand{\RBayesAME}{\acro{BayesAME-RS}}
\newcommand{\BayesAIPW}{\acro{Bayes-AIPW}}
\newcommand{\RSL}{\acro{Random-Sampling-Learn}}
\newcommand{\RS}{\acro{RS-Mean}}
\newcommand{\BayesRSL}{\acro{Bayes-RS-Learn}}
\newcommand{\SeqAPW}{\acro{Seq-APW}}

\newcommand{\ProEval}{\acro{ProEval}}
\newcommand{\IRT}{\acro{IRT}}
\newcommand{\CAT}{\acro{Cost-Accuracy Trade-Off}}
\newcommand{\RMSELog}{\acro{RMSE Log}}
\newcommand{\RMSEGain}{\acro{RMSE Gain}} 
\newcommand{\EWIRMSE}{\acro{Evaluation-Weighted Integrated RMSE}}

\title{BayesAME: Bayesian Active Model Evaluation}

\renewcommand{\copyrightext}{\footerfont *Senior co-authors.\\ \textcopyright\, \the\year{} Google DeepMind. All rights reserved}

\begin{document}

\author[1]{Paula Cordero Encinar}
\author[2]{Taylan Cemgil}
\author[2]{Arnaud Doucet}
\author[2]{Virginia Aglietti\textsuperscript{*}}
\author[2]{Silvia Chiappa\textsuperscript{*}}

\affil[1]{Imperial College London (work conducted while at Google DeepMind)}
\affil[2]{Google DeepMind}

\begin{abstract}
Evaluating large generative models across benchmarks is time-consuming and computationally expensive. This drives the need for methods that can estimate full benchmark performance by evaluating models on only a subset of items, known as a coreset. Current literature mostly requires the practitioner to input a coreset size. However, when reliable performance estimation takes priority over efficiency, an evaluation method should also be capable of automatically determining a coreset size that reflects this priority. We introduce \BayesAME, a sequential Bayesian framework specifically targeting automatic determination of the coreset size. \BayesAME models performance as a random variable by defining a latent ability for each group of items sharing the same historical model performances, with a joint prior distribution encoding the belief that the target model behaves similarly to these historical models. The posterior distribution over these abilities is used to derive performance estimators, quantify performance uncertainty, and select items to add to the coreset via an information-gain criterion. The coreset is iteratively augmented until the performance estimate fluctuation and the performance uncertainty fall below their respective user-defined thresholds. Moreover, we propose a multi-target extension that actively captures performance correlations across multiple target models to further reduce the  coreset size. Through extensive experiments across diverse benchmarks, we demonstrate that \BayesAME consistently outperforms sequential adaptations of existing methods. Crucially, our comprehensive analysis addresses recent skepticism in the literature, establishing that non-random coreset selection is advantageous over random selection, and demonstrating that one can reliably improve upon the random sample mean baseline. Additionally, we highlight that leveraging continuous response log-likelihoods over traditional binary scores significantly enhances estimation accuracy.
\end{abstract}

\maketitle

\section{Introduction}
The massive scale and frequency of generating responses render large generative model evaluation across benchmarks a time-consuming and computationally expensive process \citep{liang2023holistic}. To mitigate this, a growing number of methods estimate full benchmark performance by evaluating models on only a subset of items, known as a \textit{coreset}. Such methods mostly require the practitioner to input a coreset size \citep{berrada2025scaling,bowyer2026efficient,fisch2026collaborative,huang2026proeval,kipnis2025metabench,kossen2021active,hsu2026efficient,liao2025toward,liu2026active,perlitz2024efficient,saranathan2025sublime,vivek2024anchor,zhang2025how}. While this is most suitable when facing strict computational limits, in scenarios where reliable performance estimation takes priority over efficiency, an evaluation method should also be capable of automatically determining a coreset size that reflects this priority. In addition, the predominant approach of leveraging evaluations from historical models (\textit{reference models}) has largely operated within the regime in which the models being evaluated (\textit{target models}) fall within the performance range of these reference models (\textit{interpolation regime}). \cite{zhang2025how} underscored the importance of also testing efficient model evaluation methods in regimes where the target models significantly deviate from this range (\textit{extrapolation regimes}) to obtain a more accurate assessment of a method's robustness. In particular, through a study that considered both regimes, the authors questioned the utility of non-random coreset selection and highlighted the difficulty of outperforming the mean of a random coreset as a performance estimate in extrapolation.

In this work, we focus on efficient model evaluation leveraging reference models, specifically targeting automatic determination of the coreset size. We propose \BayesAME, a sequential Bayesian method that models performance as a random variable via underlying abilities whose joint prior distribution encodes the belief that the target model behaves similarly to the reference models. The posterior distribution over these abilities is used to derive performance estimators, quantify performance uncertainty, and select items to add to the coreset via an information-gain criterion. The coreset is iteratively augmented until the performance estimate fluctuation and the performance uncertainty fall below their respective user-defined thresholds. In the setting where multiple target models are being evaluated, we capture correlation among them using a linear model of coregionalization \citep{alvarez2012kernels,journel1978mining} with the goal of further reducing the coreset size. To validate our approach, we comprehensively evaluate \BayesAME against sequential adaptations of existing methods. 

Our main contributions are summarized as follows:\vspace{-15pt}
\begin{itemize}
    \item We formulate efficient model evaluation with automatic coreset-size determination as a Bayesian sequential process that augments the coreset until the performance estimate fluctuation and the performance uncertainty fall below their respective user-defined thresholds.
    \item We develop a computationally efficient method, equipped with robust performance estimators possessing well-defined properties.
    \item We design active selection strategies that identify benchmark items maximizing information gain.
    \item We extend the method to capture correlations across multiple target models to further reduce the coreset size.
    \item We conduct an extensive empirical analysis across varying numbers of reference models, interpolation/extrapolation regimes, and different degrees of target model correlation, showing that \BayesAME outperforms sequential adaptations of existing methods designed to automatically determine the coreset size. We further show that \BayesAME remains advantageous even when the coreset size is predetermined. 
    \item Crucially, this analysis overturns recent claims in the literature by demonstrating that non-random coreset selection does improve performance and that several methods can reliably outperform the random sample mean baseline.
    \item We demonstrate that leveraging available response log-likelihoods, rather than using binary scores, greatly improves performance estimation.
\end{itemize}

\section{Problem Setup} Consider a benchmark $E=\{(x_i, y_i)\}_{i=1}^N$, where $x_i$ is the $i$-th input instance and $y_i$ is the corresponding ground-truth response. Let $\mathcal{T}=\{t_1, \dots, t_{K_t}\}$ be a set of $K_t$ \textit{target models} that we wish to evaluate, and let $s^t_i$ be the \textit{score} of target model $t\in \mathcal{T}$ on item $(x_i,y_i)$. We consider the problem of estimating the full benchmark performance $R^\star_t=\frac{1}{N}\sum_{i=1}^N s^t_i$ by evaluating the target model $t$ only on a subset of items $\coreset^t\subseteq E$, which we refer to as a \textit{coreset}. We focus on the setting where the full benchmark scores for a set of $K_r$ \textit{reference models} $\{r_1, \dots, r_{K_r}\}$ are available.   
Although our framework is applicable to scores $s^t_i\in\mathbb{R}$, in our experiments we consider either binary scores ($s^t_i\in\{0,1\}$), indicating whether the model's response matches $y_i$, or bounded continuous scores, where, e.g., $s^t_i\in[0,1]$ represents the probability assigned by the model to the ground-truth response $y_i$. 

Our goal is to address the setting in which the user prioritizes reliable performance estimation over minimizing cost, without \textit{a priori} knowing the coreset size required to achieve this. Therefore, we focus on methods that not only operate with a user-specified coreset size, but can also automatically determine a coreset size that reflects this priority. 

In the remainder of the paper, with a slight abuse of notation, we use $E$ and $\coreset^t$ to indicate the indices of the benchmark items rather than the items themselves.

\section{\BayesAME} 
\BayesAME is a sequential Bayesian method that assigns a latent \textit{ability} to each group of benchmark items sharing the same reference model performances. It then treats the item scores as noisy versions of these abilities and models the full benchmark performance as a random variable defined as the average of the scores, $R_t = \frac{1}{N} \sum_{i=1}^N \sr_i^t$ (where, with a slight abuse of notation, we use $\sr_i^t$ to denote both the score random variable and its realization). The abilities are assigned a joint Gaussian prior distribution to encode the belief that, if the reference models achieved similar scores on two groups of items, the target model likely exhibits similar behavior. At each iteration, the posterior distribution of the latent abilities is used to derive performance estimators, quantify performance uncertainty, and select an item to add to the coreset via an information gain criterion. The sequential process terminates when the performance fluctuation and uncertainty fall below their respective user-defined thresholds.

\subsection{Single-Target Setting} \label{sec:bayesAME_single_target}
We first consider the setting of a single target model. For the remainder of this section, we omit the target model index. 
Let $(s_i^{r_1}, \ldots, s_i^{r_{K_r}})$ be the vector of scores achieved by the $K_r$ reference models on item $(x_i, y_i)$. 
Suppose that the reference models yield $B$ unique score vectors across the $N$ benchmark items, which we denote as the set $\{\phi_1,\ldots,\phi_B\}$. This partitions the items into $B$ disjoint buckets, such that all items assigned to bucket $b$ share the same reference score vector $\phi_b$.    

We consider a random vector $\theta=(\theta_1,\ldots,\theta_B)$, where $\theta_b$ represents the target model's underlying ability on items belonging to bucket $b$, with a Gaussian prior distribution $p(\theta)=\mathcal{N}(\mut, \Sigmat)$, defined as:
\begin{align}\label{eq:gaussian_prior_from_reference}
\mut_b=\frac{1}{K_r}\phi_b^\top \mathbf{1}_{K_r},\qquad \Sigmat_{b,b'}= \alpha \exp\left(-\tfrac{\beta}{K_r}\lVert\phi_b - \phi_{b'}\rVert^2\right),
\end{align}
where $\mathbf{1}_{K_r}$ denotes a $K_r$-dimensional column vector of ones, $\alpha$ and $\beta$ are parameters to be learned, and $\lVert\phi_b - \phi_{b'}\rVert^2$ is the squared Euclidean distance. This form of the covariance encodes the belief that buckets with similar reference scores yield highly correlated latent abilities for the target model\footnote{When given metadata for the reference and target models, we could employ a weighted mean and covariance, where the weights capture the similarity between the target and reference models, see Appendix \ref{sec:app:alternative_weighted_prior}.}. Applying this latent variable approach at the bucket level, rather than the item level, reduces overreliance on prior information when reference models behave similarly and thus a large number of items map to the same $\phi_b$, which most commonly manifests when $K_r$ is small or when the scores are binary.

For any item $(x_i,y_i)$ belonging to bucket $b$, we model the target model's score $s_i$ as a random variable with conditional distribution $p(\sr_i \cond \theta_b) = \mathcal{N}(\theta_b, N_b\sigma^2)$, where $N_b$ is the number of items in bucket $b$. 
Let $H$ denote an $N \times B$ binary indicator matrix, where $H_{i,b}=1$ if the $i$-th item belongs to bucket $b$, and $0$ otherwise. The target model's score vector $S=(\sr_1,\ldots,\sr_N)$ is then modeled as a random variable with conditional distribution $p(S \cond \theta) = \mathcal{N}(H\theta, D)$, where $D$ is an $N\times N$ diagonal noise matrix such that $D_{i,i}=\sigma^2 \sum_{b=1}^B H_{i,b} N_b$. Note that, when all items map to unique reference vectors ($B=N$), $D_{i,i}=\sigma^2$. 
This bucket-size-dependent noise accounts for the inherent heterogeneity within larger buckets. Even though items in a large bucket share identical reference scores, the target model's performance on each individual item might vary. By scaling the individual item variance by $N_b$, we ensure that the variance of the average bucket score remains constant at $\sigma^2$, regardless of how many items the bucket contains:
$\operatorname{Var}\left( \frac{1}{N_b} \sum_{i \in \text{bucket } b} \sr_i \right) = \frac{1}{N_b^2} \sum_{i \in \text{bucket } b} \operatorname{Var}(\sr_i) = \frac{1}{N_b^2} (N_b \cdot N_b \sigma^2) = \sigma^2$. While a more principled approach to modeling binary and bounded continuous scores would explicitly model their support, exploring this direction revealed that the approximations required not only introduced substantial computational overhead but also resulted in inferior performance (see Appendix~\ref{sec:app:logit}). In contrast, the Gaussian model admits closed-form posterior distributions, delivering superior scalability and performance. We therefore adopt a Gaussian model as the default choice.

Let $s_\coreset$ denote the subvector of $s$ corresponding to the items in $\coreset$, and let $\mut_{|\coreset}$ and $\Sigmat_{|\coreset}$ be the posterior mean and covariance of $\theta$ conditioned on $s_\coreset$, such that $p(\theta\cond s_\coreset)=\mathcal{N}(\mut_{|\coreset}, \Sigmat_{|\coreset})$. 
The posterior mean and variance of the full benchmark performance $R = \frac{1}{N}\sum_{i=1}^N \sr_i$ are given by:
\begin{align}
    \mathbb{E}[R \cond s_\coreset] &= \frac{1}{N}\left(\sum_{i \in\coreset} s_i + \sum_{j \in E\setminus \coreset} \mathbb{E}[\sr_j \cond s_\coreset] \right) = \frac{1}{N}\left(\sum_{i \in\coreset} s_i + \sum_{b =1}^B (N_b-n_b)\mut_{b| \coreset} \right),\nonumber \\
    \operatorname{Var}(R\cond s_\coreset) &= \frac{1}{N^2} \sum_{j \in E\setminus \coreset}\sum_{j' \in E\setminus \coreset} \operatorname{Cov}(s_j, s_{j'}\cond s_\coreset)=\frac{1}{N^2} \sum_{b=1}^B (N_b - n_b) \left(\sigma^2 N_b + \sum_{b'=1}^B (N_{b'} - n_{b'}) \Sigmat_{b,b'|\coreset}\right ),\label{eq:variance_performance}
\end{align}
where $n_b$ is the number of evaluated items in bucket $b$.
As more items are evaluated, the posterior mean increasingly relies on the actual scores, converging to the true performance $R^\star$ when the entire benchmark has been evaluated ($n_b = N_b$ for all $b$), at which point the posterior variance vanishes.

\paragraph{Performance Estimator}
We construct an estimator $\hat{R}$ of $R^\star$ by adapting our approach to the number of unique reference score vectors in the benchmark, introducing a small margin $N_0 > 0$ to allow for slight deviations from uniqueness.

In the \textit{nearly-unique reference regime} $B \ge N - N_0$, the natural and optimal performance estimator is the Bayesian posterior expectation:
\begin{equation}\label{eq:posterior_mean_estimator}
  \hat{R}_{\text{Bayes}} = \mathbb{E}[R \cond s_\coreset]=\frac{1}{N}\left(\sum_{i \in\coreset} s_i + \sum_{b =1}^B (N_b-n_b)\mut_{b|\coreset} \right).
\end{equation}
However, using \eqref{eq:posterior_mean_estimator} in the \textit{non-unique reference regime} $B < N- N_0$ presents a potential issue: if a heavily populated bucket $b$ has a large number of unevaluated items ($n_b\ll N_b$), its posterior mean $\mut_{b|\coreset}$ exerts a high weight on the overall estimate, amplifying any estimation errors. In this regime, we instead use the following estimator:
\begin{equation}\label{eq:performance_estimate_control_variate}
    \hat{R}_{\text{CV}} = \frac{1}{n} \sum_{i \in \coreset} s_i +  \sum_{b=1}^B \left(\frac{N_b}{N}- \frac{n_b}{n}\right)\, \mathbb{E}[\theta_b \cond s_\coreset] = \frac{1}{n} \sum_{i \in \coreset} s_i + \sum_{b=1}^B\left(\frac{N_b}{N} - \frac{n_b}{n}\right)\mut_{b|\coreset},
\end{equation}
where $n$ indicates the size of the coreset $\coreset$. This estimator serves as an in-sample proxy for the unbiased control variate estimator given in Appendix \ref{sec:app:cv_estimator}. We can unify both regimes into a single estimator:
\begin{equation}\label{eq:performance_estimate}
    \hat{R} = w \sum_{i \in \coreset}s_i + \sum_{b=1}^B\left(\frac{N_b}{N} - n_b w\right)\mut_{b|\coreset},
\end{equation}
where $w=\frac{1}{n}$ if $B < N-N_0$ and $w=\frac{1}{N}$ if $B \ge N - N_0$.

\paragraph{Selection Strategy}
At each iteration of the sequential process, we use an information-gain criterion to select the next item on which to evaluate the target model. Specifically, we choose an item that maximally reduces the expected uncertainty of the model's performance $R$, measured via differential entropy:
$\alpha_{\text{IG}}(i)={\mathcal H}(R\cond s_\coreset)-\mathbb{E}_{s_i}[{\mathcal H}(R\cond s_\coreset, s_i)]$.
Assuming candidate item $(x_i,y_i)$ belongs to bucket $b$, due to the Gaussian nature of the posterior distribution, this expected information gain simplifies to:
\begin{equation} \label{eq:information_gain}
\alpha_{\text{IG}}(i)=\frac{1}{2}\log \left( \frac{\operatorname{Var}(R\cond s_\coreset)}{\operatorname{Var}(R \cond s_\coreset, s_i)} \right) = -\frac{1}{2} \log \left(1-\frac{\big(\sum_{b'=1}^B (N_{b'} - n_{b'}) \Sigmat_{b,b'|\coreset} + \sigma^2 N_b\big)^2}{N^2\operatorname{Var}(R\cond s_\coreset) (\Sigmat_{b,b|\coreset} + \sigma^2 N_b) }\right).
\end{equation}
Note that all items in the same bucket have the same expected information-gain value. In the non-unique reference regime with imbalanced bucket sizes, purely active selection can degrade performance. This occurs because the target model may exhibit fine-grained variations in behavior that are not captured by the coarse grouping of the reference models. Consequently, when $B < N - N_0$, we default to random selection, which ensures that the expected number of sampled items from each bucket remains proportional to its size. The margin $N_0 > 0$ allows the method to use active sampling whenever the reference score vectors are nearly unique. Furthermore, when dealing with continuous scores, unique reference score vectors may become non-unique when rounded to a specified tolerance. Under such conditions, we similarly default to random selection.

We also explored alternative selection strategies that operate on batches of items rather than individual items, including non-myopic approaches. However, these provided similar performance (see Appendix~\ref{sec:app:alternative_active_selection}).

\paragraph{Stopping Criterion} 
One of our key objectives is to determine the coreset size, rather than requiring it as input from the user. Intuitively, we would like to stop evaluating additional items once they are unlikely to meaningfully change our estimate of the target model's performance. 

To formalize this, we monitor the uncertainty of $R$ as items are evaluated. We quantify this uncertainty using the width of the 95\% credible interval of $R$:
$q_{97.5}(R\cond s_\coreset) - q_{2.5}(R\cond  s_\coreset)= 2 z_{0.975} \sqrt{\operatorname{Var}(R\cond  s_\coreset)}$,
where $z_{0.975}$ denotes the standard normal $z$-score and $\operatorname{Var}(R\cond s_\coreset)$ is given in \eqref{eq:variance_performance}. We dynamically augment the coreset until a two-condition stopping criterion is satisfied: $(i)$ the performance estimate $\hat R$ stabilizes, defined as its range over a window of $W$ iterations falling below a threshold $\epsilon_1$; and $(ii)$ $2 z_{0.975} \sqrt{\operatorname{Var}(R\cond s_\coreset)}\leq\epsilon_2$. The threshold $\epsilon_2$ represents the maximum acceptable width of this uncertainty interval, translating directly into a user's tolerated margin of error. By treating $\epsilon_1$ and $\epsilon_2$ as user-defined parameters, \BayesAME enables the user to decide the precision required. For additional discussion, see Appendix~\ref{app:stopping_criterion}.

\paragraph{Posterior Distribution}
To optimize the cost required to compute $\mut_{|\coreset}$ and $\Sigmat_{|\coreset}$, throughout the sequential process we adapt our formulation as detailed below. 
Let $H_{\coreset}$ and $D_{\coreset}$ denote the submatrices of $H$ and $D$ formed by the rows corresponding to $\coreset$. 
Let $\coreset_b$ denote the subset of $n_b$ items in the coreset $\coreset$ that are in bucket $b$. 
Finally, let $\bar{s}_{\coreset_{1:B}} = (\bar{s}_{\coreset_1}, \ldots, \bar{s}_{\coreset_B})$, where $\bar{s}_{\coreset_b} = \frac{1}{n_b}\sum_{i \in \coreset_b} s_i$ (for notational simplicity we assume $n_b > 0$ for all $b$). As detailed in Appendix \ref{sec:app:equivalent_bucket_matrix_H}, the posterior distribution of $\theta$ satisfies $p(\theta \cond s_\coreset)=p(\theta \cond \bar{s}_{\coreset_{1:B}})$. Although mathematically equivalent, computing $p(\theta \cond s_\coreset)$ requires inverting an $n \times n$ matrix, whereas computing $p(\theta \cond \bar{s}_{\coreset_{1:B}})$ requires inverting at most a $B\times B$ matrix.
Indeed, $p(\theta \cond s_\coreset)$ is Gaussian with mean $\mut_{|\coreset}$ and covariance $\Sigmat_{|\coreset}$ given by:
\begin{align}\label{eq:theta_posterior_item}
    \mut_{|\coreset} &= \mut + \Sigmat H_\coreset^\top (H_\coreset \Sigmat H_\coreset^\top + D_\coreset)^{-1} (s_{\coreset} - H_{\coreset}\mut),\nonumber\\
    \Sigmat_{|\coreset} &= \Sigmat - \Sigmat H_\coreset^\top (H_\coreset \Sigmat H_\coreset^\top + D_\coreset)^{-1} H_\coreset \Sigmat,
\end{align}
while, starting from $p(\theta \cond \bar{s}_{\coreset_{1:B}})$, the mean $\mut_{|\coreset}$ and covariance $\Sigmat_{|\coreset}$ can be expressed as:
\begin{align}\label{eq:theta_posterior_bucket}
    \mut_{|\coreset} &= \mut + \Sigmat (\Sigmat + \Delta)^{-1} (\bar{s}_{\coreset_{1:B}} - \mut), \nonumber\\
    \Sigmat_{|\coreset} &= \Sigmat - \Sigmat (\Sigmat + \Delta)^{-1} \Sigmat,
\end{align}
where $\Delta$ is a diagonal noise matrix with entries $\Delta_{b,b} = \sigma^2 N_b / n_b$.
To achieve the best computational cost, we use \eqref{eq:theta_posterior_item} when $n < B$ and \eqref{eq:theta_posterior_bucket} when $n \ge B$. We employ Cholesky updates across both regimes, leveraging the fact that the prior covariance $\Sigmat$ remains fixed between hyperparameter updates, which occur every $F$ iterations. This yields a computational complexity of $\mathcal{O}(\min(n, B)^2)$ per iteration.

\begin{algorithm}[t]
\caption{\BayesAME \acro{Single-target}}
\label{algorithm:BayesAME}
\begin{algorithmic}
\REQUIRE Benchmark $E$; noise variance $\sigma^2$; margin $N_0$; stopping criterion thresholds $\epsilon_1, \epsilon_2$ and window-length $W$; \acro{Optimizer},  hyperparameter initial values $\alpha_0, \beta_0$ and update frequency $F$
\STATE Initialize coreset $\coreset \leftarrow \emptyset$
\STATE Initialize hyperparameters $\alpha \leftarrow \alpha_0, \beta\leftarrow \beta_0$
\FOR{$j \in\{1, \dots, N\}$}
\IF{$B\ge N-N_0$}
\STATE Select item via expected information gain: $i^*=\arg\max_{i \in E\setminus \coreset}\alpha_{\text{IG}}(i)$ 
\algcomment{\eqref{eq:information_gain}}
\ELSE
\STATE Select item via random sampling: $i^* \sim \text{Uniform}(E\setminus \coreset)$
\ENDIF
\STATE Obtain score $s_{i^*}$ and add $i^*$ to the coreset: $\coreset\leftarrow\coreset \cup \{i^*\}$
\STATE Compute $\mut_{|\coreset}$ and $\Sigmat_{|\coreset}$ \algcomment{\eqref{eq:theta_posterior_item} or \eqref{eq:theta_posterior_bucket}}
\STATE Compute performance estimate $\hat R$   \algcomment{\eqref{eq:performance_estimate}} 
\STATE Set $\hat R_j\leftarrow \hat R$
\IF{$j\% F=0$}
\STATE Update hyperparameters:
$\alpha, \beta \leftarrow \acro{Optimizer}(\mathcal{L}(\alpha, \beta), \alpha, \beta)$  \algcomment{\eqref{eq:NLML_objective}} 
\ENDIF
\IF{$j\ge W$ \textbf{and} $\max_{l\in[j-W+1, j]}(\hat{R}_l)-\min_{l\in[j-W+1, j]}(\hat{R}_l)\le\epsilon_1$ \textbf{and} $2 z_{0.975} \sqrt{\operatorname{Var}(R\cond s_\coreset)}\le \epsilon_2$}
\STATE \textbf{break}
\ENDIF
\ENDFOR
\STATE Return $\hat R$
\end{algorithmic}
\end{algorithm}

\paragraph{Algorithm} We summarize the method in Algorithm~\ref{algorithm:BayesAME}.
Starting with an empty coreset $\coreset$, \BayesAME iteratively selects the benchmark item maximizing the expected information gain $\alpha_{\text{IG}}(i)$ when $B\geq N-N_0$ (or at random otherwise). The target model is evaluated on this item, its index is added to $\coreset$, and the posterior mean $\mut_{|\coreset}$ and covariance $\Sigmat_{|\coreset}$ are updated, along with the performance estimate $\hat{R}$. Every $F$ iterations, the covariance hyperparameters $\alpha$ and $\beta$ are optimized by minimizing the negative log-marginal likelihood $\mathcal{L}(\alpha, \beta)$ of the coreset scores (see Appendix \ref{sec:app:hyperparameter_optimization}). This sequential process terminates once the two-condition stopping criterion is satisfied: $(i)$ the range of the performance estimate $\hat{R}$ over a window of $W$ iterations falls below threshold $\epsilon_1$; and $(ii)$ the width of the 95\% credible interval falls below threshold $\epsilon_2$.

\subsection{Multi-Target Setting}  \label{sec:bayesAME_multiple_target}
In the case of multiple target models $\mathcal{T}=\{t_1, \dots, t_{K_t}\}$, a standard approach would be to simply apply the single-target method of Section \ref{sec:bayesAME_single_target} to each model independently. However, jointly modeling target models that exhibit correlated performance behaviors can improve estimation accuracy and reduce the required coreset size. We model the targets jointly using a linear model of coregionalization \citep{alvarez2012kernels,journel1978mining}.

We capture cross-model correlations by assuming that their abilities are driven by $L$ shared, independent latent random vectors $u_l \in \mathbb{R}^B$. Specifically, let $\theta^t_b$ represent the latent ability of target model $t$ on bucket $b$, and let $\theta^t = (\theta_1^t, \dots, \theta_B^t)$. We model $\theta^t$ as:
\[\theta^t =\mu^{\theta^t} + \sum_{l=1}^L w_l^t u_l,\] 
where $\mu^{\theta^t}$ is defined as in \eqref{eq:gaussian_prior_from_reference}, $w_1^t, \dots, w_L^t$ are model-specific weights to be learned, and $u_l \sim \mathcal{N}(0, \Sigma^{\theta,l})$, where $\Sigma^{\theta,l}$ is a covariance matrix specific to latent vector $l$. This induces a Gaussian prior distribution over $\theta = (\theta^{t_1}, \dots, \theta^{t_{K_t}})$ given by:
\begin{equation*}
    p(\theta)=\mathcal{N}\left(\mut, \sum_{l=1}^L w_l w_l^\top \otimes \Sigma^{\theta,l}\right),
\end{equation*} 
where $\otimes$ denotes the Kronecker product, $w_l = (w_l^{t_1}, \dots, w_l^{t_{K_t}})^\top$ and $\mut= (\mu^{\theta^{t_1}}, \dots, \mu^{\theta^{t_{K_t}}})$. 
This formulation provides a flexible yet computationally efficient way to capture correlation across both buckets and models by combining bucket-specific ($\Sigma^{\theta,l}_{b,b'}$) and model-specific ($w_l^t$) terms. To provide greater modeling flexibility, we use two covariance families with complementary inductive biases: the squared exponential covariance defined as in \eqref{eq:gaussian_prior_from_reference}, which imposes strict smoothness,  and a Mat\'ern-3/2 kernel of the form $\Sigma^{\theta,l}_{b,b'} = \alpha_l \left( 1 + \sqrt{3} \beta_l  \|\phi_b - \phi_{b'}\|  \right) \exp\left( - \sqrt{3} \beta_l \|\phi_b - \phi_{b'}\| \right)$ with $\alpha_l, \beta_l>0$, which imposes weaker smoothness. Because the degree of correlation between the target models is typically unknown \textit{a priori}, we set $L=K_t$. This enables the framework to dynamically adapt: it captures strong correlations by sharing latent components across models, yet it retains the ability to recover the fully independent case. 
The scores are then modeled similarly to the single-target setting described in Section \ref{sec:bayesAME_single_target}.

\paragraph{Selection Strategy}
In contrast to the single-target setting, the multi-target selection step requires simultaneously selecting an item and a target model to evaluate on that item, substantially expanding the search space. 
Furthermore, relying on a purely greedy information-gain criterion across this joint item-model space can bias performance estimates, particularly during early stages when correlations across target models are still being learned. 
To mitigate this bias and establish a robust initial estimate of the correlation structure, we depart from the empty coreset initialization used in the single-target setting. Instead, we randomly select 10\% of the benchmark items and evaluate all target models on them. After this initialization step, we deploy the following selection strategy, which maintains some degree of exploration.

We iterate deterministically through the target models. For a given model $t$, we select the item that maximizes the expected information gain:
$ \alpha^t_{\text{IG}}(i) = {\mathcal H}(R^t \cond s_\coreset) - \mathbb{E}_{s_i^t}[{\mathcal H}(R^t \cond s_\coreset, s^t_i)]$,
where $s_\coreset=\{s^{t_1}_{\coreset^{t_1}},\ldots,s^{t_{K_t}}_{\coreset^{t_{K_t}}}\}$. This approach substantially reduces the search space, maintaining a selection complexity identical to that of the single-target setting. We iteratively augment the coresets $\{\coreset^{t_1},\ldots,\coreset^{t_{K_t}}\}$ until all target models have met the two-condition stopping criterion detailed above. 

We also investigated an alternative average marginalized information-gain strategy. At each iteration, we first select the item that maximizes the expected information gain averaged across all $K_t$ target models: $\alpha_{\text{IG}}(i) = \frac{1}{K_t} \sum_{t=1}^{K_t}  \left({\mathcal H}(R^{t} \cond s_\coreset) - \mathbb{E}_{s_i^t}[{\mathcal H}(R^{t} \cond s_\coreset, s^{t}_i)]\right)$.
After selecting an item, we choose the target model uniformly at random. We present these results in Figure \ref{fig:GPQA-multitarget-extra-active-learning} of Appendix \ref{sec:app:additional_results_benchmarks_multi}.

\section{Related work}\label{sec:related_work}
The issue of computational inefficiency in model evaluation is primarily tackled from two, sometimes intersecting, directions: constructing smaller, static versions of existing benchmarks (or developing methods for reducing benchmarks) \citep{bean2025scales,kipnis2025metabench,zhao2025bento}, and estimating performance without fully evaluating the target model \citep{berrada2025scaling,bowyer2026efficient,fisch2026collaborative,huang2026proeval,kipnis2025metabench,kossen2021active,hsu2026efficient,liao2025toward,liu2026active,polo2024tinybenchmarks,saranathan2025sublime,vivek2024anchor,wang2025cereval,zhang2025how}. 

\paragraph{Automatic Determination of Coreset Size} With the exception of \acro{Cer-Eval} \citep{wang2025cereval}, all methods in the second category assume that the evaluation budget is specified by the user. \acro{Cer-Eval} differs substantially from our method as it does not utilize reference models, relying instead exclusively on the responses from the target model. It repeats the following steps until a strict confidence condition is met: $(i)$ partitioning the benchmark based on previously evaluated items to isolate regions of low variance in the target model's scores, $(ii)$ computing summary statistics for each partition, $(iii)$ identifying the partition that yields the greatest expected reduction in evaluation uncertainty; and $(iv)$ sampling an item from that optimal partition.

Extending methods designed for user-specified coreset sizes to automatically determine the coreset size poses fundamental challenges. This task requires striving for accurate estimation across all coreset sizes, which, as we show in our experiments, is not always achievable. Furthermore, establishing a reliable stopping criterion requires confidence intervals that accurately reflect the true remaining uncertainty, a property that existing formulations often lack.

\paragraph{Performance Modeling and Estimation} From a modeling perspective, many works that leverage reference model information rely on Item Response Theory (IRT) to infer the target model's profile \citep{kipnis2025metabench, liao2025toward, polo2024tinybenchmarks}. However, obtaining reliable parameter estimates with IRT requires on the order of hundreds \citep{jiang2026trust} or thousands \citep{kipnis2025metabench} of reference models. Another line of work \citep{bowyer2026efficient,zhang2025how} frames the problem as a regression task. \ProEval \citep{huang2026proeval} introduces a Gaussian process approach that shares similarities with our Bayesian modeling. However, we depart critically from their formulation to address automatic coreset size determination and more robustly handle the lack of prior information about the target model.

\paragraph{Multi-Target Setting} Finally, to the best of our knowledge, the only other work that leverages performance correlations across multiple target models rather than treating them independently is \cite{fisch2026collaborative}. This approach frames model evaluation as a matrix completion problem. While the method could be run sequentially by randomly sampling the next item and model to evaluate,
its confidence intervals capture only the sampling uncertainty around the mean performance, assuming the benchmark is a finite sample drawn from an infinite population of prompts. Consequently, even when all items in the finite benchmark have been evaluated, the confidence interval does not shrink to zero. In contrast, our Bayesian credible intervals provide a direct probability statement about the true, finite-sample benchmark performance given the observed data, yielding an intuitively appealing and actionable stopping criterion.

\section{Experiments}
We evaluate our approach across seven standard benchmarks collected from three primary sources: the Open LLM Leaderboard \citep{fourrier2024open}, evaluations compiled by \citet{zhang2025how}, and the HELM Lite benchmark \citep{liang2023holistic}.

From the Open LLM Leaderboard, we extracted scores for GPQA \citep{rein2024gpqa}, MMLU-Pro \citep{wang2024mmlupro,hendrycks2021measuring}, BBH \citep{suzgun2023challenging}, ARC-Challenge \citep{clark2018think}, and MuSR \citep{sprague2024musr}. Beyond standard binary scores, we leveraged the leaderboard's raw log-likelihoods to obtain continuous scores via a softmax transformation. These continuous scores allow us to assess the impact of exploiting richer information and mitigate the issue of identical reference model scores. To ensure a representative distribution of model capabilities, we partitioned the available models into five quantiles based on accuracy. We then sampled evenly by selecting the most popular models from each quantile, resulting in an initial set of 300 models. After extracting the data for these selections, we filtered out any records containing NaN values caused by incomplete leaderboard entries, formatting discrepancies, or evaluation timeouts. This curation process yielded the following final model counts per benchmark: GPQA: 102, MMLU-Pro: 165, BBH: 279, ARC-Challenge: 125, and MuSR: 288.

From \cite{zhang2025how}, we extracted binary scores for IFEval \citep{zhou2023instruction} across 448 models. Finally, from HELM Lite (up to version 1.13.0), we extracted the F1 scores for the Natural QA Openbook long-answer scenario \citep{kwiatkowski2019natural} across 44 models.

\subsection{Single-Target Setting}
\paragraph{Baselines}
Since all existing approaches that leverage reference models assume a predefined coreset size, we adapt the most effective methods to operate in the same sequential manner as Algorithm \ref{algorithm:BayesAME}. Specifically, $\coreset$ is iteratively augmented until $\max_{l\in[j-W, j]}(\hat{R}_l)-\min_{l\in[j-W, j]}(\hat{R}_l)\le\epsilon_1$ and $2 z_{0.975} \sqrt{\operatorname{Var}_R}\leq\epsilon_2$ where $\hat R$ and $\operatorname{Var}_R$ are defined below.

\begin{enumerate}
    \item \textbf{\RBayesAME}: A simpler version of \BayesAME that selects items at random.
     \item \textbf{\SeqAPW}: An extension of the anchor points weighted method from \cite{vivek2024anchor}, in which $\coreset$ is augmented by choosing items that maximize space covering, as measured by correlations between the reference models' performances. Benchmark items are partitioned into clusters by assigning each item to its nearest anchor in $\coreset$. 
    $\hat R=\sum_{i \in \coreset} w_is_i,$
    where $w_i$ is a weight proportional to the cluster size, while $\operatorname{Var}_R$ is given by the following estimator of $\operatorname{Var}(R)$:
    \begin{equation}
        \operatorname{Var}_R= s^2 \left(\sum_{i\in \coreset} w_i^2\right)\frac{N-n_{\text{eff}}}{N}, \quad n_{\text{eff}} = \frac{\left(\sum_{i\in \coreset} w_i\right)^2}{\sum_{i\in \coreset} w_i^2},
        \label{eq:SeqAPW_Variance}
    \end{equation}
    where $s$ denotes the sample standard deviation. 
    By construction, $\hat R$ recovers the true performance $R^\star$ once the entire benchmark has been evaluated. Moreover, $\operatorname{Var}_R$ captures the sampling variance arising from the randomness in the coreset selection. Due to the finite-population correction factor, this variance decreases as the coreset grows and converges to zero when the full benchmark is evaluated.
    \item \textbf{\RS}: The simplest baseline that augments $\coreset$ by selecting items at random and estimates the full benchmark performance as the empirical average performance on $\coreset$. $\operatorname{Var}_R$ is given by \eqref{eq:SeqAPW_Variance} using $w_i=1/n$. The same properties regarding $\hat{R}$ and $\operatorname{Var}_R$ discussed for \SeqAPW apply.
    \item \textbf{\BayesAIPW}: A Bayesian ridge regression extension of \AIPW \citep{zhang2025how}, obtained by placing a Gaussian prior distribution on the coefficients $\beta$ of the model $s_i = \phi_i^\top \beta + \varepsilon_i$, where $\phi_i=(s_i^{r_1}, \ldots, s_i^{r_{K_r}})$, and $\varepsilon_i\sim \mathcal{N}(0, \alpha^{-1})$ with $\alpha\sim\operatorname{Gamma}(\alpha_1, \alpha_2)$. 
    By conditioning on $\{(\phi_i,s_i)\}_{i\in\coreset}$, we obtain a posterior distribution over $\beta$ with mean $\mu^\beta_{|\coreset}$ and covariance $\Sigma_{|\coreset}^\beta$. As $\hat{R}$, we use the \AIPW estimator:
    \[\hat{R}=\frac{1}{n}\sum_{i \in \coreset} s_i+\frac{1}{1+\frac{n}{N-n}}\Bigg(\frac{1}{N-n}\sum_{i \in E\setminus \coreset} \mu_{i|\coreset}-\frac{1}{n}\sum_{i \in \coreset} \mu_{i|\coreset}\Bigg),\]
    where $\mu_{i|\coreset}= \phi_i^\top \mu^\beta_{|\coreset}$. Note that $\hat{R}$ recovers the true performance $R^\star$ when all items have been evaluated.
    Posterior uncertainty is obtained by propagating the posterior distribution of $\beta$ through the \AIPW correction, yielding: 
    \[
    \operatorname{Var}_R = \frac{(\bar{\phi}_{E\setminus\coreset}-\bar{\phi}_\coreset)^\top \Sigma_{|\coreset}^\beta(\bar{\phi}_{E\setminus\coreset} - \bar{\phi}_\coreset) + \hat{\alpha}^{-1}\left(\tfrac{1}{n} + \tfrac{1}{N-n}\right)}{\left(1 + \frac{n}{N-n}\right)^2},
    \]
    where $\bar{\phi}_\coreset = \frac{1}{n}\sum_{i \in \coreset} \phi_i$, $\bar{\phi}_{E\setminus\coreset} = \frac{1}{N-n}\sum_{i \in E\setminus\coreset} \phi_i$, and $\hat{\alpha}$ is the posterior mean of $\alpha$.
    $\coreset$ is augmented by selecting items at random. 
    Figure \ref{fig:GPQA-MMLUPRO-SAMPLING-SCORING-AIPW} in Appendix \ref{sec:app:additional_results_benchmarks} shows that this Bayesian extension matches or exceeds the performance of the original \AIPW method.
    \item \textbf{\BayesRSL}: A Bayesian ridge regression extension of \RSL \citep{zhang2025how}, obtained by placing a Gaussian prior on the coefficients $\beta$ of the model $R =  s_{\coreset}^\top \beta + \varepsilon$, where $\varepsilon \sim \mathcal{N}(0,\alpha^{-1})$, $\alpha\sim\operatorname{Gamma}(\alpha_1, \alpha_2)$. We perform inference on $\beta$ by conditioning on $\{(s_\coreset^{r_k}, R^\star_{r_k})\}_{k=1}^{K_r}$. This yields a posterior distribution over $\beta$ with mean $\mu^\beta_{|\coreset}$ and covariance $\Sigma_{|\coreset}^\beta$.    
    We then obtain $\hat R$ and $\operatorname{Var}_R$ as: 
    \begin{align*}
    \hat R = s_{\coreset}^\top\mu^\beta_{|\coreset}, \quad 
    \operatorname{Var}_R = s_{\coreset}^{\top}\Sigma^\beta_{|\coreset}s_{\coreset} + \hat{\alpha}^{-1},
    \end{align*}
    where $\hat{\alpha}^{-1}$ is the posterior mean of $\alpha^{-1}$.
    \item \textbf{\ProEval}: The method in \cite{huang2026proeval} for the setting with no prior performance data for the target model (called New Model setting with Score Features in the original paper), without abstention rule. The method defines the target model's performance as $R=\frac{1}{N}\sum_{i=1}^N \theta_i$, where $\theta=(\theta_1,\ldots,\theta_N)$ is a Gaussian random vector whose prior covariance matrix is given by the empirical covariance of the reference model scores. $\hat{R}=\mathbb{E}[R \cond s_\coreset]$ and $\operatorname{Var}_R=\operatorname{Var}(R\cond s_\coreset)$. Because $R$ is defined as the average of the latent abilities, the estimator $\hat{R}$ can deviate significantly from the true performance $R^\star$ even for large coreset sizes, as discussed in \secref{sec:related_work}. The confidence interval induced by $\operatorname{Var}_R$ can be interpreted as a Bayesian credible interval.
    \item \textbf{\IRT}: Two variants of \acro{IRT} where $\coreset$ is selected randomly, depending on the score type: one for binary scores based on the 2-PL model \citep{polo2024tinybenchmarks} (\acro{p-IRT}) and one for continuous scores based on the Beta distribution, specifically the extension of \cite{noel2007beta} provided in \cite{chen2019beta3}. The item parameters are learned using the reference models' scores, and the scores $s_{\coreset}$ are used to fit the target model-specific parameters. As the performance estimator, we deploy \eqref{eq:performance_estimate} with $w=1/N$. Because we fit the model parameters separately for each benchmark, treating them as scalars rather than vector-valued as in \cite{polo2024tinybenchmarks}, applying the generalized variant (\acro{gp-IRT}) does not improve performance. 
    We fit the parameters using MCMC. This provides samples from the posterior distribution of the parameters, allowing us to directly obtain the posterior distribution of $R$ and its variance, which we use as $\hat{R}$ and  $\operatorname{Var}_R$. Due to the computational cost of MCMC, we update the parameters only every 50 evaluations (resulting in piecewise-constant curves in the plots).
\end{enumerate}
%Because the baselines do not all produce uncertainty estimates with the same interpretation, the stopping-rule comparison should be read as an empirical comparison of practical sequential adaptations rather than a comparison of identically calibrated confidence statements.

\paragraph{Metrics} We evaluate our framework using four metrics: \RMSELog, \RMSEGain, \EWIRMSE, and \CAT. 
By computing the Root Mean Square Error (RMSE) over 200 initialization seeds, all metrics intrinsically account for the empirical variance of the performance estimates across different samples, thereby providing a rigorous evaluation of each method's precision. 
The first three metrics evaluate the methods when stopping at fixed coreset sizes $n=1,\ldots,N$ (thereby not using the $\epsilon_1,\epsilon_2$ stopping criterion). 
Evaluating across the entire range of coreset sizes as input stopping points enables us to determine whether any method dominates the others regardless of the stopping point. In contrast, the \CAT measures estimation error at the specific coreset sizes reached when the $\epsilon_1,\epsilon_2$ stopping criterion is met, demonstrating which method achieves the lowest estimation error at the smallest automatically determined coreset size.
\begin{enumerate}
\item \textbf{\RMSELog}: The RMSE on a logarithmic scale, multiplied by 100.
\item \textbf{\RMSEGain}: The relative improvement of a given method over the \RS baseline, calculated as: 
\begin{equation*}
\RMSEGain(\text{Method}) = 20 \log_{10} \left( \frac{\text{RMSE}(\text{\RS})}{\text{RMSE}(\text{Method})} \right).
\end{equation*}
A positive gain indicates a relative reduction in estimation error compared to the baseline.
\item \textbf{\EWIRMSE}: The weighted integral $100 \int_0^{N} n' \text{RMSE}(n')\, dn'$, which ensures that late-stage errors are penalized more heavily than early-stage ones.
\item \textbf{\CAT}: The RMSE (multiplied by 100) computed at the coreset sizes dynamically determined 
by the $\epsilon_1,\epsilon_2$ stopping criterion, plotted against the average coreset size $\bar{n}$ across seeds. We vary $\epsilon_2\in \{0.005, 0.0075, 0.01, 0.015, 0.02, 0.025, 0.03\}$, and fix $\epsilon_1 = 0.005$ since empirical results indicate that $\epsilon_2$ is the primary factor governing the stopping decision.
\end{enumerate}
We also evaluate our method using Spearman's rank correlation to assess how well the estimated performance preserves the true relative ordering of the models. The results, presented in Appendix \ref{sec:app:additional_results_benchmarks}, demonstrate that our method works well for tasks such as model comparison and ranking.

\paragraph{Interpolation and Extrapolation Reference Setup}
To investigate the interpolation and extrapolation regimes with granularity, we partition the available models into a pool of potential reference models and a fixed set of 10 target models, following an approach similar to \cite{zhang2025how}. 
In the interpolation regime, the target models are selected uniformly at random, with all remaining models assigned to the reference pool. In the extrapolation regime, we induce a distribution shift by designating the 10 highest-performing models as the target set, discarding the next 20\% of models to create a performance gap, and using the remaining lower-performing models as potential reference models. To assess the impact of reference model quantity, we vary the size of the set of reference models by sampling 10\%, 50\%, or 90\% of the reference pool uniformly at random.

\begin{figure}[t!]
\centering
\renewcommand{\arraystretch}{1.2} 
\begin{tabular}{c @{\hspace{3pt}} c @{} c}
% --- X-Axis Column Headers (Top) ---
& 
% Headers for the FIRST 2x3 PDF
\makebox[0.22\textwidth][c]{\scriptsize{Interpolation}} \makebox[0.22\textwidth][c]{\scriptsize{Extrapolation}} &
% Headers for the SECOND 2x3 PDF 
\makebox[0.22\textwidth][c]{\scriptsize{Interpolation}} \makebox[0.22\textwidth][c]{\scriptsize{Extrapolation}} \\

\begin{tabular}{@{}c@{}}
    \rotatebox{90}{\scriptsize{10\%}} \\[1.5cm] %
    \rotatebox{90}{\scriptsize{50\%}} \\[1.5cm] %
    \rotatebox{90}{\scriptsize{90\%}}
\end{tabular} &
% First 2x3 PDF
\includegraphics[width=0.48\textwidth, valign=c]
{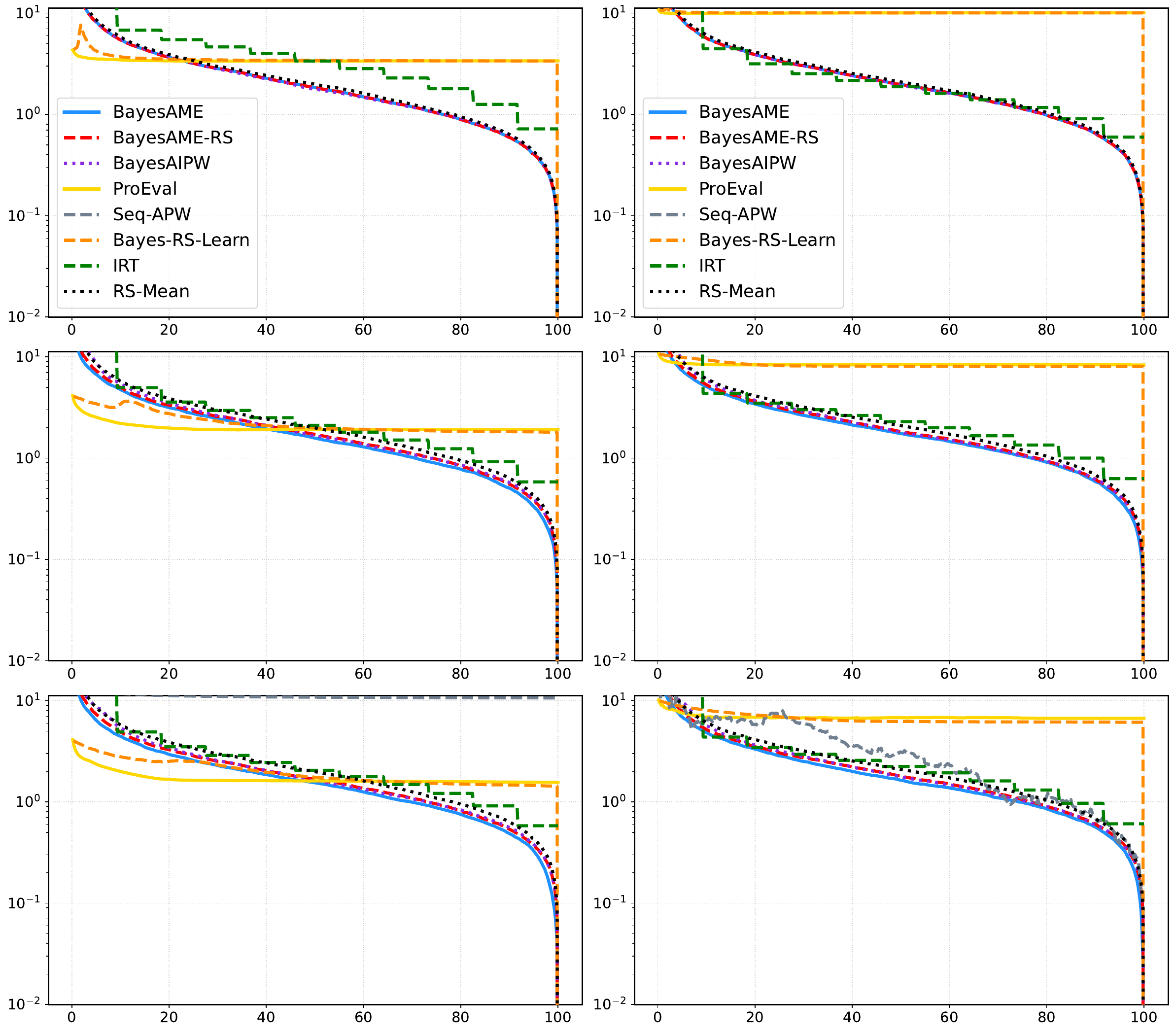} &
% Second 2x3 PDF
\includegraphics[width=0.48\textwidth, valign=c]
{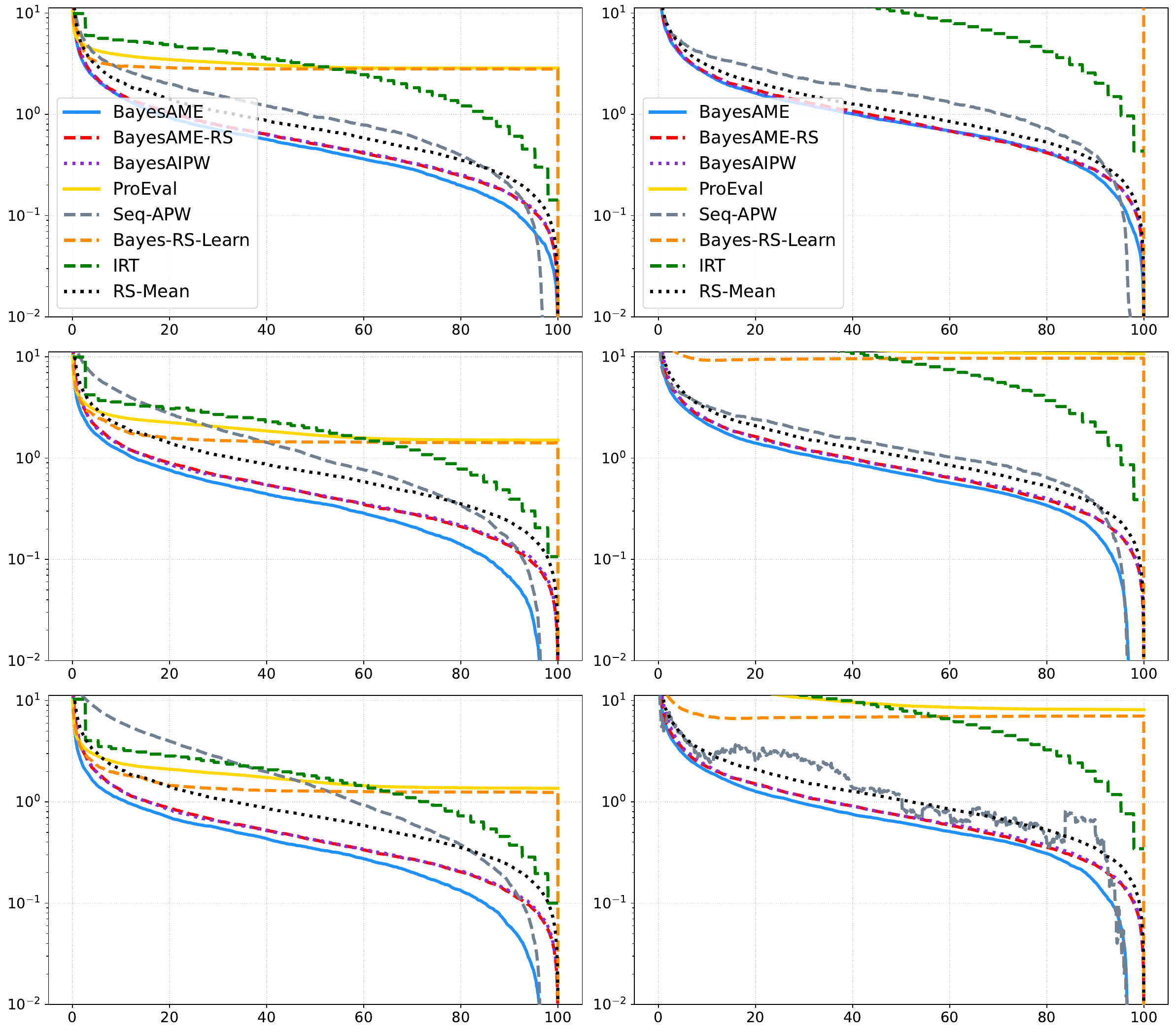} \\
\end{tabular}
\begin{tabular}{c @{\hspace{3pt}} c @{} c}
% --- X-Axis Column Headers (Top) ---
& 
% Headers for the FIRST 2x3 PDF
\makebox[0.22\textwidth][c]{\scriptsize{Interpolation}} \makebox[0.22\textwidth][c]{\scriptsize{Extrapolation}} &
% Headers for the SECOND 2x3 PDF 
\makebox[0.22\textwidth][c]{\scriptsize{Interpolation}} \makebox[0.22\textwidth][c]{\scriptsize{Extrapolation}} \\

\begin{tabular}{@{}c@{}}
    \rotatebox{90}{\scriptsize{10\%}} \\[1.5cm] %
    \rotatebox{90}{\scriptsize{50\%}} \\[1.5cm] %
    \rotatebox{90}{\scriptsize{90\%}}
\end{tabular} &
% First 2x3 PDF
\includegraphics[width=0.48\textwidth, valign=c]
{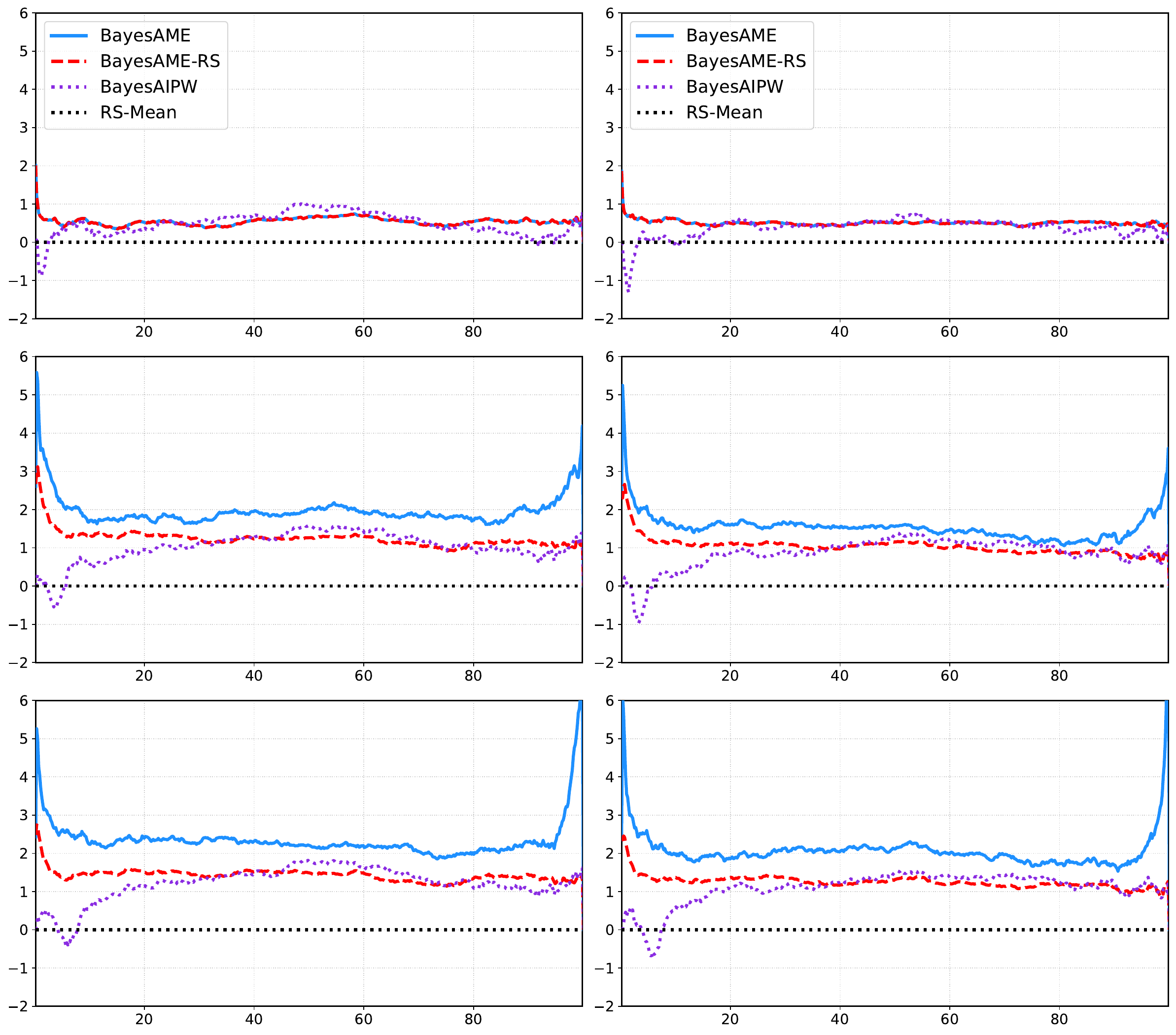} &
% Second 2x3 PDF
\includegraphics[width=0.48\textwidth, valign=c]{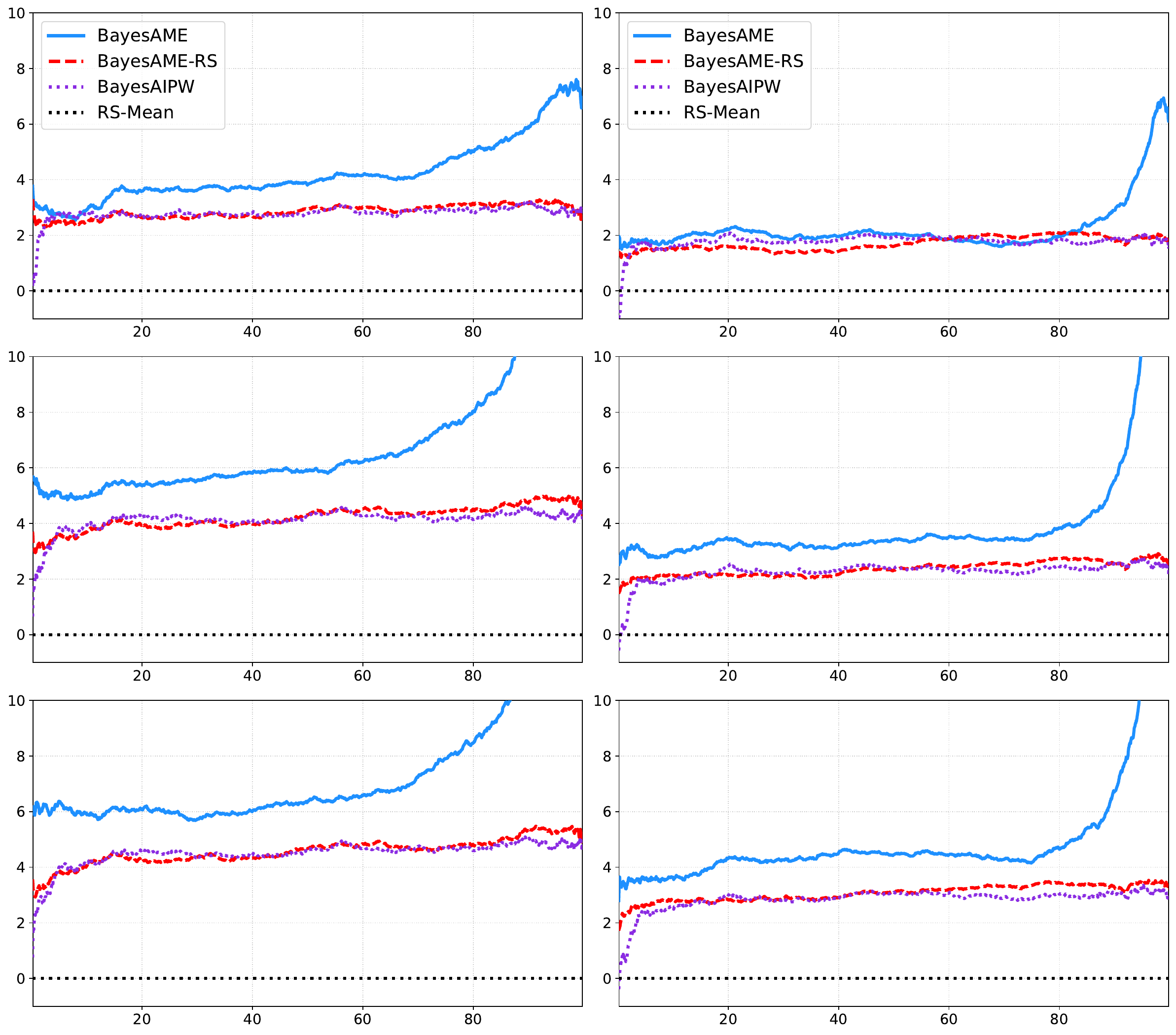} \\
\end{tabular}
\caption{\textbf{Single-target Setting.} \textbf{GPQA} with binary scores (left two columns) and \textbf{MMLU-Pro} with continuous scores (right two columns). \RMSELog (top three rows) and \RMSEGain (bottom three rows) across varying proportions of reference models (10\%, 50\%, and 90\%) for the interpolation and extrapolation regimes. The $x$-axis denotes the coreset size $n$ as a percentage of the benchmark size $N$.}
\label{fig:GPQA-MMLUPro-RMSELog-GAIN}
\end{figure}
\subsubsection{Results}
Figure \ref{fig:GPQA-MMLUPro-RMSELog-GAIN} shows the \RMSELog (top three rows) and \RMSEGain (bottom three rows) for GPQA with $s_i\in\{0,1\}$ (left two columns) and MMLU-Pro with $s_i\in[0,1]$ (right two columns) across different percentages of reference models (10\%, 50\%, and 90\%) for the interpolation and extrapolation regimes. Overall, when considering the entire range of coreset sizes, \BayesAME consistently achieves the lowest \RMSELog, demonstrating superior estimation accuracy and precision. The relative gains of \BayesAME and the two other most competitive methods, \RBayesAME and \BayesAIPW, over the \RS baseline can be fully appreciated in the \RMSEGain plots (bottom three rows), which exclude the less competitive methods (\SeqAPW, \BayesRSL, \ProEval, and \IRT) for clarity. The plots also illustrate that the gain scales positively with the number of reference models, and that this trend is more pronounced for \BayesAME. This confirms that a richer reference pool establishes a stronger prior, which \BayesAME successfully exploits to isolate and select the most critical benchmark items. In the two cases where $B < N - N_0$, which occur for GPQA with 10\% of reference models, \BayesAME and \RBayesAME coincide by construction and perform comparably to \BayesAIPW. 
 
\begin{figure}[t!]
\centering
\renewcommand{\arraystretch}{1.2} 
\begin{tabular}{c @{\hspace{3pt}} c @{} c}
% --- X-Axis Column Headers (Top) ---
& 
% Headers for the FIRST 2x3 PDF
\makebox[0.22\textwidth][c]{\scriptsize{Interpolation}} \makebox[0.22\textwidth][c]{\scriptsize{Extrapolation}} &
% Headers for the SECOND 2x3 PDF 
\makebox[0.22\textwidth][c]{\scriptsize{Interpolation}} \makebox[0.22\textwidth][c]{\scriptsize{Extrapolation}} \\

\begin{tabular}{@{}c@{}}
    \rotatebox{90}{\scriptsize{10\%}} \\[1.5cm] %
    \rotatebox{90}{\scriptsize{50\%}} \\[1.5cm] %
    \rotatebox{90}{\scriptsize{90\%}}
\end{tabular} &
% First 2x3 PDF
\includegraphics[width=0.48\textwidth, valign=c]
{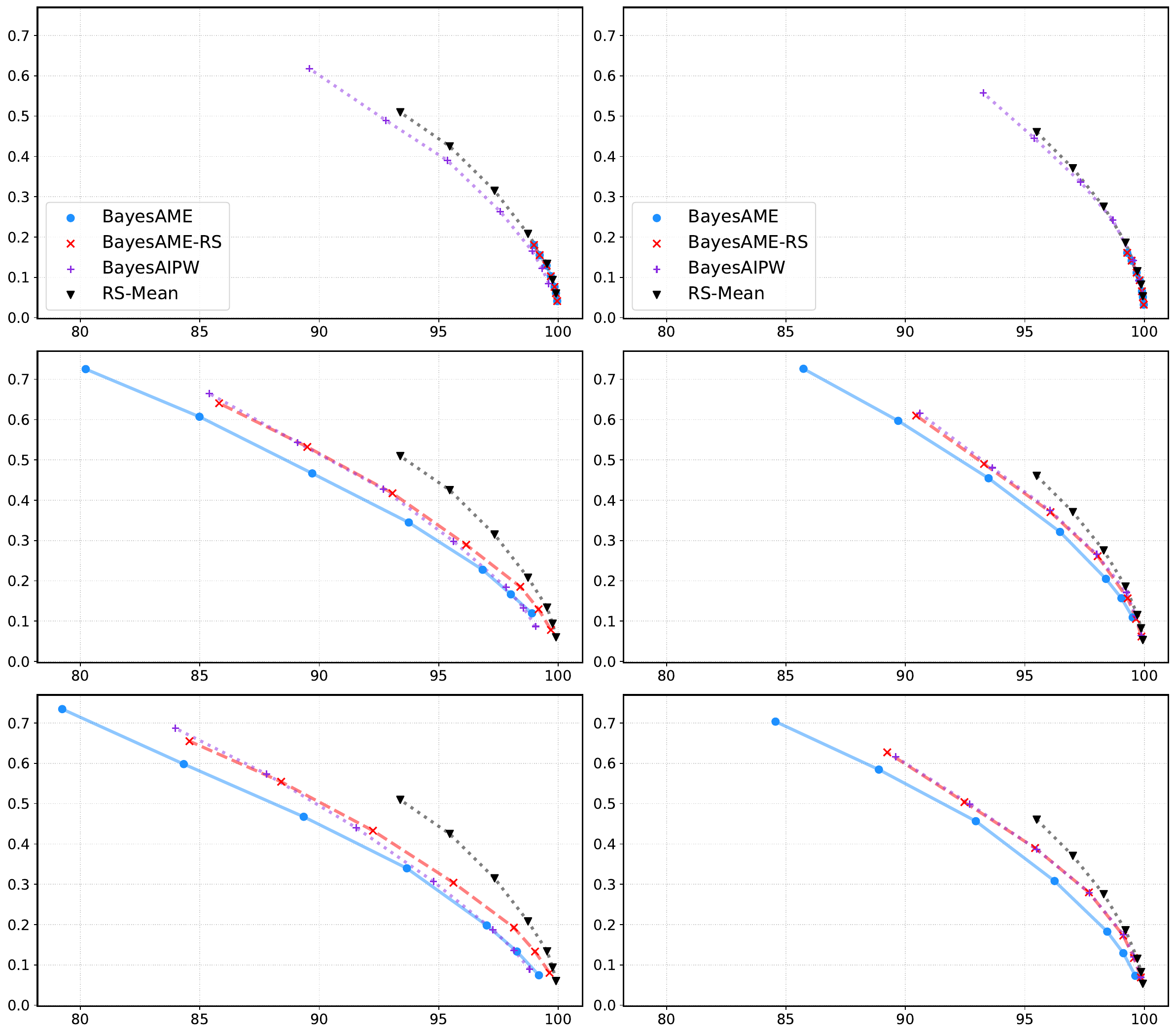} &
% Second 2x3 PDF
\includegraphics[width=0.48\textwidth, valign=c]{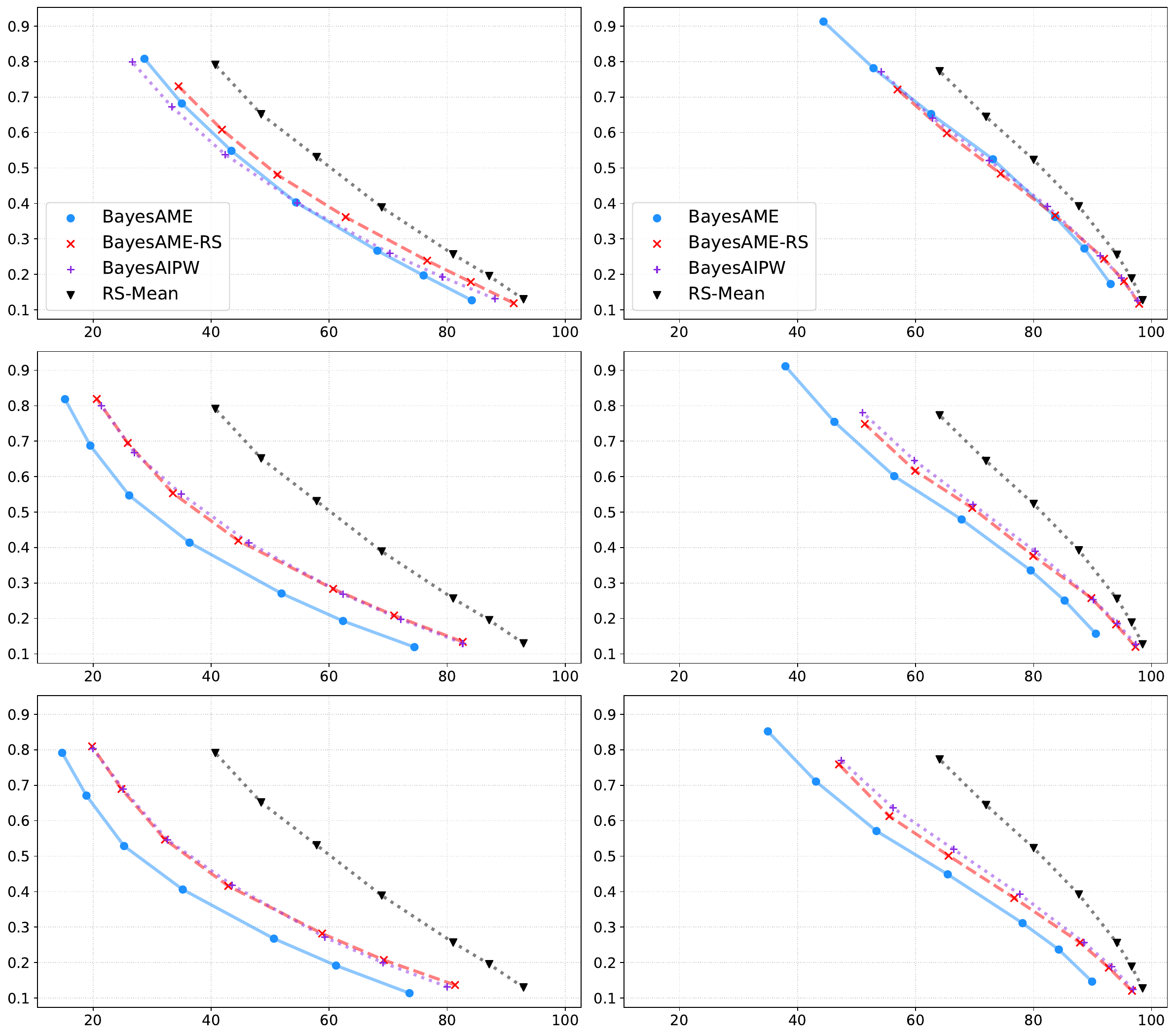} \\
\end{tabular}
\caption{\textbf{Single-target Setting.} \textbf{GPQA} with binary scores (left two columns) and \textbf{MMLU-Pro} with continuous scores (right two columns). \CAT across varying proportions of reference models (10\%, 50\%, and 90\%) for the interpolation and extrapolation regimes. The $x$-axis denotes the average coreset size $\bar{n}$ as a percentage of the benchmark size $N$.}
\label{fig:GPQA-MMLUPro-CAT}
\end{figure}

Figure \ref{fig:GPQA-MMLUPro-CAT} displays the \CAT for GPQA with $s_i\in\{0,1\}$ (left two columns) and MMLU-Pro with $s_i\in[0,1]$ (right two columns). By providing the most accurate performance estimates for any threshold value at the lowest cost across both the interpolation and extrapolation regimes, \BayesAME establishes a strictly superior Pareto frontier.

Plots corresponding to Figures \ref{fig:GPQA-MMLUPro-RMSELog-GAIN} and \ref{fig:GPQA-MMLUPro-CAT} for the remaining benchmarks, which show similar conclusions, are provided in Appendix \ref{sec:app:additional_results_benchmarks}. We summarize these results for 50\% of reference models via the \EWIRMSE in Table \ref{table:BudgetIntegratedRSME}. As the table shows, \BayesAME outperforms the other methods in nearly all cases. 
For binary scores, we have $B<N-N_0$ for MMLU-Pro, ARC-Challenge, and MuSR in both the interpolation and extrapolation regimes, and for IFEVAL in the extrapolation regime. For continuous scores, we have that $B<N-N_0$ up to some tolerance for ARC-Challenge in the extrapolation regime. In these cases, \RBayesAME coincides with \BayesAME by construction.
The table clearly quantifies the overall benefit of using richer continuous scores over binary scores. In Appendix \ref{sec:app:additional_results_benchmarks}, we demonstrate this advantage via the other three metrics.

Our results demonstrate that active coreset selection becomes significantly advantageous over random coreset selection whenever reference models provide sufficiently unique item representations (e.g., via larger reference pools or continuous scores), and show that it is possible to strongly outperform \RS. While non-unique representations cause \BayesAME to default to random selection (matching \RBayesAME and \BayesAIPW), richer item signals enable active selection to effectively target informative items and strongly outperform \RS, addressing the concerns raised by \cite{zhang2025how}.

\subsection{Multi-Target Setting}
We compare the multi-target version of \BayesAME (\secref{sec:bayesAME_multiple_target}) against its single-target counterpart (\secref{sec:bayesAME_single_target}) applied independently to each target model. To evaluate the multi-target formulation under varying degrees of correlation, we construct sets of 10 target models exhibiting either low or high correlation. Specifically, we form the low-correlation set using a random selection of 10 models, and the high-correlation set using the 10 most correlated models overall. 
We construct the set of reference models by sampling 50\% of the reference model pool uniformly at random.

\begin{figure}[t]
\centering
\renewcommand{\arraystretch}{1.2} 

\begin{tabular}{c @{\hspace{10pt}} c @{} c}
% --- X-Axis Column Headers (Row 1) ---
% --- X-Axis Column Headers (Row 2) ---

% Headers for the FIRST  PDF
\makebox[0.22\textwidth][c]{\scriptsize Low correlation} \makebox[0.22\textwidth][c]{\scriptsize High correlation} 
% Headers for the SECOND PDF 
\makebox[0.22\textwidth][c]{\scriptsize Low correlation} \makebox[0.22\textwidth][c]{\scriptsize High correlation} \\

% --- Y-Axis Label and Images ---
% First PDF
\includegraphics[width=0.48\textwidth, valign=c]{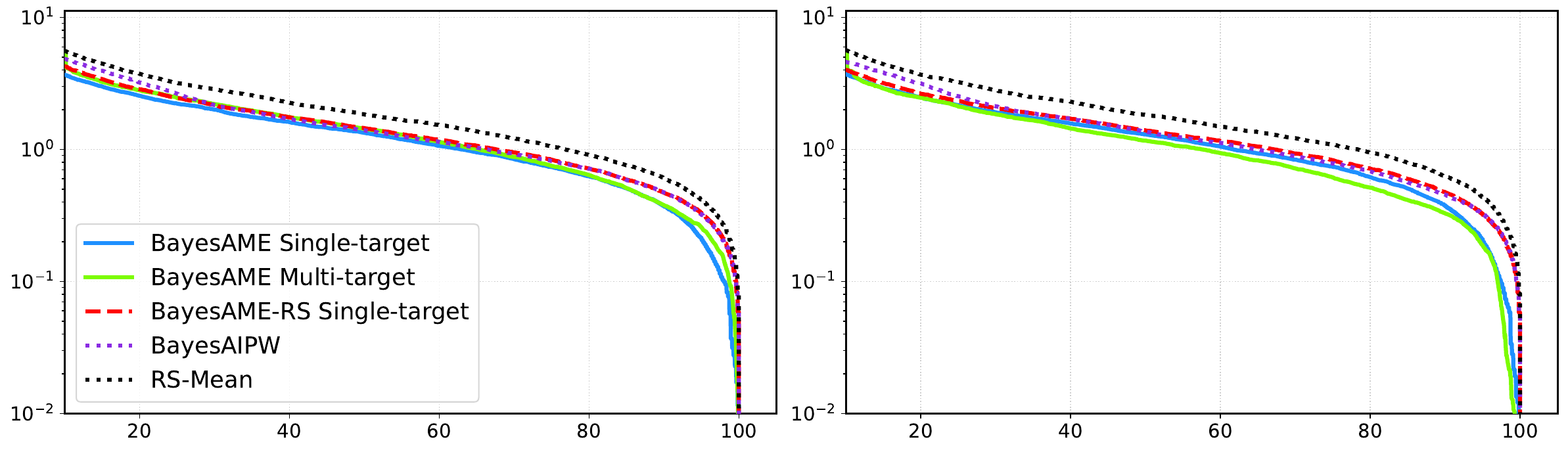} 
\includegraphics[width=0.48\textwidth, valign=c]{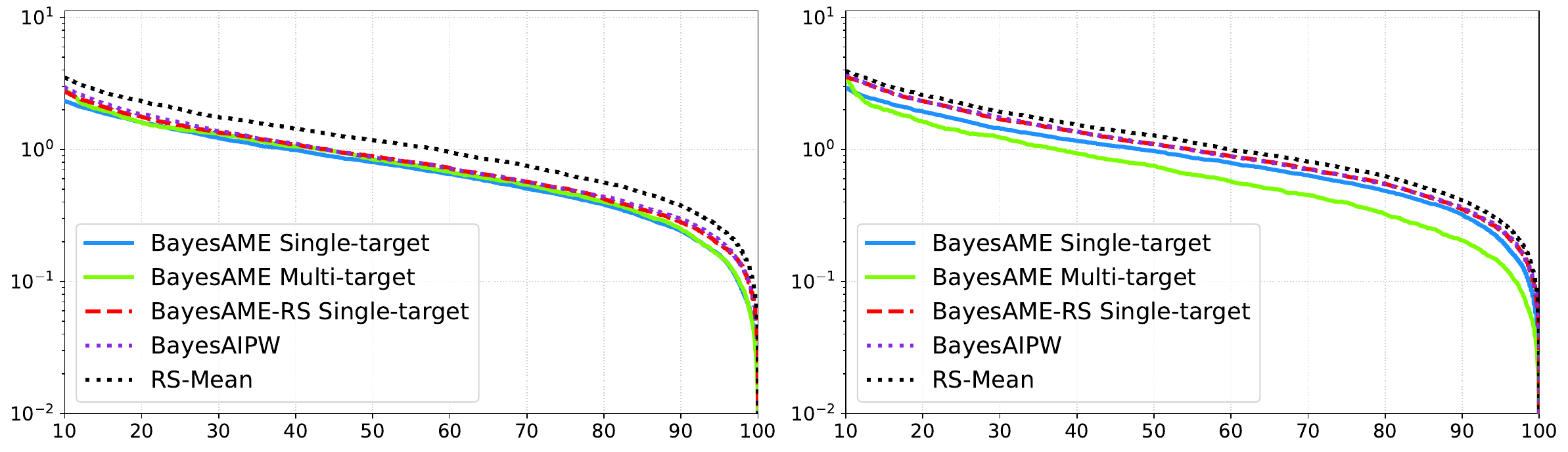} \\
% Second PDF
\includegraphics[width=0.48\textwidth, valign=c]{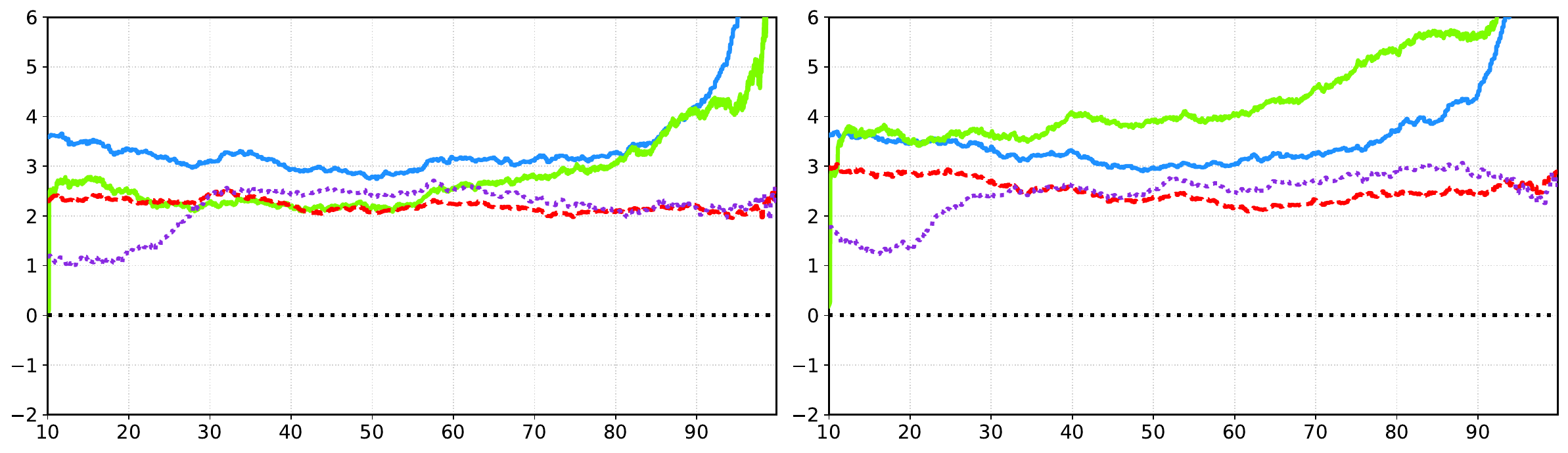} \includegraphics[width=0.48\textwidth, valign=c]{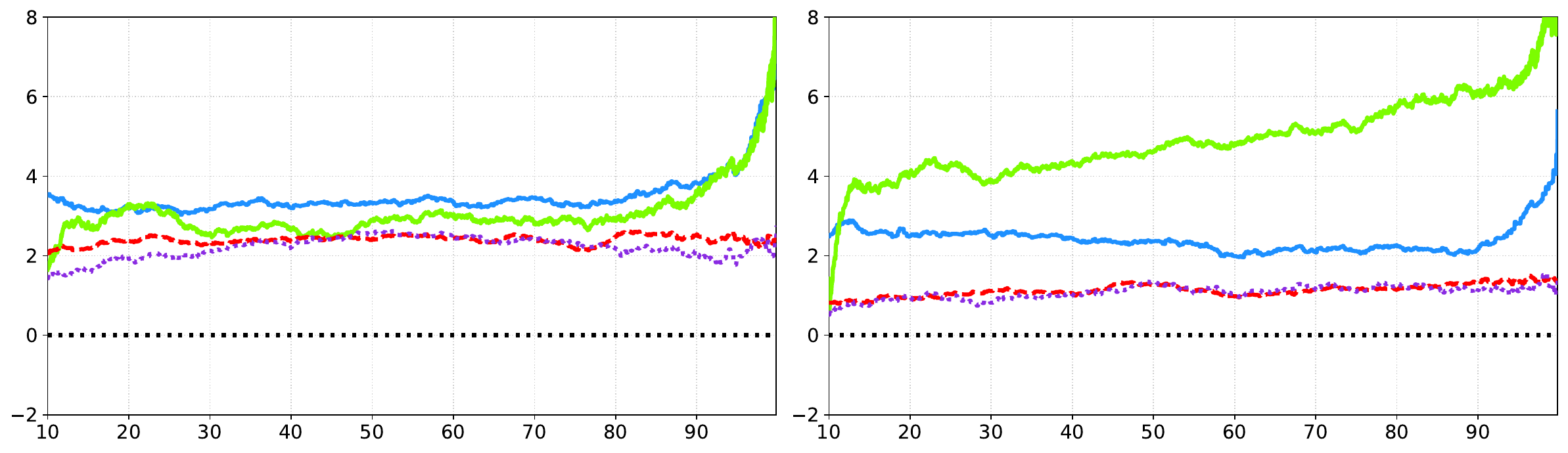} 
\\
% Second PDF
\includegraphics[width=0.48\textwidth, valign=c]{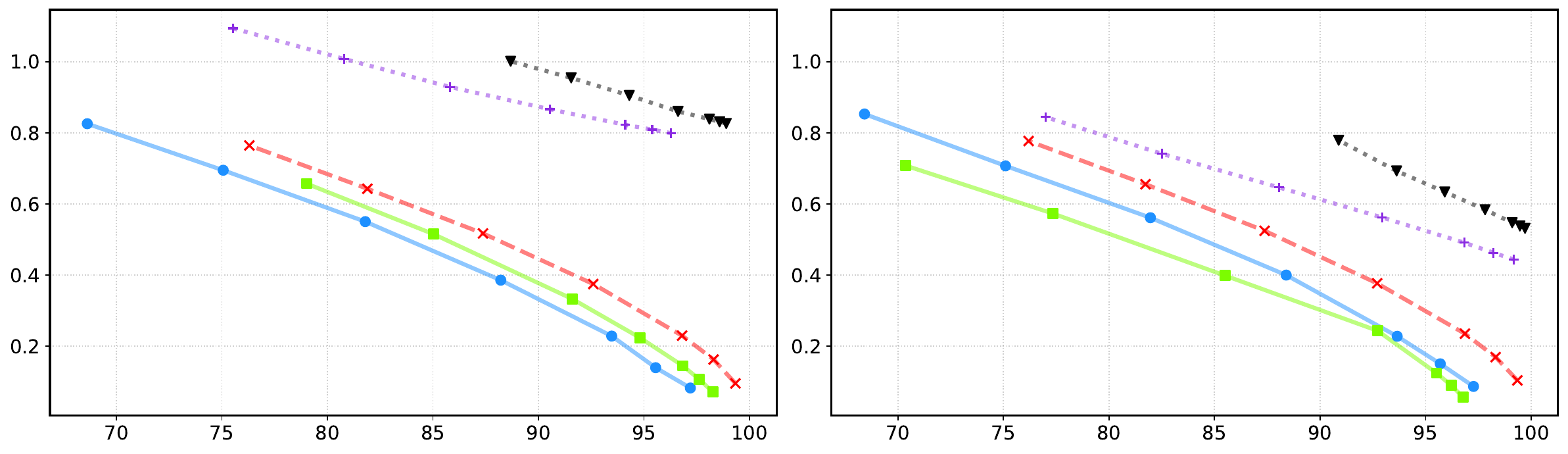} \includegraphics[width=0.48\textwidth, valign=c]{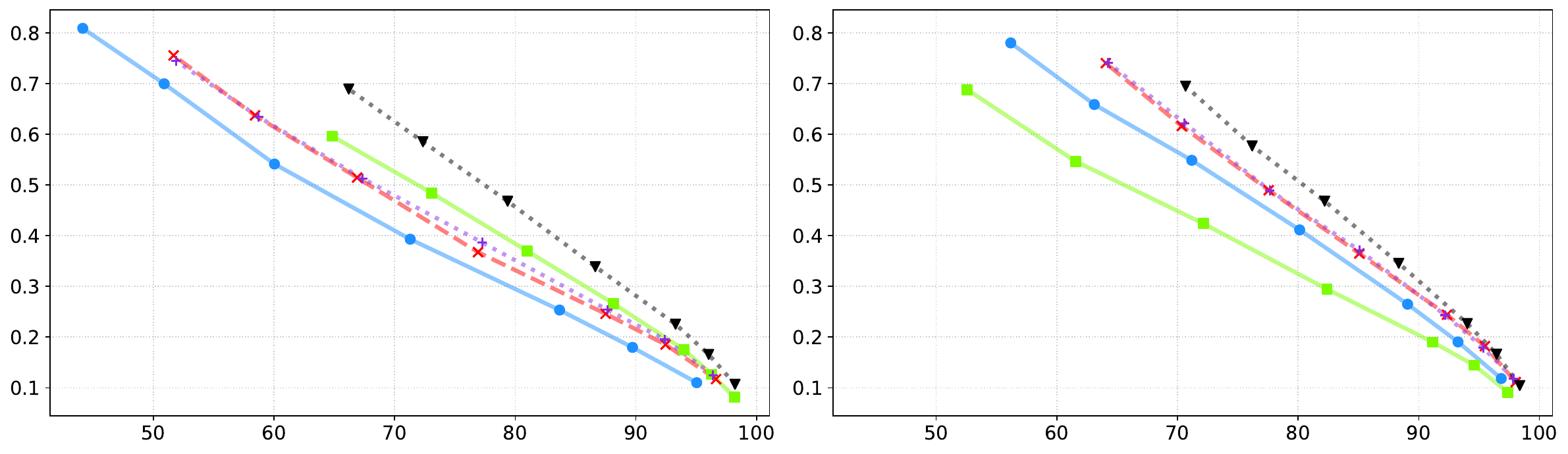} 
\\
\end{tabular}
\caption{\textbf{Multi-target Setting.} \textbf{IFEval} with binary scores (left two columns) and \textbf{GPQA} with continuous scores (right two columns). The rows represent \RMSELog, \RMSEGain, and \CAT, respectively, for 50\% of reference models for the low and high correlation regimes. 
}
\label{fig:GPQA-IFEVAL-multitarget}
\end{figure}

Figure \ref{fig:GPQA-IFEVAL-multitarget} shows \RMSELog, \RMSEGain, and \CAT for IFEval with $s_i\in\{0,1\}$ (left two columns) and GPQA with $s_i\in[0,1]$ (right two columns). Additional results for GPQA with $s_i\in\{0,1\}$ are given in Fig. \ref{fig:GPQA-multitarget} of Appendix \ref{sec:app:additional_results_benchmarks_multi}. The primary conclusion drawn from these plots is that jointly evaluating multiple target models provides a significant efficiency advantage when the models exhibit correlated behaviors. Specifically, in the high-correlation regime, \BayesAME \acro{Multi-target} demonstrates a steeper decline in \RMSELog and reach a higher \RMSEGain much earlier compared to \BayesAME \acro{Single-target}. This is further reflected in the \CAT plots, where \BayesAME \acro{Multi-target} establishes a strictly superior Pareto frontier, achieving the same level of estimation accuracy at a notably lower coreset size across all threshold values. However, the low-correlation regime plots reveal a practical nuance of the multi-target framework. When target models do not share meaningful behavioral similarities, attempting to jointly model their latent abilities introduces a degree of negative transfer. As observed in the \CAT plots, the multi-target approach slightly underperforms \BayesAME \acro{Single-target}, exhibiting higher estimation errors for a given evaluation budget. Crucially, however, this degradation is not catastrophic, the performance of the multi-target method degrades gracefully to match that of the competitive baseline \BayesAIPW. This suggests that while the active selection process expends some resources attempting to learn nonexistent cross-model correlations, the underlying framework remains highly robust. Nonetheless, to maximize evaluation efficiency, practitioners should apply the multi-target extension primarily when there is a reasonable prior expectation of correlated performance. 

\begin{table*}[t!]
    \centering
    \resizebox{\textwidth}{!}{
    \begin{tabular}{ll cccccccc}
        \toprule
        \multicolumn{2}{l}{\textbf{\textsc{Benchmarks}}} & \textbf{\BayesAME} & \textbf{\RBayesAME} & \textbf{\RS} & \textbf{\SeqAPW} & \textbf{\BayesAIPW} & \textbf{\BayesRSL} & \textbf{\ProEval} & \textbf{\IRT} \\
        
        \midrule
        \multicolumn{10}{l}{\textit{\textsc{Binary}}} \\
        \midrule
        \textsc{GPQA}          & Interpolation & \textbf{61.54} & 66.38 & 76.48& 866.61 & 67.21&99.06 & 96.29 & 89.61\\
                               & Extrapolation & \textbf{68.91}& 72.65 & 81.73&751.08 & 73.41& 403.29 & 418.11 & 91.40 \\ \addlinespace
        \textsc{MMLU-Pro}      & Interpolation & \textbf{30.18}& \textbf{30.18} & 40.20&  1141.06& 30.68& 90.59 & 95.56 & 46.38 \\
                               & Extrapolation & \textbf{37.20} & \textbf{37.20}& 44.89 & 821.40 & 37.69& 670.20& 701.84 & 156.33\\ \addlinespace
        \textsc{BBH}           & Interpolation & \textbf{31.70} & 36.35 & 50.42 & 495.01 & 35.54 &  50.63& 57.82 & 50.95\\
                               & Extrapolation & \textbf{38.35} & 42.02& 52.59& 448.04& 41.34 & 448.53 & 600.98 & 158.41 \\ \addlinespace
        \textsc{ARC-Challenge} & Interpolation & \textbf{36.53}& \textbf{36.53} & 56.09& 1391.22 & 36.68& 70.84 & 74.06  & 44.19\\
                               & Extrapolation & \textbf{36.56}& \textbf{36.56}& 53.21& 768.83& 38.03& 327.45 & 364.21  & 88.58 \\ \addlinespace
        \textsc{MuSR}          & Interpolation & \textbf{47.86} & \textbf{47.86} & 69.27& 949.90& 49.18& 69.12& 68.54 & 84.51\\
                               & Extrapolation & \textbf{54.19}& \textbf{54.19}& 71.28 &719.86 & 56.72 &289.76 & 353.44 &  104.41\\ \addlinespace
        \textsc{IFEval}        & Interpolation & \textbf{49.53} & 56.14& 72.45& 1464.91 & 56.71 & 76.22 & 78.53 & 79.06 \\
                               & Extrapolation & \textbf{58.08} & \textbf{58.08} & 64.54 & 488.54 & 60.52& 466.45& 1017.33 & 212.27 \\
        
        \midrule
        \multicolumn{10}{l}{\textit{\textsc{Continuous}}} \\
        \midrule
        \textsc{GPQA}                & Interpolation & \textbf{30.58} & 34.42& 45.25 &50.22 &35.21 & 49.08& 54.03 & 81.44\\
                                     & Extrapolation & \textbf{40.40} & 40.75& 56.64 & 62.52& 46.77& 262.65& 313.65 &  142.56\\ \addlinespace
        \textsc{MMLU-Pro}            & Interpolation & \textbf{13.11}&17.10 & 27.80 & 39.99& 17.19& 72.66 & 82.62 & 66.44\\
                                     & Extrapolation & \textbf{26.82} & 31.10 & 40.83 & 47.07 & 31.33& 482.43 & 570.76 & 311.12 \\ \addlinespace
        \textsc{BBH}                 & Interpolation & \textbf{13.04}& 17.54& 35.21& 49.38& 18.01 & 26.13 & 40.45 & 61.84\\
                                     & Extrapolation & \textbf{29.80} & 35.57 & 47.10 & 48.50 & 33.64& 303.09 & 612.49 & 388.23\\ \addlinespace
        \textsc{ARC-Challenge}       & Interpolation & \textbf{26.04} &26.47 & 47.67 & 93.82 & 26.19 &  46.84 & 53.57 & 46.10 \\
                                     & Extrapolation & 33.16 & 33.16 & 49.07& 43.52& \textbf{33.05}& 292.45 & 354.75 & 144.55 \\ \addlinespace
        \textsc{MuSR}                & Interpolation  & \textbf{27.83}& 33.42& 50.67& 76.21& 34.34& 50.02& 62.88 & 57.39\\
                                     & Extrapolation& \textbf{36.96}& 42.56& 60.19 & 70.64 & 43.69 & 246.90 & 372.08 & 211.37 \\ \addlinespace
        \textsc{Natural QA Openbook} & Interpolation & \textbf{28.26} & 29.68 & 45.91 & 1185.87 & 29.12 & 275.90 & 286.37 & 129.11\\
                                     & Extrapolation & \textbf{20.83}& 28.36 & 41.76 & 1044.02 & 27.70 & 226.18 & 262.55 &127.84  \\
        \bottomrule
    \end{tabular}
    }
    \caption{\textbf{Single-target Setting.} \EWIRMSE for 50\% of reference models. 
    }
    \label{table:BudgetIntegratedRSME}
\end{table*}
\section{Discussion and Conclusions}
A primary objective of this work was to provide an efficient model evaluation method that is capable of automatically determining a coreset size when reliable performance estimation takes priority over efficiency, while remaining flexible to accept a predefined coreset size. To this end, we introduced a simple, high-performing method, \BayesAME. Furthermore, we proposed an extension to the multi-target setting that leverages performance correlations among target models to further reduce the coreset size.

Another objective was to address skepticism in recent literature, which suggests that random coreset selection and simply using a random sample mean baseline gives competitive performance. Our study overturns this belief by showing that, when leveraging richer information (such as continuous scores and sufficiently varied sets of reference models), looking across the entire spectrum of coreset sizes, and accounting for the empirical variance of the performance estimates across different samples, non-random coreset selection and more robust performance estimation remain highly advantageous. 

Interestingly, while our results challenge the literature's reliance on random selection, they do validate the underlying intuition that simple methods are difficult to outperform. We found that introducing highly sophisticated, principled modeling of the scores did not translate to better performance. For example, when dealing with binary or bounded continuous scores, a more principled model of the latent abilities using logit transformations (as described in Appendix \ref{sec:app:logit}) failed to outperform the coarser joint Gaussian approach. Though relying on an improper prior for bounded data, the Gaussian approach proved far more effective because it allowed for highly efficient, closed-form evaluations of information gain. Similarly, our attempts to find richer, more complex representations of items (for example, including prompt embeddings), rather than simply relying on reference scores, did not yield tangible improvements. 

There are several limitations of \BayesAME that provide promising avenues for future research. Currently, our framework assumes scalar scores. A natural step is to adapt it to handle more complex, multi-dimensional, or unstructured evaluation formats. 
%Furthermore, future work could expand the framework to multi-step agentic trajectories. This could involve modeling the latent abilities required for planning and tool use, while actively sampling the most informative simulation environments to reduce the high computational costs associated with agent evaluation. 
In addition, future iterations could investigate ways to integrate architecture details, training hyperparameters, or training data distributions to further refine the prior covariance.

\section*{Acknowledgments}
The authors would like to thank Michalis Titsias for valuable discussions and providing feedback on the manuscript.

\bibliographystyle{abbrvnat}
\bibliography{refs}

\newpage
\appendix
\section{\BayesAME Details}

\subsection{Posterior Distribution Computation}\label{sec:app:equivalent_bucket_matrix_H}
In this section, we show the equivalence between $p(\theta\cond s_\coreset)$ and $p(\theta\cond \bar{s}_{\coreset_{1: B}})$ stated in \secref{sec:bayesAME_single_target}. Based on this, we derive the two equivalent expressions for $\mut_{|\coreset}$ and $\Sigmat_{|\coreset}$ given in \eqref{eq:theta_posterior_item} and \eqref{eq:theta_posterior_bucket}.

Recall that $\bar{s}_{\coreset_{1:B}} = (\bar{s}_{\coreset_1}, \ldots, \bar{s}_{\coreset_B})$, where $\bar{s}_{\coreset_b} = \frac{1}{n_b}\sum_{i \in \coreset_b} s_i$ and $\coreset_b$ denotes the subset of $n_b$ items in $\coreset$ that are in bucket $b$.
As $p(\bar{s}_{\coreset_b} \cond \theta_b)=\mathcal{N}\left(\theta_b, \frac{\sigma^2 N_b}{n_b}\right)$, we have:
\begin{align*}
p(s_{\coreset_b} \cond \theta_b) &\propto \exp\left( -\frac{1}{2 \sigma^2 N_b} \sum_{i \in \coreset_b} (s_i - \theta_b)^2 \right)\\
& =\exp\left( -\frac{1}{2 \sigma^2 N_b} \sum_{i \in \coreset_b}((s_i - \bar{s}_{\coreset_b}) + (\bar{s}_{\coreset_b} - \theta_b))^2 \right)\\
& =\exp\left( -\frac{1}{2 \sigma^2 N_b}  \left(\sum_{i \in \coreset_b} (s_i - \bar{s}_{\coreset_b})^2 + n_b(\bar{s}_{\coreset_b} - \theta_b)^2 \right)\right)\\
& =\exp\left( -\frac{1}{2 \sigma^2 N_b}  \sum_{i \in \coreset_b} (s_i - \bar{s}_{\coreset_b})^2 \right) \exp\left( -\frac{n_b}{2 \sigma^2 N_b}  (\bar{s}_{\coreset_b} - \theta_b)^2 \right)\\
& \propto \exp\left( -\frac{n_b}{2 \sigma^2 N_b}  (\bar{s}_{\coreset_b} - \theta_b)^2 \right)\\
& \propto  p(\bar{s}_{\coreset_b} \cond\theta_b),
\end{align*}
where the proportionality is with respect to $\theta_b$. This results into:
\begin{equation*}
    p(\theta\cond s_\coreset) \propto p(\theta) p(s_\coreset\cond \theta) = p(\theta) \prod_{b=1;\, n_b>0}^B p(s_{\coreset_b}\cond \theta_b)\propto p(\theta) \prod_{b=1;\, n_b>0}^B p(\bar s_{\coreset_b}\cond \theta_b) =  p(\theta) p(\bar{s}_{\coreset_{1: B}}\cond \theta) \propto p(\theta\cond \bar{s}_{\coreset_{1: B}}),
\end{equation*}
which establishes that $p(\theta\cond s_\coreset) = p(\theta\cond \bar{s}_{\coreset_{1: B}})$. 

Having shown this equivalence, we now derive the explicit form for $p(\theta\cond s_\coreset)=p(\theta\cond \bar{s}_{\coreset_{1: B}})$ through the Gaussian conditioning rule, which states that, for a joint distribution $p(x_1, x_2)$, the conditional distribution $p(x_1 \cond x_2)$ has mean $\mu_1 + \Sigma_{12}\Sigma_{22}^{-1}(x_2 - \mu_2)$ and covariance $\Sigma_{11} - \Sigma_{12}\Sigma_{22}^{-1}\Sigma_{21}$.

Recall that $H_{\coreset}$ and $D_{\coreset}$ denote the submatrices of $H$ and $D$ formed by the rows corresponding to $\coreset$. The joint distribution of $(\theta,s_\coreset)$ can be written as: 
\begin{equation*}
\begin{bmatrix} \theta \\ s_\coreset \end{bmatrix} \sim \mathcal{N}\left( \begin{bmatrix} \mut \\ H_\coreset \mut \end{bmatrix}, \begin{bmatrix} \Sigmat & \Sigmat H_\coreset^\top \\ H_\coreset \Sigmat & H_\coreset \Sigmat H_\coreset^\top + D_\coreset \end{bmatrix} \right).
\end{equation*}
Applying the Gaussian conditioning rule, we obtain:
\begin{align*}
\mut_{|\coreset} &= \mut + \Sigmat H_\coreset^\top (H_\coreset \Sigmat H_\coreset^\top + D_\coreset)^{-1} (s_\coreset - H_\coreset\mut),\\
\Sigmat_{|\coreset} &= \Sigmat - \Sigmat H_\coreset^\top (H_\coreset \Sigmat H_\coreset^\top + D_\coreset)^{-1} H_\coreset \Sigmat,
\end{align*}
which corresponds to \eqref{eq:theta_posterior_item}.

Let $\Delta$ be a diagonal noise matrix with entries $\Delta_{b,b} = \sigma^2 N_b / n_b$. The joint distribution of $(\theta,\bar{s}_{\coreset_{1:B}})$ can be written as:
$$\begin{bmatrix} \theta \\ \bar{s}_{\coreset_{1:B}} \end{bmatrix} \sim \mathcal{N}\left( \begin{bmatrix} \mut \\ \mut \end{bmatrix}, \begin{bmatrix} \Sigmat & \Sigmat \\ \Sigmat & \Sigmat + \Delta \end{bmatrix} \right).$$ 
Applying the Gaussian conditioning rule, we obtain:
\begin{align*}
\mut_{|\coreset} &= \mut + \Sigmat (\Sigmat + \Delta)^{-1} (\bar{s}_{\coreset_{1:B}} - \mut),\\
\Sigmat_{|\coreset} &= \Sigmat - \Sigmat (\Sigmat + \Delta)^{-1} \Sigmat,
\end{align*}
which corresponds to \eqref{eq:theta_posterior_bucket}.

\subsection{Performance Estimator}\label{sec:app:cv_estimator}
In this section, we discuss the estimator $\hat{R}_{CV}$ (\eqref{eq:performance_estimate_control_variate}) that we use in the non-unique reference regime $B < N - N_0$. To simplify the exposition, we adopt an item-level, rather than bucket-level, description. Specifically, for an item $(x_i,y_i)$ belonging to bucket $b$, we use $\mu_{i|\coreset}$ to denote $\mathbb{E}[ \theta_b\cond s_\coreset]$.
We can rewrite \eqref{eq:performance_estimate_control_variate} at the item-level as:
\begin{equation*}
    \hat{R}_{\text{CV}} = \frac{1}{n}\sum_{i\in \coreset} \left(s_i - \mu_{i|\coreset}\right) + \frac{1}{N}\sum_{j=1}^N \mu_{j|\coreset}.
\end{equation*}
This estimator can be seen as an in-sample proxy of the following control variate estimator that uses out-of-sample posteriors:
\begin{equation*}
    \hat{R}_{\text{CVO}} = \frac{1}{n}\sum_{i\in \coreset} \left(s_i - \mu_{i|\coreset\setminus\{i\}}+ \frac{1}{N}\sum_{j=1}^N \mu_{j|\coreset\setminus \{i\}} \right).
\end{equation*}
Note that the in-sample proxy estimator avoids the computational bottleneck of recomputing the posterior for every held-out item.

$\hat{R}_{\text{CVO}}$ is unbiased with respect to the random selection of the coreset as shown below.
Let $\coreset = (i_1, \dots, i_n)$ be constructed by sampling items uniformly at random with replacement, such that $i_k\sim U(1, \dots, N)$.
$\hat{R}_{\text{CVO}}$ can be expressed as: 
\begin{equation*}
    \hat{R}_{\text{CVO}} = \frac{1}{n}\sum_{k=1}^n \left( s_{i_k} - \mu_{i_k|\coreset\setminus\{i_k\}} + \frac{1}{N}\sum_{j=1}^N \mu_{j|\coreset\setminus\{i_k\}}\right).
\end{equation*}
Taking the unconditional expectation with respect to the random selection of the coreset $\coreset$ yields:
\begin{align}\label{eq:intermediate_unbiased_cv_estimator}
     \mathbb{E}[\hat{R}_{\text{CVO}}] &= \frac{1}{n}\sum_{k=1}^n \mathbb{E}[\sr_{i_k}] - \frac{1}{n}\sum_{k=1}^n\left( \mathbb{E}\left[\mu_{i_k|\coreset\setminus\{i_k\}}\right]-\frac{1}{N}\sum_{j=1}^N\mathbb{E}\left[\mu_{j|\coreset\setminus\{i_k\}}\right]\right).
\end{align}
We first focus on the term $\mathbb{E}\left[\mu_{i_k|\coreset\setminus\{i_k\}}\right]$. Using the Law of Total Expectation, we have:
\begin{equation*}
    \mathbb{E}\left[\mu_{i_k|\coreset\setminus\{i_k\}}\right] = \mathbb{E}_{\coreset\setminus\{i_k\}}\left[\mathbb{E}\left[\mu_{i_k|\coreset\setminus\{i_k\}}|\coreset\setminus\{i_k\}\right]\right].
\end{equation*}
Because $i_k$ is drawn independently from the remaining coreset items $\coreset\setminus\{i_k\}$, the inner expectation over the uniform draw of $i_k$ is exactly the population mean:
\begin{equation*}
    \mathbb{E}\left[\mu_{i_k|\coreset\setminus\{i_k\}}\right] = \mathbb{E}_{\coreset\setminus\{i_k\}}\left[\frac{1}{N}\sum_{j=1}^N\mu_{j|\coreset\setminus\{i_k\}}\right] = \frac{1}{N} \sum_{j=1}^N\mathbb{E}\left[\mu_{j|\coreset\setminus\{i_k\}}\right].
\end{equation*}
Plugging this into \eqref{eq:intermediate_unbiased_cv_estimator}, the second term cancels out, resulting in:
\begin{equation*}
     \mathbb{E}[\hat{R}_{\text{CVO}}] = \frac{1}{n}\sum_{k=1}^n \mathbb{E}[\sr_{i_k}] =\frac{1}{n}\sum_{k=1}^n \mathbb{E}_{\coreset\setminus \{i_k\}}[\mathbb{E}[\sr_{i_k}\cond \coreset\setminus \{i_k\}]] =\frac{1}{n}\sum_{k=1}^n \mathbb{E}_{\coreset\setminus \{i_k\}}\left[\frac{1}{N}\sum_{j=1}^N \sr_j\right] = \frac{1}{N}\sum_{j=1}^N \sr_j= R.
\end{equation*}
\textit{Remark}: In practice, the coreset is sampled without replacement. This breaks strict independence, introducing a weak dependence between $i_k$ and $\coreset\setminus\{i_k\}$ and yielding a residual bias. However, this value is negligible when $N \gg 1$.

\subsection{Stopping Criterion}\label{app:stopping_criterion}
Recall that, to monitor posterior uncertainty about the random variable $R$, we consider the width of the 95\% credible interval for $R$, which is governed by $\operatorname{Var}(R\cond s_\coreset)$. Crucially, this quantity is not tied to a particular choice of point estimator. This distinction is important in the non-unique reference regime, $B<N-N_0$, where we use the control-variate estimator $\hat{R}_{\text{CV}}$ in \eqref{eq:performance_estimate_control_variate}. 
Our stopping criterion deliberately separates these two roles: the stability condition $(i)$ is applied to the reported estimator $\hat{R}_{\text{CV}}$, whereas the credible-interval condition $(ii)$ is applied to the target quantity $R$.

A natural alternative would be to define a random variable ${R}_{\text{CV}} = \frac{1}{n} \sum_{i \in \coreset} s_i +  \sum_{b=1}^B \left(\frac{N_b}{N}- \frac{n_b}{n}\right)\,\theta_b$ and use $\operatorname{Var}(R_{\text{CV}}\cond s_\coreset)$ in the stopping criterion. However, this variance is not reliable as a measure of uncertainty about $R$ since it vanishes whenever $N_b/N=n_b/n$, which occurs, e.g., when there is only one bucket. Another possible estimator-specific diagnostic is the posterior MSE, $\mathbb{E}[(R-\hat R_{\text{CV}})^2\cond s_C]=\operatorname{Var}(R\cond s_\coreset)+(\hat R_{\text{Bayes}}-\hat R_{\text{CV}})^2$. This is a valid posterior risk criterion under squared-error loss: the first term captures residual uncertainty about $R$, while the second captures the conditional bias of ${R}_{\text{CV}}$ relative to the posterior mean. 
We use  $\operatorname{Var}(R\cond s_\coreset)$ as the default uncertainty measure because it preserves a direct credible-interval interpretation for the benchmark performance and avoids conflating posterior uncertainty with the deliberate correction introduced by the control-variate estimator. The estimator-specific discrepancy is instead monitored through the stability condition $(i)$.

\subsection{Hyperparameter Optimization}\label{sec:app:hyperparameter_optimization}
\paragraph{Single-target Setting}
We tune the hyperparameters $\alpha$ and $\beta$, which govern the prior covariance $\Sigmat$, by minimizing the negative log-marginal likelihood $\mathcal{L}(\alpha, \beta)$ of the coreset scores. 
As established in Appendix \ref{sec:app:equivalent_bucket_matrix_H}, the conditional likelihoods $p(s_\coreset \cond \theta)$ and $p(\bar{s}_{\coreset_{1:B}} \cond \theta)$ are proportional up to a constant independent of $\theta$. As a result, their negative log-marginal likelihoods differ only by a constant independent of $\alpha$ and $\beta$. Therefore, to evaluate the objective efficiently, we define $\mathcal{L}(\alpha, \beta)$ as:
\begin{equation}\label{eq:NLML_objective}
\mathcal{L}(\alpha, \beta) = 
\begin{cases}
\frac{1}{2} (s_\coreset - H_\coreset \mut)^\top (H_\coreset \Sigmat H_\coreset^\top + D_\coreset)^{-1} (s_\coreset - H_\coreset \mut) + \frac{1}{2} \log |H_\coreset \Sigmat H_\coreset^\top + D_\coreset| & \text{if } n < B\\
\frac{1}{2} (\bar{s}_{\coreset_{1:B}} - \mut)^\top (\Sigmat + \Delta)^{-1} (\bar{s}_{\coreset_{1:B}} - \mut) + \frac{1}{2} \log |\Sigmat + \Delta| & \text{if } n \geq B
\end{cases}
\end{equation}
For additional efficiency and to maintain numerical stability, we employ a Cholesky decomposition of the respective covariance matrix. 
We perform this minimization via an L-BFGS-B optimizer \citep{byrd1995limited}, constraining the parameters with a small lower bound of $10^{-5}$ to ensure the covariance matrix remains positive definite. 

\paragraph{Multi-target Setting} 
We tune the hyperparameters $\alpha_l$ and $\beta_l$ which govern the prior covariance $\Sigma^{\theta,l}$, as well as the model-specific weights $\{w^t_l\}_{t=1}^{K_t}$ for each $l=1,\ldots,L$. Additionally, we learn the noise variance $\sigma^2$ instead of keeping it fixed as for the single-target setting. This turns minimizing the negative log-marginal likelihood into a highly non-convex, joint optimization problem. Consequently, instead of L-BFGS-B, we employ AdamW \citep{loshchilov2017decoupled}. AdamW leverages momentum to successfully navigate complex loss surfaces and provides adaptive, per-parameter learning rates, which are essential for handling parameters that operate on fundamentally different scales. Because the hyperparameters naturally stabilize as the coreset grows, we deploy an adaptive computational schedule that executes 500 gradient steps during the initial active learning iterations and progressively decreases this count in later iterations to significantly reduce computational overhead. 
To ensure numerical stability throughout  the training process, we clip the values of $\alpha_l$, $\beta_l$, and $\sigma^2$ after each update, enforcing a lower bound of $10^{-4}$.

\section{\BayesAME \acro{Single-target} Variants}\label{app:bayesame_variants}
\subsection{Batch Selection Strategies}\label{sec:app:alternative_active_selection} 
We extend the approach from the main text by introducing selection strategies that identify an optimal batch of items $A$, rather than a single item. We focus on the nearly unique reference regime $B\geq N - N_0$. To simplify the presentation, we denote $\Sigma_{|\coreset} = H \Sigmat_{|\coreset} H^\top + D$, which in the case $B=N$ simplifies to $\Sigma_{|\coreset} = \Sigmat_{|\coreset} + \sigma^2 I_N$.

The extension of the expected information gain \eqref{eq:information_gain} from one item to a batch of items $A$ is given by:
\begin{align*}
    \alpha_{\text{IG}}(A) &= \mathcal{H}(R\cond s_\coreset) - \mathbb{E}_{s_A}[\mathcal{H}(R\cond s_\coreset, s_A)]\\
    &= -\frac{1}{2}\log\left(1-\frac{1}{N^2\operatorname{Var}(R\cond s_\coreset)}( \Sigma_{A, U| \coreset} \mathbf{1}_{\vert U\vert})^\top\Sigma_{A, A| \coreset}^{-1}( \Sigma_{A, U| \coreset} \mathbf{1}_{\vert U\vert})\right),
\end{align*}
where $\Sigma_{A, U| \coreset}$ denotes the submatrix of $\Sigma_{|  \coreset}$ of columns and rows corresponding to $A$ and the unevaluated set $U = E\setminus\coreset$, respectively, and $\mathbf{1}_{\vert U\vert}$ is a $\vert U\vert$-dimensional vector of ones. 
A fundamental property of Gaussian conditioning where the noise is not a function of the score value implies that $\Sigma_{|\coreset}$ depends only on the score indexes rather than their actual values. This means that batch selection strategies simplify to a purely combinatorial problem of identifying an optimal batch of indices. Once this optimal batch is identified, the specific sequence in which these items are evaluated is arbitrary.

Because the expected information gain is strictly monotonically increasing with respect to the variance reduction, maximizing $\alpha_{\text{IG}}(A)$ is equivalent to maximizing the variance reduction term:
\begin{equation}\label{eq:variance_reduction_batch}
    \beta(A) = \frac{1}{N^2}( \Sigma_{A, U| \coreset} \mathbf{1}_{\vert U\vert})^\top\Sigma_{A, A | \coreset}^{-1}( \Sigma_{A, U| \coreset} \mathbf{1}_{\vert U\vert}).
\end{equation}
Therefore, for a batch $A$ of specified size $\vert A\vert=k$, we obtain the following optimization problem:
\begin{equation}\label{eq:combinatorial_optimization_problem} A^\star = \operatorname{argmax}_{A\subset U, \vert A\vert = k} \beta(A).\end{equation}
\paragraph{Batch Size $\vert A\vert=2$} When $\vert A\vert=2$, we can derive a closed-form expression of $\beta(A=\{i,j\})$ that can be efficiently vectorized:
\begin{align*}
    \beta(\{i,j\}) &= \frac{1}{N^2} \begin{pmatrix}
        T_i &
        T_j
    \end{pmatrix} 
    \begin{pmatrix}
       \Sigma_{i,i|\coreset} & \Sigma_{i,j|\coreset}\\
       \Sigma_{i,j|\coreset} & \Sigma_{j,j|\coreset}
    \end{pmatrix}^{-1}\begin{pmatrix}
        T_i\\
        T_j
    \end{pmatrix}
    =\frac{1}{N^2}\frac{T_i^2\Sigma_{j,j| \coreset} + T_j^2\Sigma_{i,i| \coreset} - 2 T_iT_j\Sigma_{i,j| \coreset}}{\Sigma_{i,i| \coreset}\Sigma_{j,j| \coreset} - \Sigma_{i,j| \coreset}^2},
\end{align*}
where $T_i = \sum_{k\in U} \Sigma_{i,k| \coreset}$ and similarly for $T_j$.

\paragraph{Batch Size $\vert A\vert>2$} 
As the batch size $\vert A\vert$ grows, an exhaustive search over all $\binom{\vert U\vert}{\vert A\vert}$ subsets requires $\mathcal{O}(\vert U\vert^{\vert A\vert} \vert A\vert^3)$ operations, which quickly becomes computationally infeasible even for moderate $\vert U\vert$. We address this by replacing the exhaustive search with an approximation algorithm that scales linearly with $\vert U\vert$.
Specifically, we use heuristic warm-start strategies to identify initial candidate batches, followed by a combination of local search and Sequential Monte Carlo (SMC). Coupling local search with SMC prevents the algorithm from getting stuck in local optima: we maintain a population of candidate batches, apply local search to each independently, and resample them according to their fitness $\beta(A)$. We use the following warm-start strategies:
\begin{itemize}
    \item \textbf{\textsc{LARS}}:
    Solve an $\ell_1$-penalized continuous relaxation of \eqref{eq:combinatorial_optimization_problem}. In particular,  for  ${w} \in \mathbb{R}^N$ define the following function:
    \begin{equation*}
    g({w})
    =
    \frac{1}{2}\,{w}^\top
    {\Sigma}_{| \coreset}\,{w}
    -
    {w}^\top {v},
  \end{equation*}
  where ${v} = \Sigma_{E, U| \coreset} \mathbf{1}_{\vert U\vert}$. Fix a subset $A$ and let ${w}_{A^c} = \mathbf{0}$. The gradient of $g$ with respect to ${w}_A$ is
  $$
    \nabla_{{w}_A} g
    =
    {\Sigma}_{A,A| \coreset}\,{w}_A
    -
    {v}_A,
  $$
  where ${v}_A = \Sigma_{A, U| \coreset} \mathbf{1}_{\vert U\vert}$. Setting the gradient to zero yields the optimizer ${w}_A^\star ={\Sigma}_{A,A| \coreset}^{-1}{v}_A$. Substituting ${w}_A^\star$ into $g$ gives
  \begin{equation*}
      g({w}_A^\star) = -\frac{1}{2}{v}_A^\top {\Sigma}_{A,A| \coreset}^{-1} {v}_A = - \frac{N^2}{2} \beta(A).
  \end{equation*}
  Therefore, maximizing $\beta(A)$ over all subsets $A$ of size $k$ is equivalent to solving the $\ell_0$-constrained quadratic program
  \begin{equation*} 
    \min_{{w}\in\mathbb{R}^N}
    \,
    \frac{1}{2}\,{w}^\top {\Sigma}_{| \coreset}{w}
    -
    {w}^\top{v}
    \quad
    \text{subject to}
    \quad
    \Vert{w}\Vert_0 \le k.
  \end{equation*}
  Since this problem is NP-hard, we instead consider the standard $\ell_1$ relaxation
  \begin{equation*}
    \min_{{w}\in\mathbb{R}^N}
    \,
    \frac{1}{2}\,{w}^\top \Sigma_{| \coreset}{w}
    -
    {w}^\top{v}+\lambda\Vert {w}\Vert_1, \quad \lambda >0.
  \end{equation*}
    To solve this, we apply the least angle regression (LARS) algorithm to efficiently extract an initial active set of $\vert A\vert$ items. 
    \item \textbf{\textsc{DPP}}: Sample multiple diverse initial batches using a $k$-determinantal point process ($k$-DPP) \citep{kulesza2011k-dpps} restricted to size $\vert A\vert$, using the posterior covariance matrix of the candidates as the likelihood kernel.
    \item \textbf{\textsc{Clust}}: Perform $K$-means clustering on the reference model score vectors $\phi_i$ to partition the candidate pool into $\vert A\vert$ distinct clusters. Generate multiple initial sets by uniformly sampling one item from each cluster.
\end{itemize}

\begin{figure}[h!]
\centering
\renewcommand{\arraystretch}{1.2} 

\begin{tabular}{c @{\hspace{3pt}} c @{} c}
% --- X-Axis Column Headers (Top) ---
& 
% Headers for the FIRST 2x3 PDF
\makebox[0.22\textwidth][c]{\scriptsize{Interpolation}} \makebox[0.22\textwidth][c]{\scriptsize{Extrapolation}} &
% Headers for the SECOND 2x3 PDF 
\makebox[0.22\textwidth][c]{\scriptsize{Interpolation}} \makebox[0.22\textwidth][c]{\scriptsize{Extrapolation}} \\

\begin{tabular}{@{}c@{}}
    \rotatebox{90}{\scriptsize{50\%}} \\[1.5cm] %
    \rotatebox{90}{\scriptsize{90\%}}
\end{tabular} &
% First 2x3 PDF
\includegraphics[width=0.48\textwidth, valign=c]{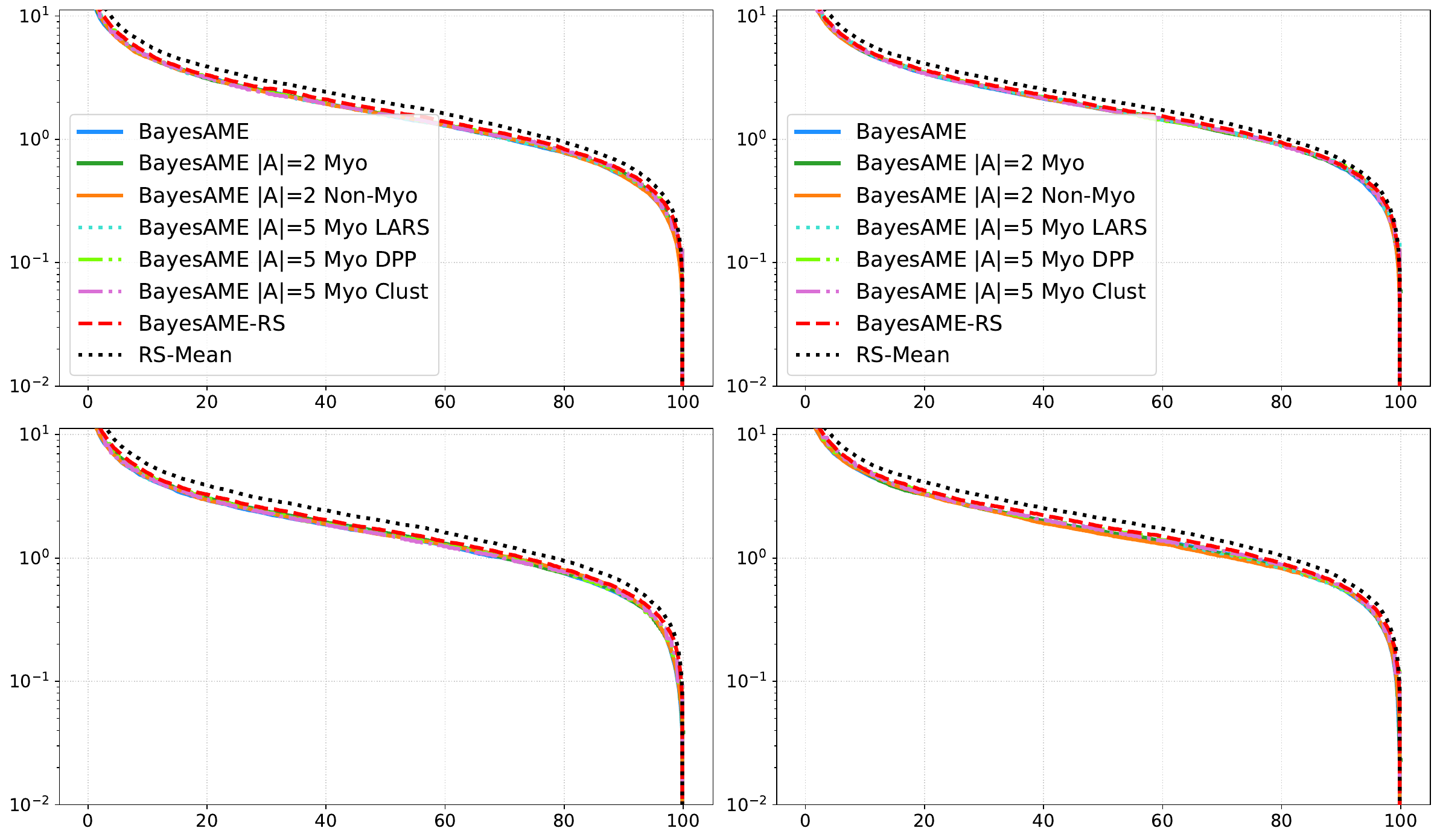} &
% Second 2x3 PDF
\includegraphics[width=0.48\textwidth, valign=c]{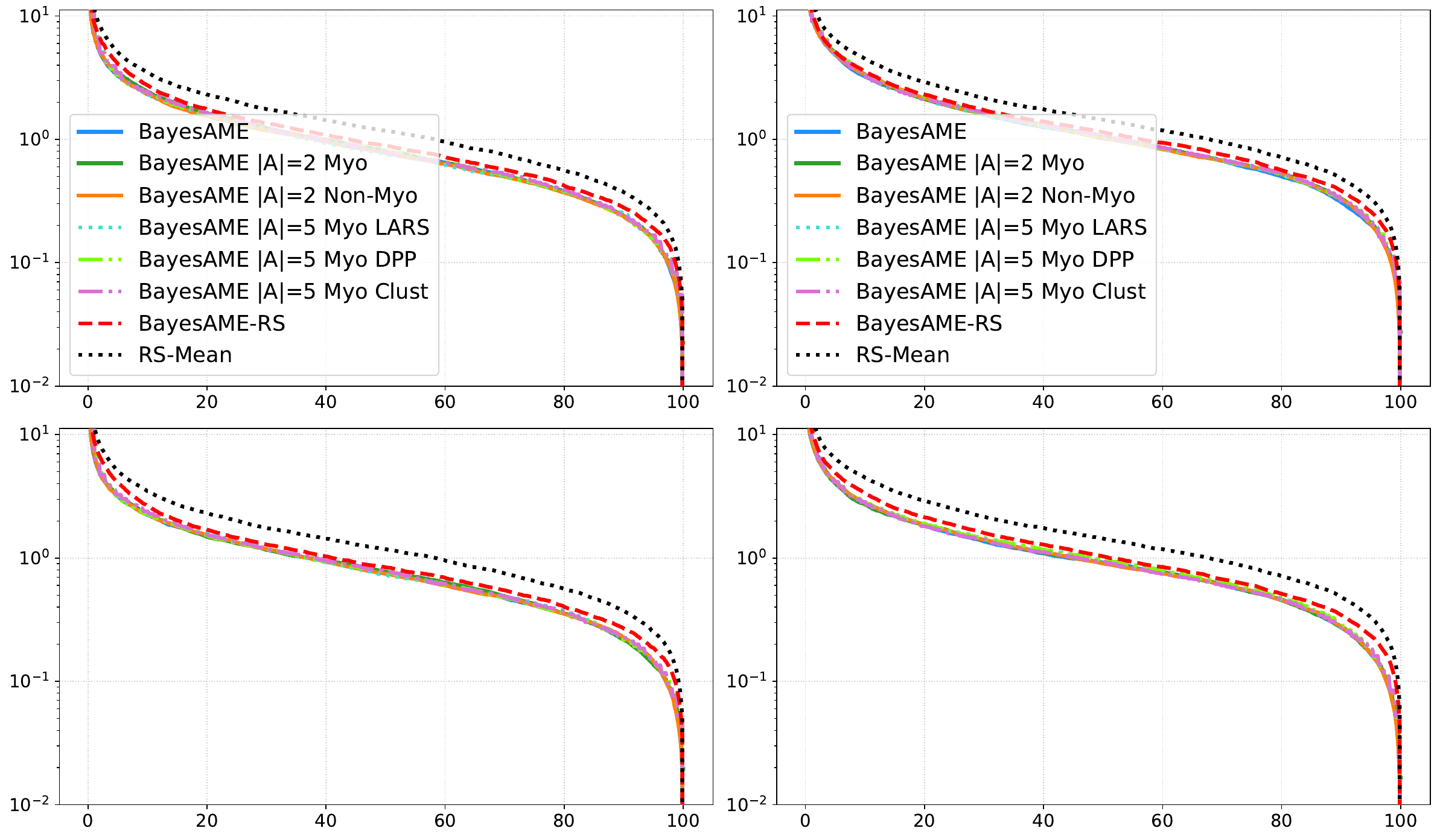} \\
\end{tabular}
\begin{tabular}{c @{\hspace{3pt}} c @{} c}
% --- X-Axis Column Headers (Top) ---
& 
% Headers for the FIRST 2x3 PDF
\makebox[0.22\textwidth][c]{\scriptsize{Interpolation}} \makebox[0.22\textwidth][c]{\scriptsize{Extrapolation}} &
% Headers for the SECOND 2x3 PDF 
\makebox[0.22\textwidth][c]{\scriptsize{Interpolation}} \makebox[0.22\textwidth][c]{\scriptsize{Extrapolation}} \\

\begin{tabular}{@{}c@{}}
    \rotatebox{90}{\scriptsize{50\%}} \\[1.5cm] %
    \rotatebox{90}{\scriptsize{90\%}}
\end{tabular} &
% First 2x3 PDF
\includegraphics[width=0.48\textwidth, valign=c]{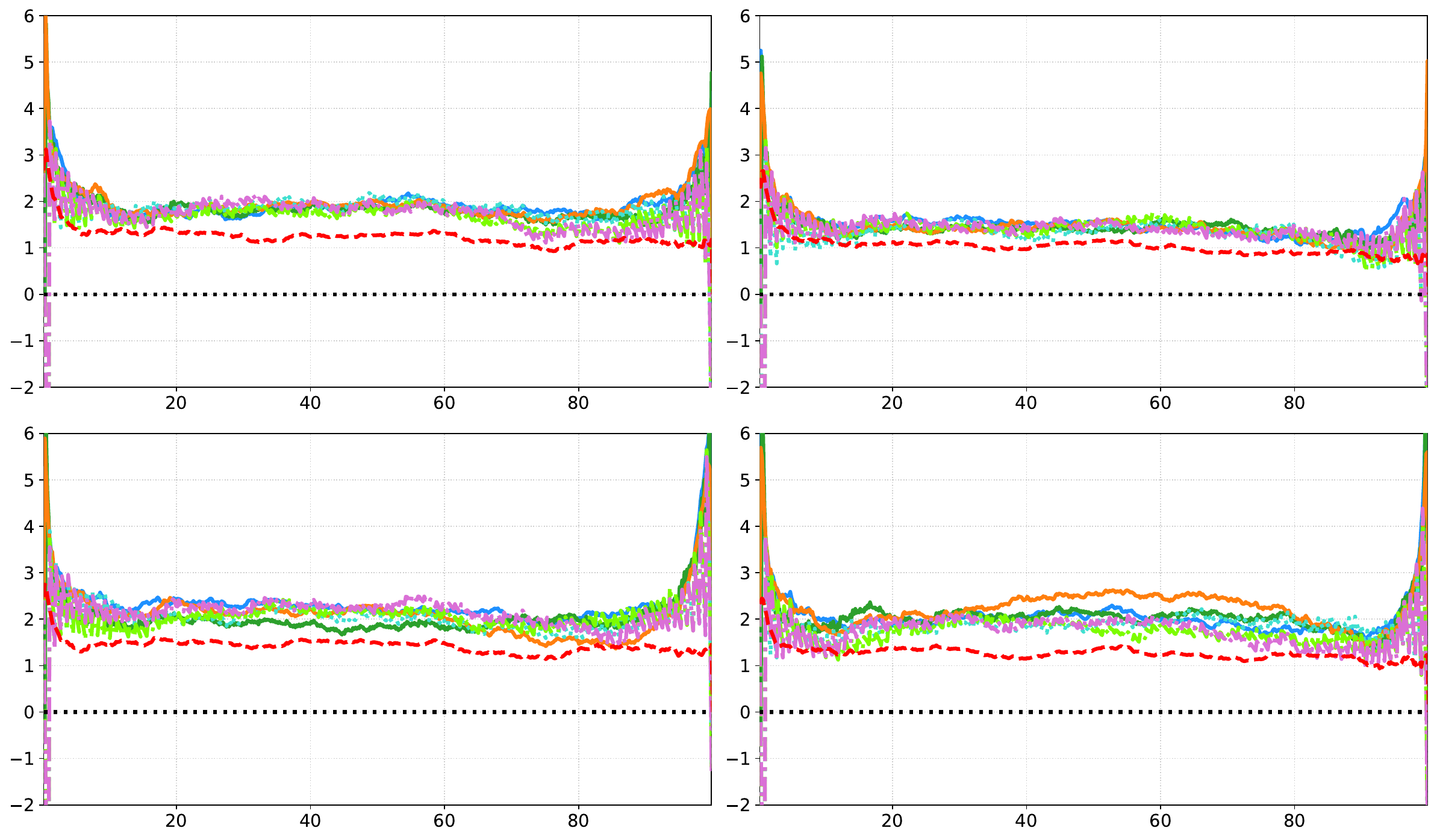} &
% Second 2x3 PDF
\includegraphics[width=0.48\textwidth, valign=c]{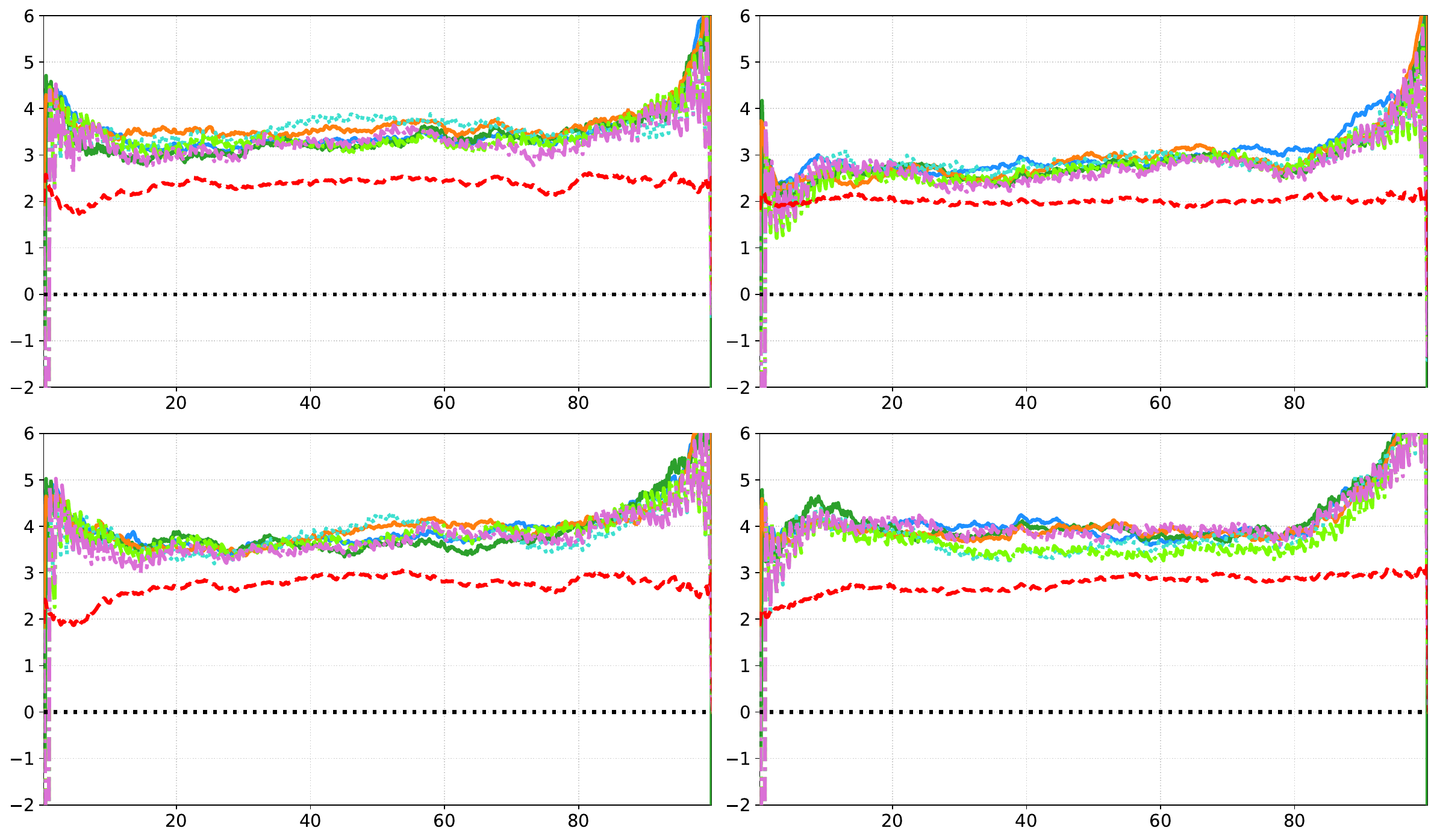} \\
\end{tabular}
\renewcommand{\arraystretch}{1.2} 

\begin{tabular}{c @{\hspace{3pt}} c @{} c}
% --- X-Axis Column Headers (Top) ---
& 
% Headers for the FIRST 2x3 PDF
\makebox[0.22\textwidth][c]{\scriptsize{Interpolation}} \makebox[0.22\textwidth][c]{\scriptsize{Extrapolation}} &
% Headers for the SECOND 2x3 PDF 
\makebox[0.22\textwidth][c]{\scriptsize{Interpolation}} \makebox[0.22\textwidth][c]{\scriptsize{Extrapolation}} \\

\begin{tabular}{@{}c@{}}
    \rotatebox{90}{\scriptsize{50\%}} \\[1.5cm] %
    \rotatebox{90}{\scriptsize{90\%}}
\end{tabular} &
% First 2x3 PDF
\includegraphics[width=0.48\textwidth, valign=c]{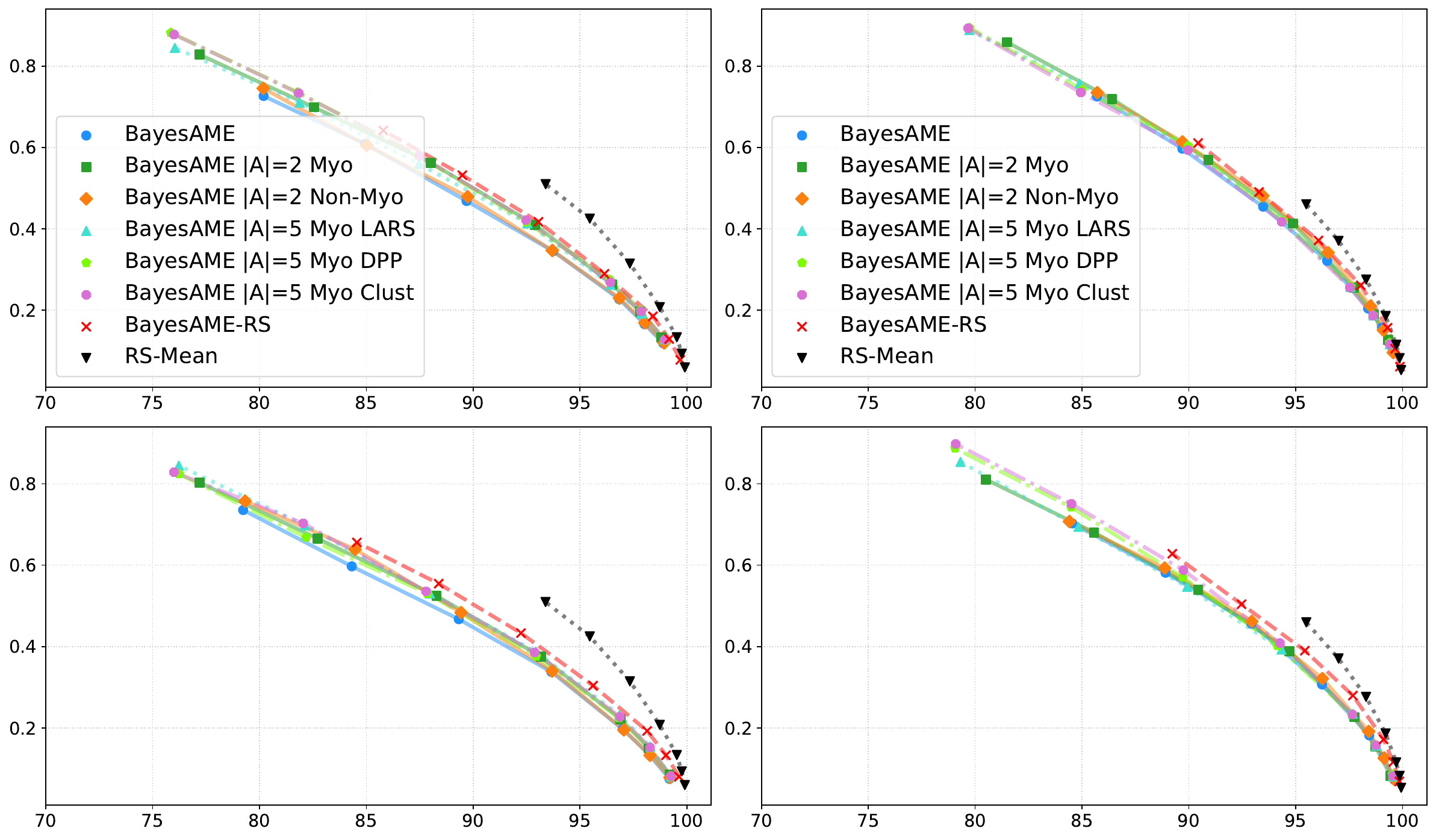} &
% Second 2x3 PDF
\includegraphics[width=0.48\textwidth, valign=c]{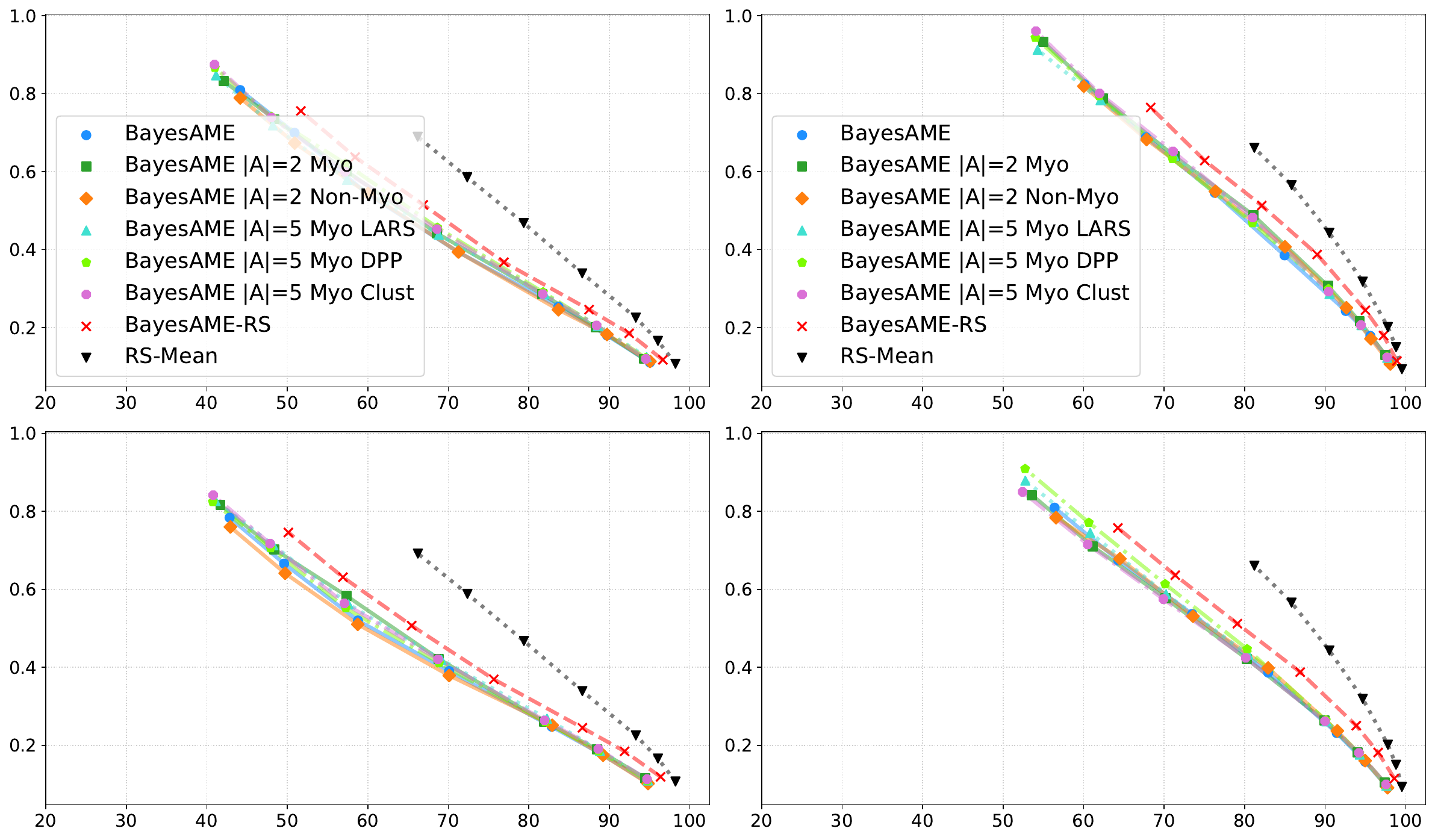} \\
\end{tabular}
\caption{\textbf{Single-target Setting.} \textbf{GPQA} with binary scores (left two columns) and continuous scores (right two columns). \RMSELog (top two rows) and \RMSEGain (central two rows), and \CAT (bottom two rows) across 50\%, and 90\% of reference models (percentages with unique reference scores $B=N$). 
}
\label{fig:GPQA-SAMPLING-SCORING-RMSELog-GAIN-CAT-ACTIVE-LEARNING}
\end{figure}
Crucially, during the local search phase, evaluating a neighbor where only a single item in the candidate batch is swapped can be computed highly efficiently using rank-2 updates. Suppose we wish to recompute the inverse in \eqref{eq:variance_reduction_batch} after a swap. Let $M_{A} = \Sigma_{A,A|\coreset}$ be the original $\vert A\vert \times \vert A\vert$ matrix and $M_{A}' = \Sigma_{\tilde A,\tilde A|\coreset}$ be the updated matrix. If the $m$-th item is swapped from $A$ to $\tilde{A}$, then $M_{A}$ and $M_{A}'$ differ only by their $m$-th row and column. Let $e_m\in\mathbb{R}^{\vert A\vert}$ denote the $m$-th standard basis vector (a vector of zeros with a $1$ at the $m$-th coordinate). Let $c_m$ be the $m$-th column of $M_{A}$ and $\tilde{c}_m$ be the corresponding column in $M_{A}'$. Defining the difference vector $d = \tilde c_m - c_m$, we can express the update as:
$$
M_{A}' = M_{A} + d e_m^\top + e_m d^\top - d_m e_m e_m^\top,
$$
where $d_m$ denotes the $m$-th coordinate of $d$. This can be factorized into $M_{A}' = M_{A} + U V^\top$, where
\begin{align*}
U = \begin{bmatrix} e_m & d - \frac{1}{2} d_m e_m \end{bmatrix}, \quad
V = \begin{bmatrix} d - \frac{1}{2} d_m e_m & e_m \end{bmatrix}.
\end{align*}
This rank-2 factorization allows us to apply the Woodbury matrix identity:
$$
(M_{A} + UV^\top)^{-1} = M_{A}^{-1} - M_{A}^{-1} U (I_2 + V^\top M_{A}^{-1} U)^{-1} V^\top M_{A}^{-1}.
$$
Because $V^\top M_{A}^{-1} U$ is only a $2\times 2$ matrix, evaluating this update is extremely fast. We can plug this inverted matrix directly back into \eqref{eq:variance_reduction_batch}, reducing the computational complexity of evaluating a neighbor to just $\mathcal{O}(\vert A\vert^2)$ instead of $\mathcal{O}(\vert A\vert^3)$.

Once the optimal batch $A^\star$ is found, we can proceed in two different ways: (i) \textbf{\textsc{Myo}}: We evaluate the target model on all items in $A^\star$ simultaneously. Batch evaluation allows us to maximize throughput when multiple evaluations can be processed in parallel; or (ii) \textbf{\textsc{Non-Myo}}: We select a single item from it at random. Because this item is selected based on an optimal future horizon rather than just the immediate next step, this renders the active selection non-myopic.  

\begin{table*}[t]
    \centering
    \resizebox{\textwidth}{!}{
    \begin{tabular}{ll ccccc}
        \toprule
        \multicolumn{2}{l}{\textbf{\textsc{Benchmarks}}} & \textbf{\BayesAME} & \textbf{\RBayesAME} & \textbf{\BayesAME |A|=2 \textsc{Myo}} & \textbf{\BayesAME |A|=2  \textsc{Non-Myo}} & \textbf{\BayesAIPW} \\
        
        \midrule
        \multicolumn{7}{l}{\textit{\textsc{Binary}}} \\
        \midrule
        \textsc{GPQA}          & Interpolation & \underline{61.54} & 66.38 & 62.08& \textbf{61.51} & 67.21\\
                               & Extrapolation & \textbf{68.91}& 72.65 & \underline{69.43} & 69.50 & 73.41 \\ \addlinespace
        \textsc{MMLU-Pro}      & Interpolation & \textbf{30.18}& \textbf{30.18} & -&  -& \underline{30.68} \\
                               & Extrapolation & \textbf{37.20} & \textbf{37.20}& - & - &\underline{37.69}\\ \addlinespace
        \textsc{BBH}           & Interpolation & {31.70} & 36.35 & \textbf{30.51} & \underline{30.98} & 35.54\\
                               & Extrapolation & \textbf{38.35} & 42.02& \underline{38.47} & 38.66 & 41.34  \\ \addlinespace
        \textsc{ARC-Challenge} & Interpolation & \textbf{36.53}& \textbf{36.53} & -& - & \underline{36.68}\\
                               & Extrapolation & \textbf{36.56}& \textbf{36.56}& - & - & \underline{38.03} \\ \addlinespace
        \textsc{MuSR}          & Interpolation & \textbf{47.86} & \textbf{47.86} & -& - & \underline{49.18}\\
                               & Extrapolation & \textbf{54.19}& \textbf{54.19}& - & - & \underline{56.72}\\ \addlinespace
        \textsc{IFEval}        & Interpolation & \textbf{49.53} & 56.14& \underline{49.99}& 50.00 & 56.71  \\
                               & Extrapolation & \textbf{58.08} & \textbf{58.08} & - & - & \underline{60.52} \\
        
        \midrule
        \multicolumn{7}{l}{\textit{\textsc{Continuous}}} \\
        \midrule
        \textsc{GPQA}                & Interpolation & \underline{30.58} & 34.42& 30.84 &\textbf{29.88} &35.21\\
                                     & Extrapolation & \textbf{40.40} & 40.75& 41.17 & \underline{40.76}& 46.77\\ \addlinespace
        \textsc{MMLU-Pro}            & Interpolation & \textbf{13.11}&17.10 & \underline{13.14} & 13.32& 17.19\\
                                     & Extrapolation & {26.82} & 31.10 & \textbf{26.24} & \underline{26.60} & 31.33 \\ \addlinespace
        \textsc{BBH}                 & Interpolation & \underline{13.04}& 17.54& \textbf{12.81}& 13.13& 18.01 \\
                                     & Extrapolation & \textbf{29.80} & 35.57 & 30.62 & \underline{30.49} & 33.64\\ \addlinespace
        \textsc{ARC-Challenge}       & Interpolation & \textbf{26.04} &26.47 & - & - & \underline{26.19}  \\
                                     & Extrapolation & \underline{33.16} & \underline{33.16} & - & - & \textbf{33.05} \\ \addlinespace
        \textsc{MuSR}                & Interpolation  & \underline{27.83}& 33.42& 27.94& \textbf{27.43}& 34.34\\
                                     & Extrapolation& {36.96}& 42.56& \textbf{35.70} & \underline{36.57} & 43.69  \\ \addlinespace
        \textsc{Natural QA Openbook} & Interpolation & \textbf{28.26} & 29.68 & \underline{28.41} & 28.89 & 29.12\\
                                     & Extrapolation & {20.83}& 28.36 & \textbf{20.43} & \underline{20.79} & 27.70 \\
        \bottomrule
    \end{tabular}
    }
    \caption{\EWIRMSE
    for 50\% reference models. A dash - denotes that we are in a non-unique reference regime. Therefore, selection defaults to random and the entry is omitted.}
    \label{table:BudgetIntegratedRSME_Active_learning}
\end{table*}
Figure \ref{fig:GPQA-SAMPLING-SCORING-RMSELog-GAIN-CAT-ACTIVE-LEARNING} presents the \RMSELog, \RMSEGain, and \CAT for GPQA with binary (left two columns) and continuous (right two columns) scores.
We evaluate the selection strategies outlined above using batch sizes of $\vert A\vert=2$ and $\vert A\vert=5$. We observe that the overall performance of these batch approaches is similar to that of the single-item selection strategy.

Table \ref{table:BudgetIntegratedRSME_Active_learning} compares the \EWIRMSE achieved by \BayesAME with single-candidate selection, random selection, bath selection with  $\vert A\vert=2$ using both myopic (\textsc{Myo}) and non-myopic (\textsc{Non-Myo}) criteria, and \BayesAIPW across the different benchmarks. 
Consistent with the results of Figure \ref{fig:GPQA-SAMPLING-SCORING-RMSELog-GAIN-CAT-ACTIVE-LEARNING}, \BayesAME with batch size $\vert A \vert = 2$ exhibits performance similar to that of myopic single-candidate selection under both the myopic and non-myopic variants.

\subsection{Metadata-Weighted Prior Distribution} \label{sec:app:alternative_weighted_prior}
When rich metadata (such as, model size, training data, or hyperparameter configurations) is available for both the reference and target models, we can formulate a weighted prior distribution. 
Specifically, we define a weight vector $w = (w_1, \dots, w_{K_r})$, where $w_k \ge 0$ captures the prior positive correlation between reference model $k$ and the target model, with $\sum_{k=1}^{K_r} w_k=1$. We then consider the prior distribution $p(\theta) = \mathcal{N}(\mut, \Sigmat)$ with
\begin{gather*}
\mut_b =\phi_b^\top w, \qquad \Sigmat_{b, b'} = \alpha \exp\left(-\frac{\beta}{K_r}(  \phi_b - \phi_{b'})^\top W (  \phi_b - \phi_{b'})\right),
\end{gather*}
where $W = \text{diag}(w)$. In practice, public leaderboards rarely provide sufficiently detailed metadata to calibrate these weights reliably, so we assign uniform weights across the reference models $w_k=1/K_r$.

\subsection{Logit-Normal Prior Distribution}\label{sec:app:logit}
While achieving strong empirical performance, a Gaussian prior distribution on $\theta$ is not the best choice for binary and bounded continuous scores from a modeling viewpoint, as it assigns positive probability mass outside the true support. Below, we describe more appropriate alternative formulations that we considered.

\paragraph{Case $s_i\in[0,1]$} We introduce a logit-normal model that respects the support of scores. Specifically, we map both the latent abilities and the observed scores to the unconstrained real line using the logit transformation. We define the latent variable vector $f= (f_1, \dots, f_B)$, where $f_b = \operatorname{logit}(\theta_b)=\ln\left(\frac{\theta_b}{1 - \theta_b}\right)$, and assign a Gaussian prior distribution on it. The scores are similarly transformed, and we assume a Gaussian conditional distribution $p(\operatorname{logit}(s_i)\cond f_b) = \mathcal{N}(f_b, N_b\sigma^2)$ for an item $i$ belonging to bucket $b$. The posterior distribution of $f=(f_1, \dots, f_B)$ conditioned on $s_{\coreset}$ is given by: 
$$
p(f\cond s_\coreset) \propto p(f) \prod _{i\in \coreset}p(\operatorname{logit}(s_i)\cond f).
$$
Under this specification, $p(f \cond s_{\coreset})$ is itself Gaussian and inference over $f$ remains fully tractable.
\paragraph{Case $s_i\in\{0,1\}$}  Similarly to above, consider the latent variable vector $f=(f_1, \dots, f_b)$, where $f_b = \operatorname{logit}(\theta_b)$, and assign a Gaussian prior distribution on it. In the case of binary responses, a more appropriate model would use a Bernoulli likelihood of the form
\begin{equation*}
p(s_i \cond f) = \sigma(f_b)^{s_i} (1 - \sigma(f_b))^{1-s_i},
\end{equation*}
for an item $i$ belonging to bucket $b$. Then, the posterior distribution of $f= (f_1, \dots, f_B)$ conditioned on $s_\coreset$ is given by:
$$
p(f\cond s_\coreset) \propto p(f) \prod _{i\in \coreset}p(s_i\cond f).
$$
In this case, we can resort to variational inference or a Laplace approximation \citep{rasmussen2006Gaussian} to compute the posterior.

\begin{figure}[t]
\centering
\renewcommand{\arraystretch}{1.2} 

\begin{tabular}{c @{\hspace{3pt}} c @{} c}
% --- X-Axis Column Headers (Top) ---
& 
% Headers for the FIRST 2x3 PDF
\makebox[0.22\textwidth][c]{\scriptsize{Interpolation}} \makebox[0.22\textwidth][c]{\scriptsize{Extrapolation}} &
% Headers for the SECOND 2x3 PDF 
\makebox[0.22\textwidth][c]{\scriptsize{Interpolation}} \makebox[0.22\textwidth][c]{\scriptsize{Extrapolation}} \\

\begin{tabular}{@{}c@{}}
    \rotatebox{90}{\scriptsize{50\%}} \\[1.5cm] %
    \rotatebox{90}{\scriptsize{90\%}}
\end{tabular} &
% First 2x3 PDF
\includegraphics[width=0.48\textwidth, valign=c]{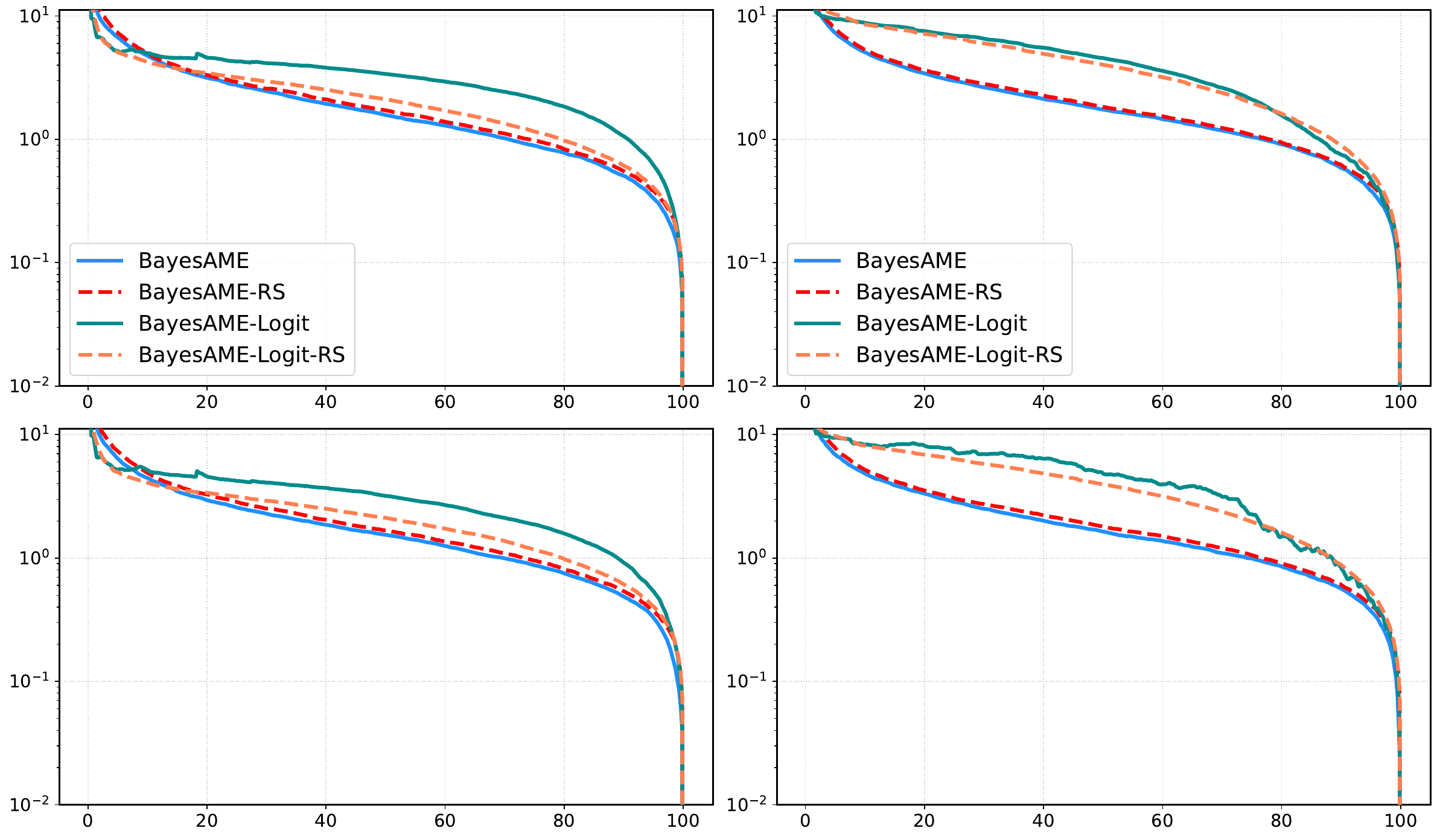} &
% Second 2x3 PDF
\includegraphics[width=0.48\textwidth, valign=c]{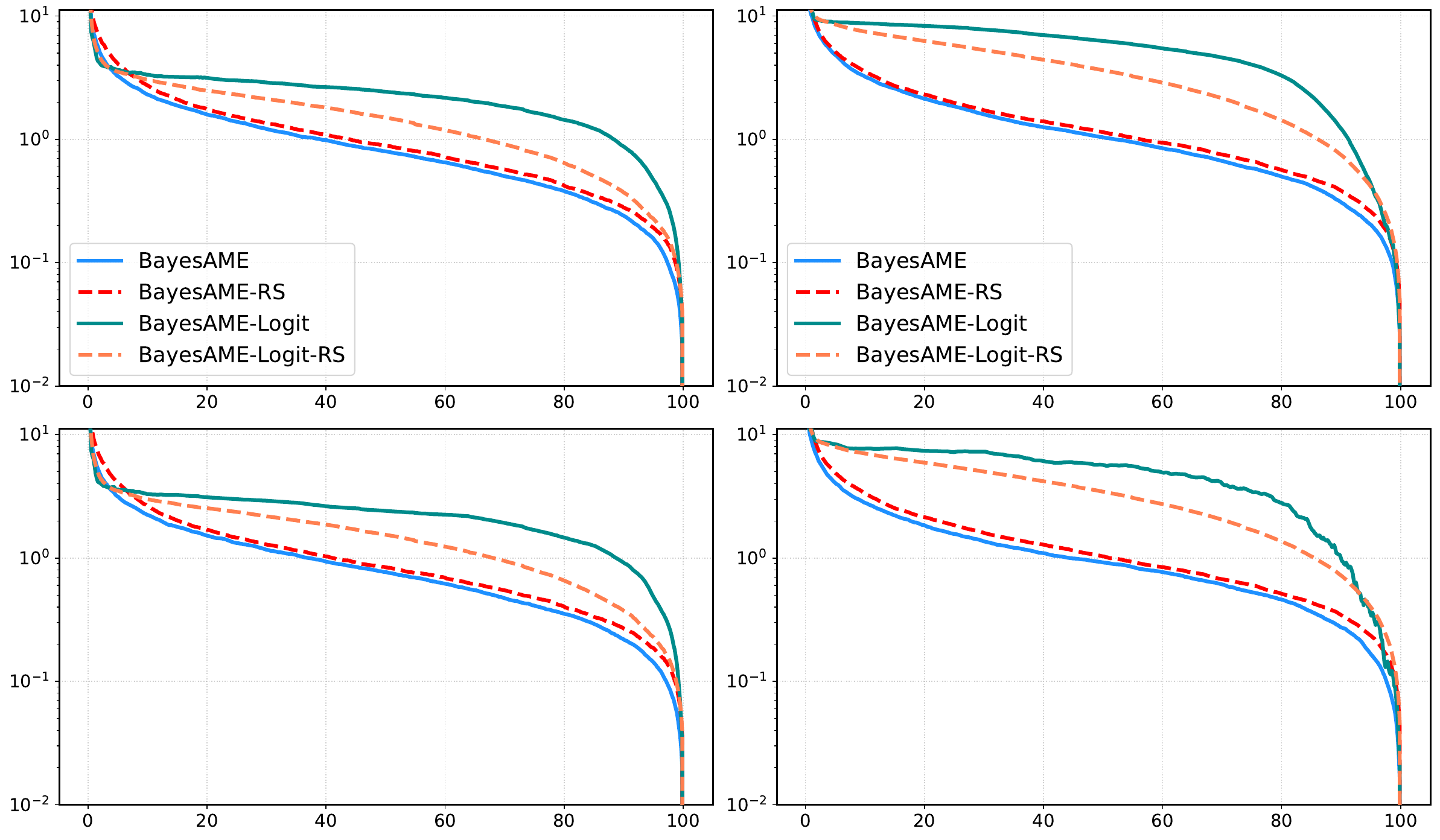} \\
\end{tabular} 
\caption{\textbf{Single-target Setting.} \textbf{GPQA} with binary scores (left two columns) and continuous scores (right two columns). \RMSELog across 50\% and 90\% of reference models.}
\label{fig:logit_ablation}
\end{figure}

However, we are interested in the posterior of $\theta = \sigma(f)=({1 + e^{-f}})^{-1}$, conditioned on $s_\coreset$. The posterior mean can be computed as:
\begin{equation*}
    \mathbb{E}[\theta_b\cond s_\coreset] = \int \sigma(f_b) p(f\cond s_\coreset) \, \text{d} f,
\end{equation*}
which does not admit a closed-form solution because of the nonlinearity of $\sigma$. We consider three strategies for approximating this integral.
\begin{itemize}
    \item \textbf{Monte Carlo estimation.} The main drawback is that 
    obtaining accurate estimates for all $N$ items requires a large number of 
    samples, which is prohibitive inside the sequential active learning loop.
    \item \textbf{Other numerical integration methods.} We can approximate $\sigma(f_b)$ in the integral with a rescaled probit function that has the same slope at the origin \citep{mackay92evidence}, obtaining:
    \begin{equation*}
         \mathbb{E}[\theta_b\cond s_\coreset]\approx \sigma\left(\mathbb{E}[f_b \cond s_\coreset]{\sqrt{1+\pi\operatorname{Var}(f_b|s_\coreset)/8}}\right).
    \end{equation*}
    \item \textbf{Linear approximation.} Following \citet{osborne2012bayesian}, we linearize the logistic term around a reference point $f_{b, 0}$, that is, \begin{equation*}
        \sigma(f_b) = \sigma(f_{b, 0}) + \sigma(f_{b, 0})(1-\sigma(f_{b, 0})) (f_{b} - f_{b, 0})
    \end{equation*}
    Under this approximation, the posterior mean of $\theta$ admits a 
    closed-form expression.
    To select the reference point, we introduce an auxiliary multivariate Gaussian model in the original space and set $\sigma(f_{b,0})$ to be its posterior mean given $s_\coreset$, i.e., $\sigma(f_{b,0}) = \mathbb{E}[\theta_b\cond s_\coreset]$, clipping the value to $[0,1]$ if necessary to remain within the valid range. Although this auxiliary Gaussian model is not appropriate as it does not have the correct range, it is used solely to inform the choice of $f_{b,0}$. 
\end{itemize}
While these alternative formulations offer a more principled treatment of the  score support, they require approximate inference at every step and introduce substantial computational overhead. In practice, we found that these approximations resulted in worse performance. For example, Figure~\ref{fig:logit_ablation} compares \BayesAME and \RBayesAME with their corresponding versions using logit-normal models (\textsc{\BayesAME-Logit} and \textsc{\BayesAME-Logit-RS}) on GPQA for both binary and continuous scores, using the linear approximation described above to derive the posterior for $\theta$. For both settings and regardless of the selection strategy, \BayesAME and \RBayesAME outperform the logit-normal formulation.

\section{Experimental Details and Additional Results}

\subsection{Implementations Details}

\paragraph{\BayesAME}
For the single-target setting, we use Algorithm \ref{algorithm:BayesAME} with $\sigma^2 = 10^{-4}$ when $B\ge N-N_0$ and $\sigma^2 = 10^{-1}$ otherwise, $N_0=10$, $W=30$. As described in Appendix \ref{sec:app:hyperparameter_optimization}, optimization is performed using L-BFGS-B with $\alpha_0=\beta_0=1$, and $F=50$.

For the multi-target setting, we set $N_0 = 10$, and $W=30$. Of the $L$ covariances, half are squared exponential (\eqref{eq:gaussian_prior_from_reference}) and the rest are Matérn-3/2. As described in Appendix \ref{sec:app:hyperparameter_optimization}, optimization is performed using AdamW, with $\alpha_l$ and $\beta_l$ initialized to $1$, each $w_i^t$ to a sample from $\mathcal{N}(0, 0.01)$, and $\sigma^2$ to $0.1$ for binary scores and to $0.01$ for continuous scores. We use $F = 50$, and learning rate $\eta=0.01$ for binary scores and $\eta=0.001$ for continuous scores. 

We set the margin $N_0 = 10$ to safely handle regimes where reference scores are nearly unique across items. Empirically, we identified a specific failure mode for active selection: when almost all items have unique combinations of reference scores (effectively forming singleton buckets), but one single bucket contains a significantly larger number of items, the active selection biases the posterior estimation and degrades performance. Enforcing $N_0 = 10$ prevents the algorithm from relying on active selection in this skewed regime, yielding stable results across all scenarios considered in this work. While fixing $N_0 = 10$ serves as a safe value, because the total number of buckets and the exact distribution of items across them are fully known to the algorithm, a more sophisticated future approach could explicitly incorporate this distributional information to set more suitable,  benchmark-specific values. We conjecture that larger threshold values may be admissible when bucket sizes are strictly balanced, as the observed estimation bias appears to be driven by the presence of a single, severely unbalanced bucket.

When selecting items in batches, we scale the window size $W$ proportionally. Specifically, for a batch size $|B|$, we set $W=30/|B|$.

\paragraph{Baselines}  
For \BayesAIPW, \BayesRSL, and \SeqAPW, we adapt the implementations of \AIPW, \RSL, and \textsc{APW}, respectively, provided by \cite{zhang2025how} in \small{\url{https://github.com/socialfoundations/benchmark-prediction}}\normalsize. In  \BayesAIPW and \BayesRSL, we replace the linear ridge regression with its Bayesian counterpart as implemented in \texttt{scikit-learn} library using the default setting \citep{pedregosa2011scikit-learn}. For \ProEval, we use the implementation provided by \cite{huang2026proeval} in \small{\url{https://github.com/google-deepmind/proeval}}\normalsize. 
For IRT, we implement the binary and continuous approaches based on the methodology outlined in \citet{polo2024tinybenchmarks} and \citet{chen2019beta3}, respectively.

\subsection{Additional Benchmark Details}
For BBH, we only consider a subset of scenarios: temporal sequences, salient translation error detection, tracking shuffled objects (seven objects), geometric shapes, and reasoning about colored objects.
For MMLU-Pro, we also restrict our evaluation to a subset of subjects, namely business and law, resulting in a total of 1890 items.

\subsection{Additional Results Single Target Setting}\label{sec:app:additional_results_benchmarks}

\begin{figure}[!h]
\centering
\renewcommand{\arraystretch}{1.2} 

\begin{tabular}{c @{\hspace{3pt}} c @{} c}
% --- X-Axis Column Headers (Top) ---
& 
% Headers for the FIRST 2x3 PDF
\makebox[0.22\textwidth][c]{\scriptsize{Interpolation}} \makebox[0.22\textwidth][c]{\scriptsize{Extrapolation}} &
% Headers for the SECOND 2x3 PDF 
\makebox[0.22\textwidth][c]{\scriptsize{Interpolation}} \makebox[0.22\textwidth][c]{\scriptsize{Extrapolation}} \\

\begin{tabular}{@{}c@{}}
    \rotatebox{90}{\scriptsize{10\%}} \\[1.5cm] %
    \rotatebox{90}{\scriptsize{50\%}} \\[1.5cm] %
    \rotatebox{90}{\scriptsize{90\%}}
\end{tabular} &
% First 2x3 PDF
\includegraphics[width=0.48\textwidth, valign=c]{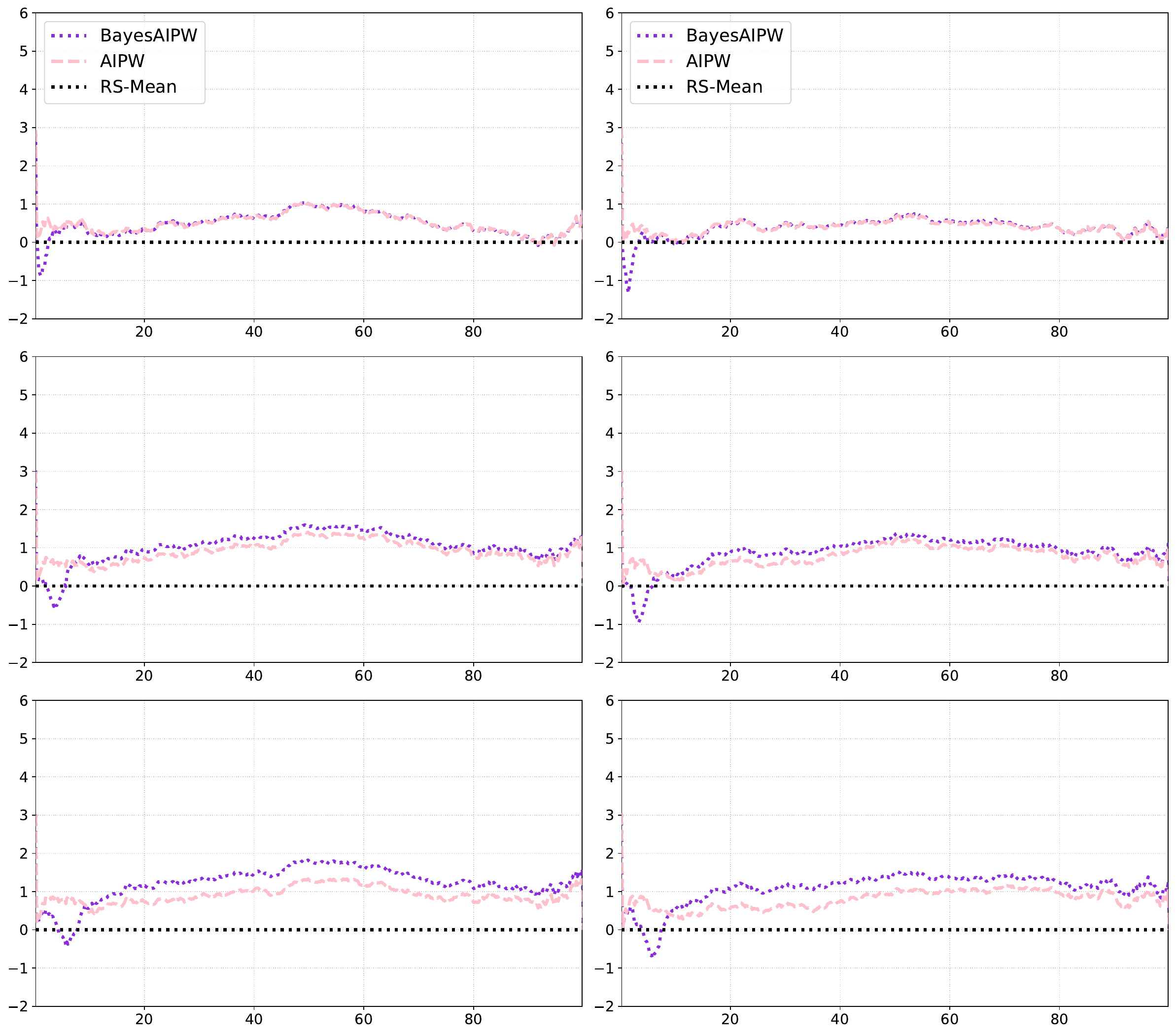} &
% Second 2x3 PDF
\includegraphics[width=0.48\textwidth, valign=c]{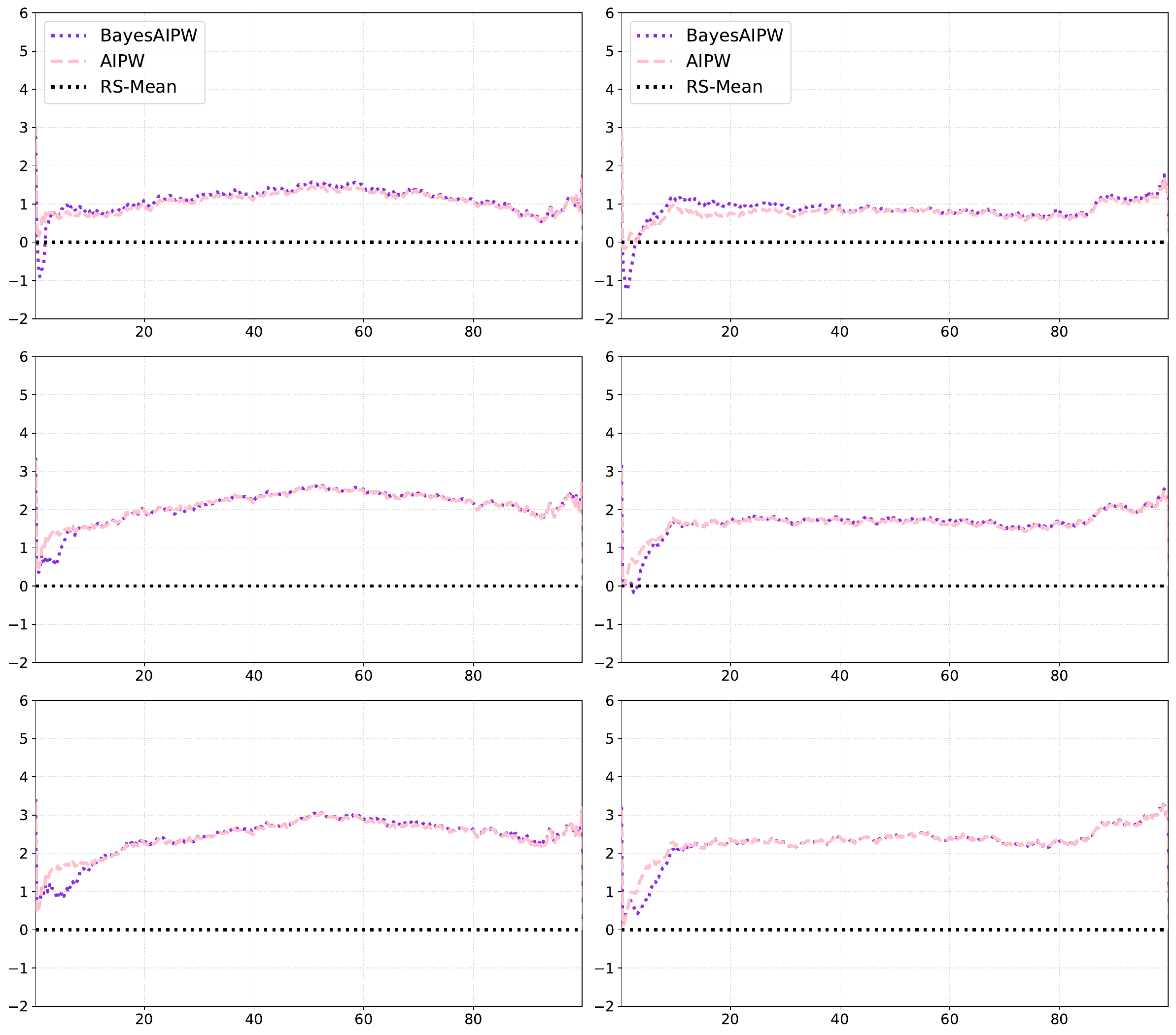} \\
\end{tabular}

\begin{tabular}{c @{\hspace{3pt}} c @{} c}
% --- X-Axis Column Headers (Top) ---
& 
% Headers for the FIRST 2x3 PDF
\makebox[0.22\textwidth][c]{\scriptsize{Interpolation}} \makebox[0.22\textwidth][c]{\scriptsize{Extrapolation}} &
% Headers for the SECOND 2x3 PDF 
\makebox[0.22\textwidth][c]{\scriptsize{Interpolation}} \makebox[0.22\textwidth][c]{\scriptsize{Extrapolation}} \\

\begin{tabular}{@{}c@{}}
    \rotatebox{90}{\scriptsize{10\%}} \\[1.5cm] %
    \rotatebox{90}{\scriptsize{50\%}} \\[1.5cm] %
    \rotatebox{90}{\scriptsize{90\%}}
\end{tabular} &
% First 2x3 PDF
\includegraphics[width=0.48\textwidth, valign=c]{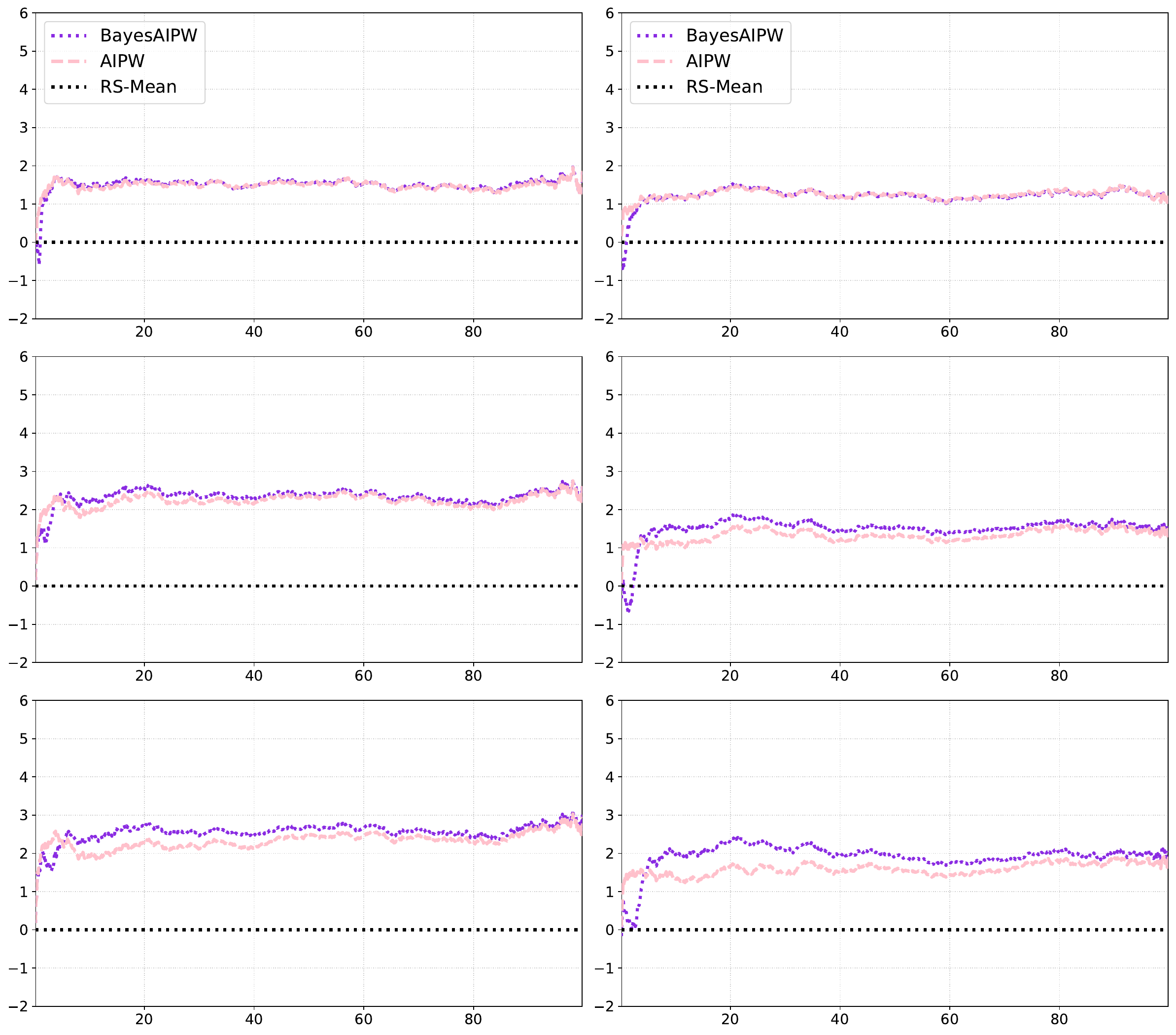} &
% Second 2x3 PDF
\includegraphics[width=0.48\textwidth, valign=c]{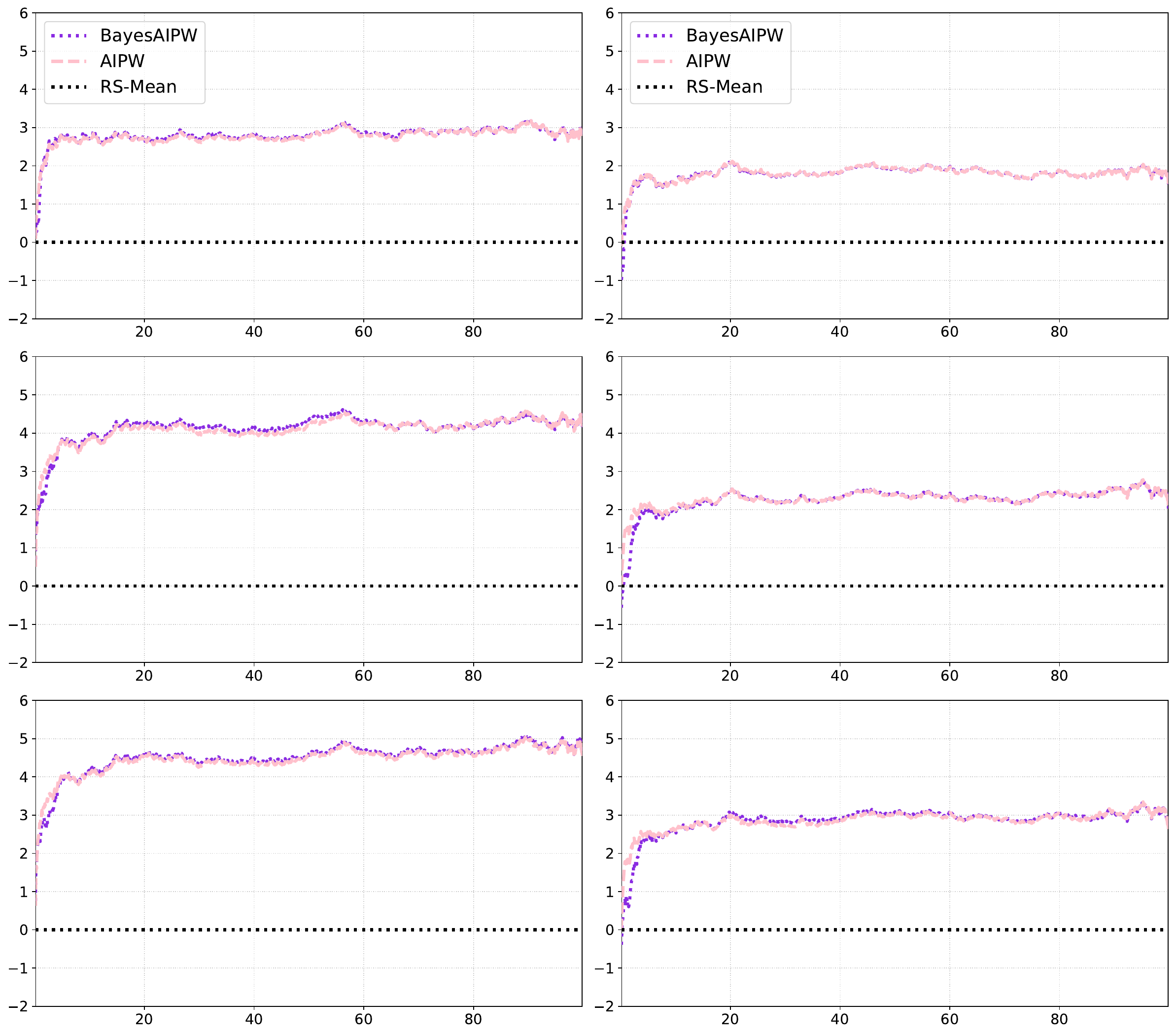} \\
\end{tabular}
\caption{\textbf{Single-target Setting.} \textbf{GPQA} (top three rows) and \textbf{MMLU-Pro} (bottom three rows) with binary scores (left two columns) and continuous scores (right two columns). \RMSELog for the AIPW ablation study across varying proportions of reference models (10\%, 50\%, and 90\%) ablation for AIPW.}
\label{fig:GPQA-MMLUPRO-SAMPLING-SCORING-AIPW}
\end{figure}

Figure \ref{fig:GPQA-MMLUPRO-SAMPLING-SCORING-AIPW} shows that our Bayesian extension of \AIPW matches or outperforms the original method of \cite{zhang2025how}.

We present extended results for all benchmarks considered. To facilitate navigation of the plots, Table~\ref{table:figure_identification} summarizes the figures corresponding to each benchmark and evaluation metric.

\begin{table*}[t]
    \centering
    \resizebox{\textwidth}{!}{
    \begin{tabular}{l ccccccc}
        \toprule
        \multicolumn{1}{l}{\textbf{\textsc{Plot Type}}} & \textsc{GPQA}   & \textsc{MMLU-Pro} & \textsc{BBH} & \textsc{ARC-Challenge} & \textsc{MuSR} & \textsc{IFEval} & \textsc{Natural QA Openbook} \\
        \midrule
        \RMSELog, \RMSEGain\ \& Cost-Accuracy Tradeoff &  \ref{fig:GPQA-SAMPLING-SCORING-RMSELog-GAIN-CAT} & \ref{fig:MMLU-SAMPLING-SCORING-RMSELog-GAIN-CAT} &  \ref{fig:BBH-SAMPLING-SCORING-RMSELog-GAIN-CAT} & \ref{fig:ARC-CHALLENGE-SAMPLING-SCORING-RMSELog-GAIN-CAT} &  \ref{fig:MuSR-SAMPLING-SCORING-RMSELog-GAIN-CAT} & \ref{fig:IFEVAL-NaturalQAOpenbook-RMSELog-GAIN-CAT} & \ref{fig:IFEVAL-NaturalQAOpenbook-RMSELog-GAIN-CAT}\\ \addlinespace
        \textsc{Spearman Correlation}     &  \ref{fig:GPQA-BBH-SAMPLING-SCORING-SPEARMAN} & -- &  \ref{fig:GPQA-BBH-SAMPLING-SCORING-SPEARMAN} & -- &  -- & -- & -- \\ \addlinespace
        \bottomrule
    \end{tabular}
    }
    \caption{Figure index for the extended experimental results across all benchmarks.}
    \label{table:figure_identification}
\end{table*}

\subsection{Multi-target Setting}\label{sec:app:additional_results_benchmarks_multi}
\begin{figure}[t!]
\centering
\renewcommand{\arraystretch}{1.2} 

\begin{tabular}{c @{\hspace{10pt}} c @{} c}
% --- X-Axis Column Headers (Row 1) ---
% --- X-Axis Column Headers (Row 2) ---

% Headers for the FIRST  PDF
\makebox[0.22\textwidth][c]{\scriptsize Low correlation} \makebox[0.22\textwidth][c]{\scriptsize High correlation} 
% Headers for the SECOND PDF 
\makebox[0.22\textwidth][c]{\scriptsize Low correlation} \makebox[0.22\textwidth][c]{\scriptsize High correlation} \\

% --- Y-Axis Label and Images ---
% First PDF
\includegraphics[width=0.48\textwidth, valign=c]{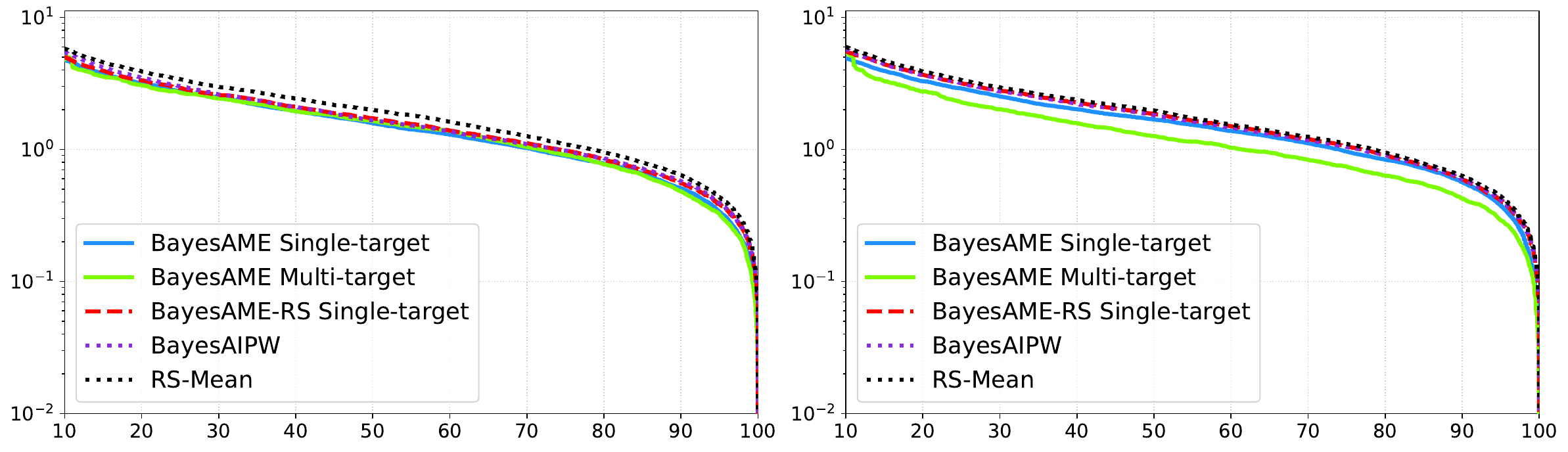} 
\includegraphics[width=0.48\textwidth, valign=c]{figures/gpqa_scoring/plots_comparison_openllm_gpqa_extended_extra_546_267155721_268348824_multitarget_trajectories.pdf} \\
% Second PDF
\includegraphics[width=0.48\textwidth, valign=c]{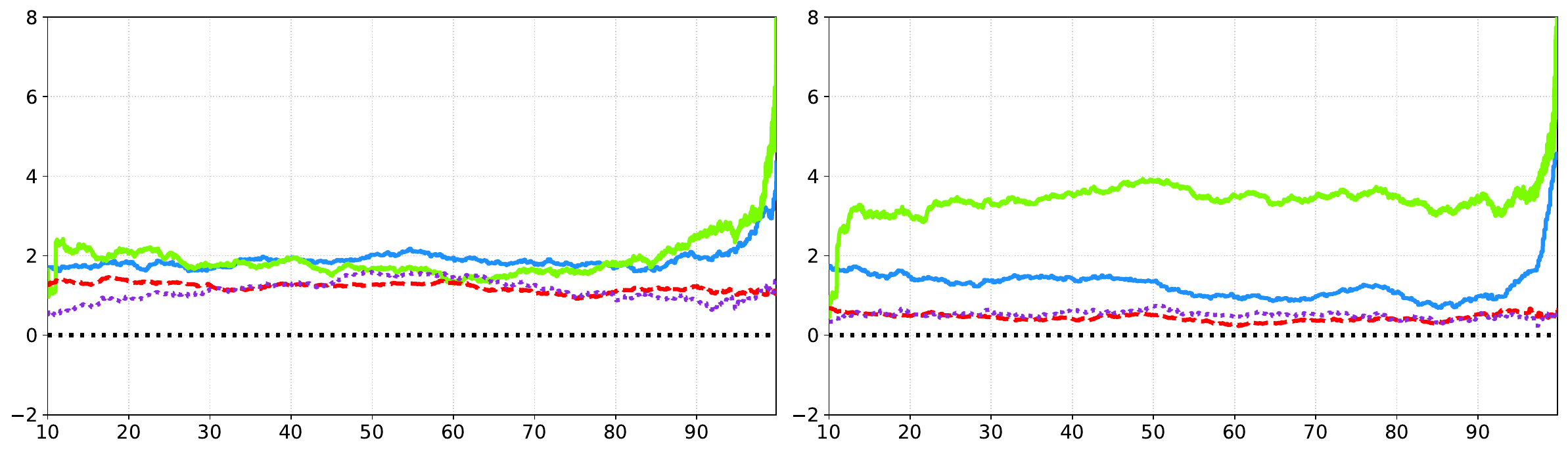} \includegraphics[width=0.48\textwidth, valign=c]{figures/gpqa_scoring/plots_comparison_openllm_gpqa_extended_extra_546_267155721_268348824_multitarget_gain.pdf} 
\\
% Second PDF
\includegraphics[width=0.48\textwidth, valign=c]{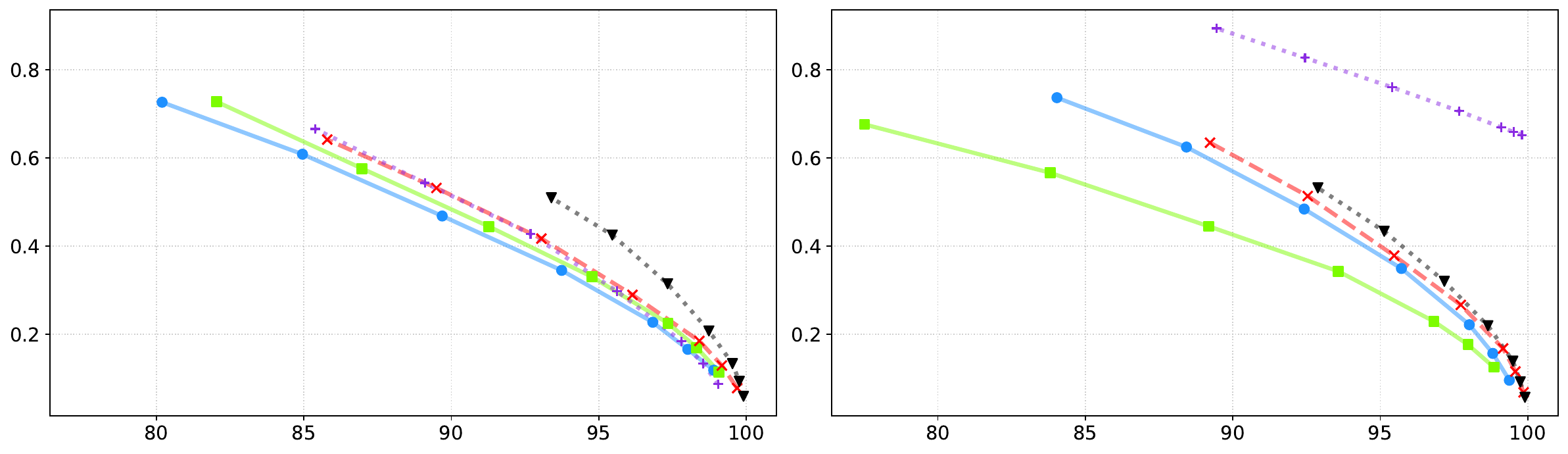} \includegraphics[width=0.48\textwidth, valign=c]{figures/gpqa_scoring/plots_comparison_openllm_gpqa_extended_extra_546_267155721_268348824_multitarget_tradeoff.pdf} 
\\
\end{tabular}
\caption{\textbf{Multiple-target Setting.} \textbf{GPQA} with binary scores (left two columns) and with continuous scores (right two columns). The rows represent \RMSELog, \RMSEGain, and \CAT, respectively. We compare the multitarget method against methods that do not model correlations across targets.}
\label{fig:GPQA-multitarget}
\end{figure}

Figure~\ref{fig:GPQA-multitarget} show the results comparing \BayesAME \acro{Multi-target} with \BayesAME \acro{Single-target} on GPQA using both binary and continuous scores. 

\begin{figure}[t!]
\centering
\renewcommand{\arraystretch}{1.2} 

\begin{tabular}{c @{\hspace{10pt}} c @{} c}
% --- X-Axis Column Headers (Row 1) ---
% --- X-Axis Column Headers (Row 2) ---

% Headers for the FIRST  PDF
\makebox[0.22\textwidth][c]{\scriptsize Low correlation} \makebox[0.22\textwidth][c]{\scriptsize High correlation} 
% Headers for the SECOND PDF 
\makebox[0.22\textwidth][c]{\scriptsize Low correlation} \makebox[0.22\textwidth][c]{\scriptsize High correlation} \\

% --- Y-Axis Label and Images ---
% First PDF
\includegraphics[width=0.48\textwidth, valign=c]{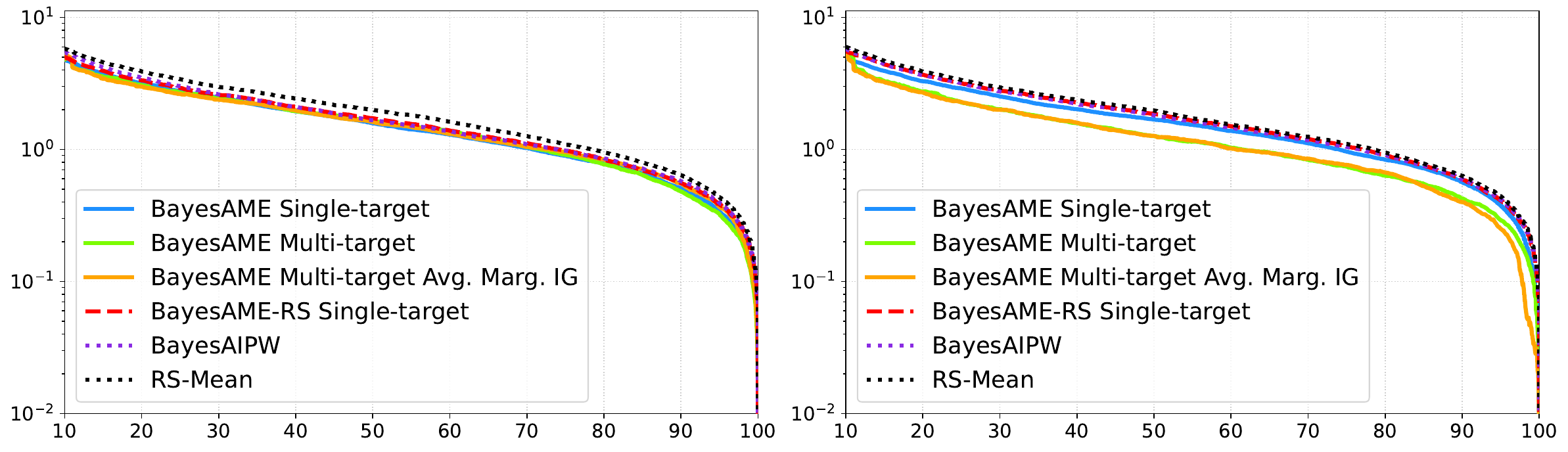} 
\includegraphics[width=0.48\textwidth, valign=c]{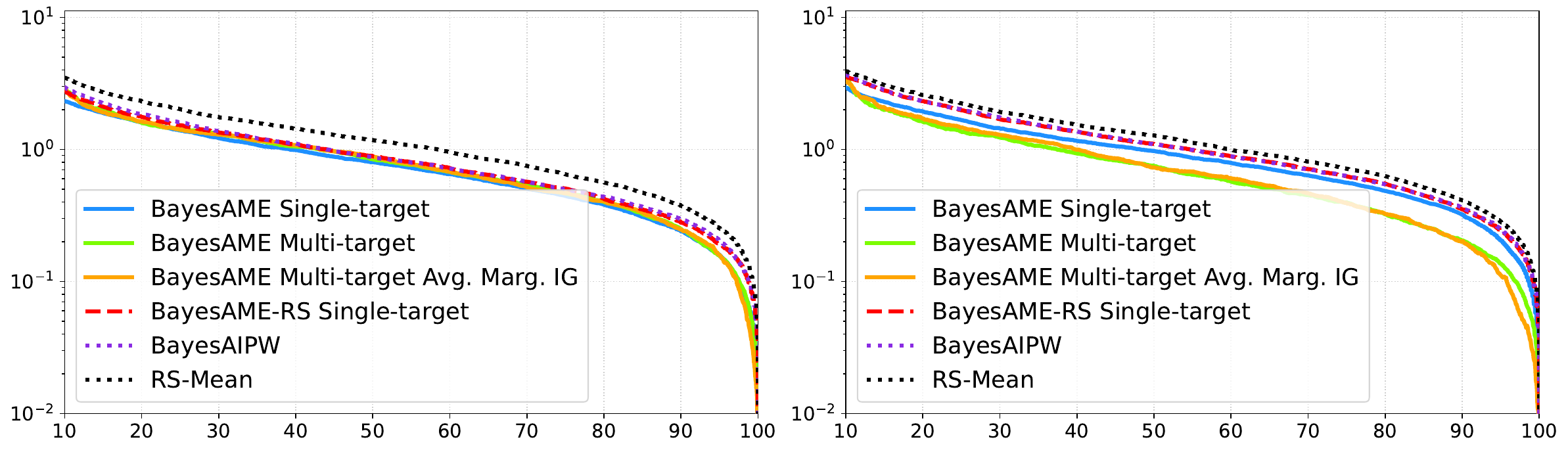} \\
% Second PDF
\includegraphics[width=0.48\textwidth, valign=c]{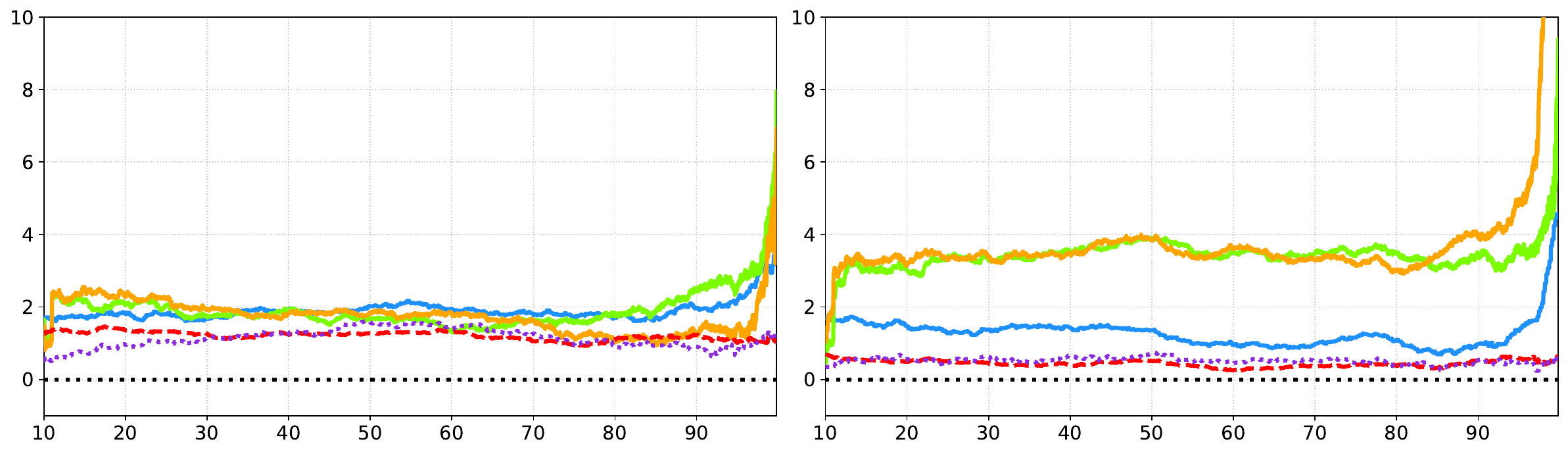} \includegraphics[width=0.48\textwidth, valign=c]{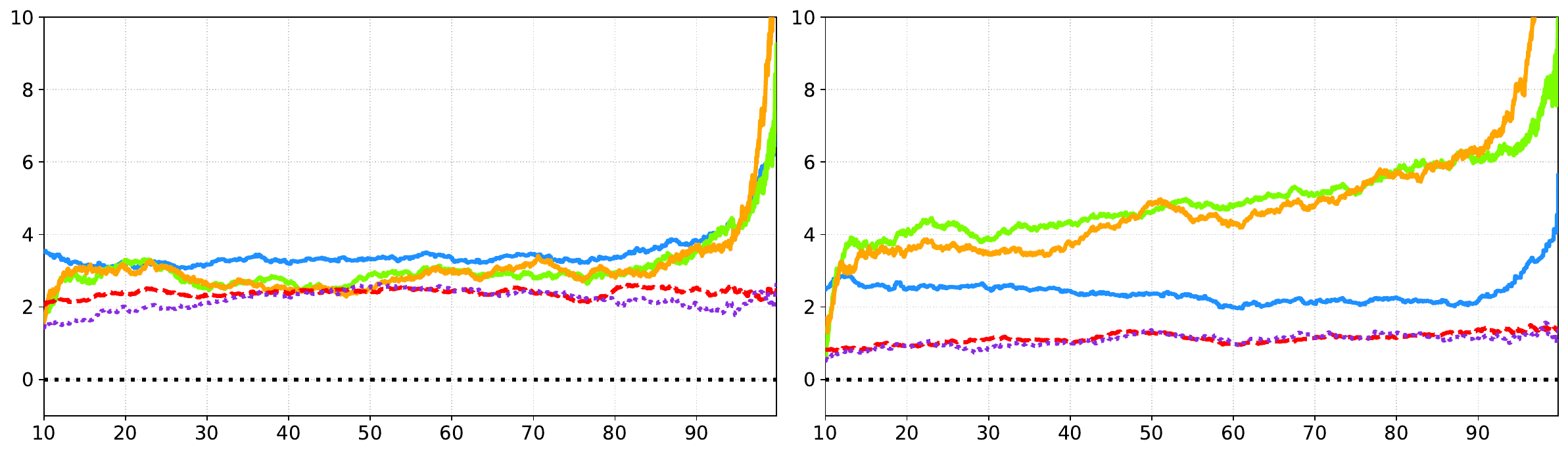} 
\\
% Second PDF
\includegraphics[width=0.48\textwidth, valign=c]{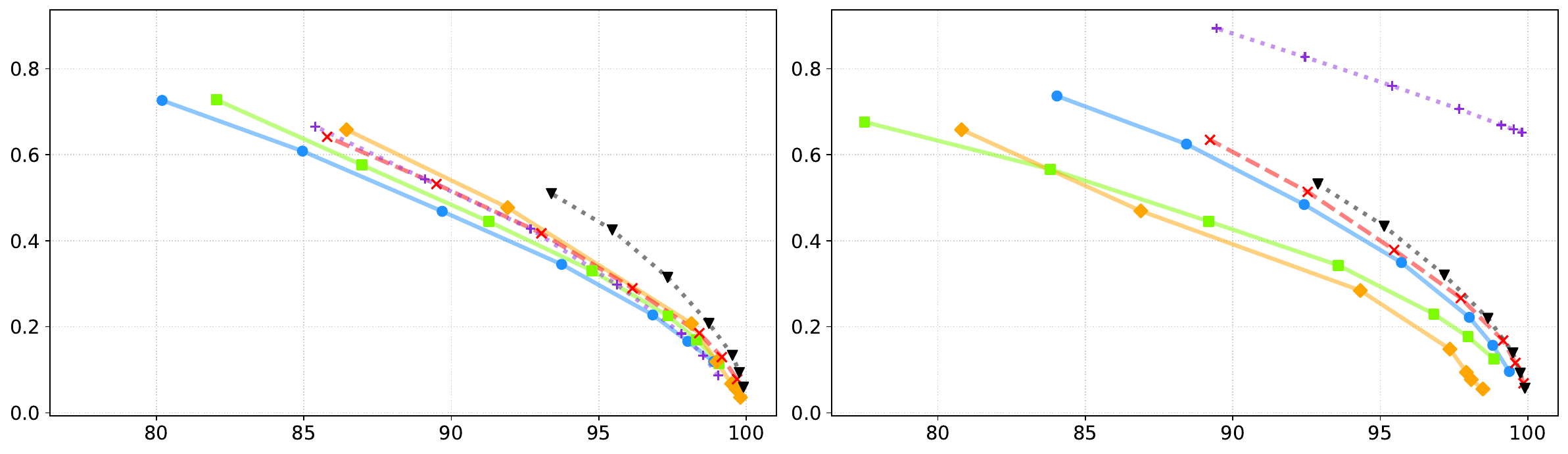} \includegraphics[width=0.48\textwidth, valign=c]{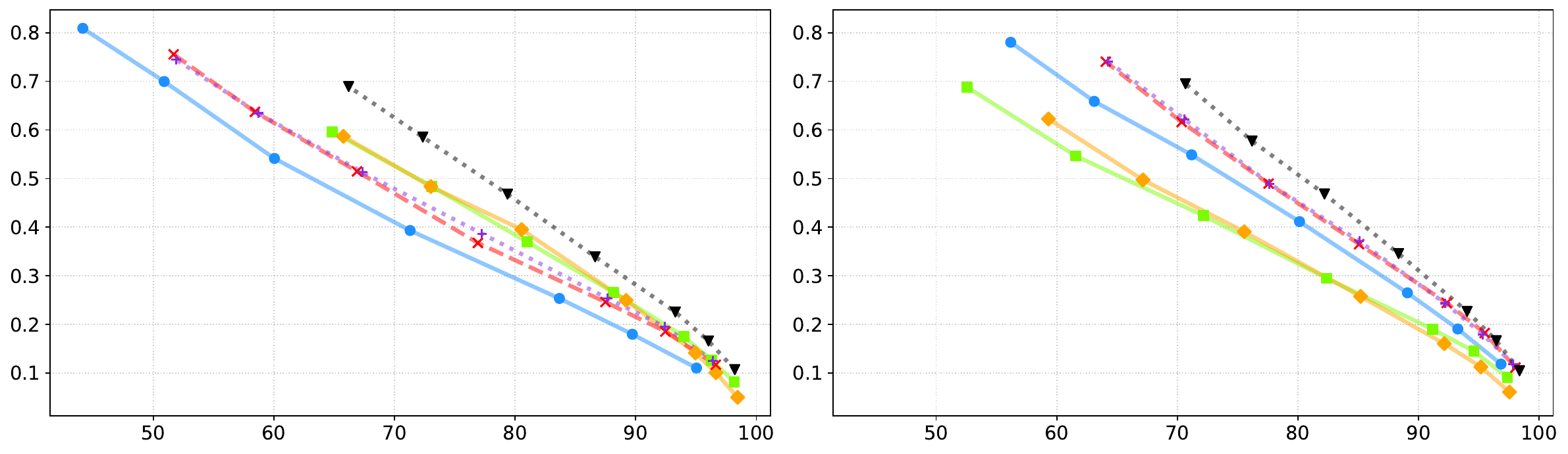} 
\\
\end{tabular}
\caption{\textbf{Multiple-target Setting.} \textbf{GPQA} with binary scores (left two columns) and with continuous scores (right two columns). The rows represent \RMSELog, \RMSEGain, and \CAT, respectively. We compare the different  selection strategies in the multitarget setting.}
\label{fig:GPQA-multitarget-extra-active-learning}
\end{figure}

Figure~\ref{fig:GPQA-multitarget-extra-active-learning} reports results for the alternative selection strategy introduced in Section~\ref{sec:bayesAME_multiple_target}, which selects items based on the average marginalized information gain across target models.

\begin{figure}[!h]
\centering
\renewcommand{\arraystretch}{1.2} 

\begin{tabular}{c @{\hspace{3pt}} c @{} c}
% --- X-Axis Column Headers (Top) ---
& 
% Headers for the FIRST 2x3 PDF
\makebox[0.22\textwidth][c]{\scriptsize{Interpolation}} \makebox[0.22\textwidth][c]{\scriptsize{Extrapolation}} &
% Headers for the SECOND 2x3 PDF 
\makebox[0.22\textwidth][c]{\scriptsize{Interpolation}} \makebox[0.22\textwidth][c]{\scriptsize{Extrapolation}} \\

\begin{tabular}{@{}c@{}}
    \rotatebox{90}{\scriptsize{10\%}} \\[1.5cm] %
    \rotatebox{90}{\scriptsize{50\%}} \\[1.5cm] %
    \rotatebox{90}{\scriptsize{90\%}}
\end{tabular} &
% First 2x3 PDF
\includegraphics[width=0.48\textwidth, valign=c]
{figures/gpqa_sampling/plots_comparison_openllm_gpqa_extended_acc_extra_546_271985553_trajectories.pdf} &
% Second 2x3 PDF
\includegraphics[width=0.48\textwidth, valign=c]
{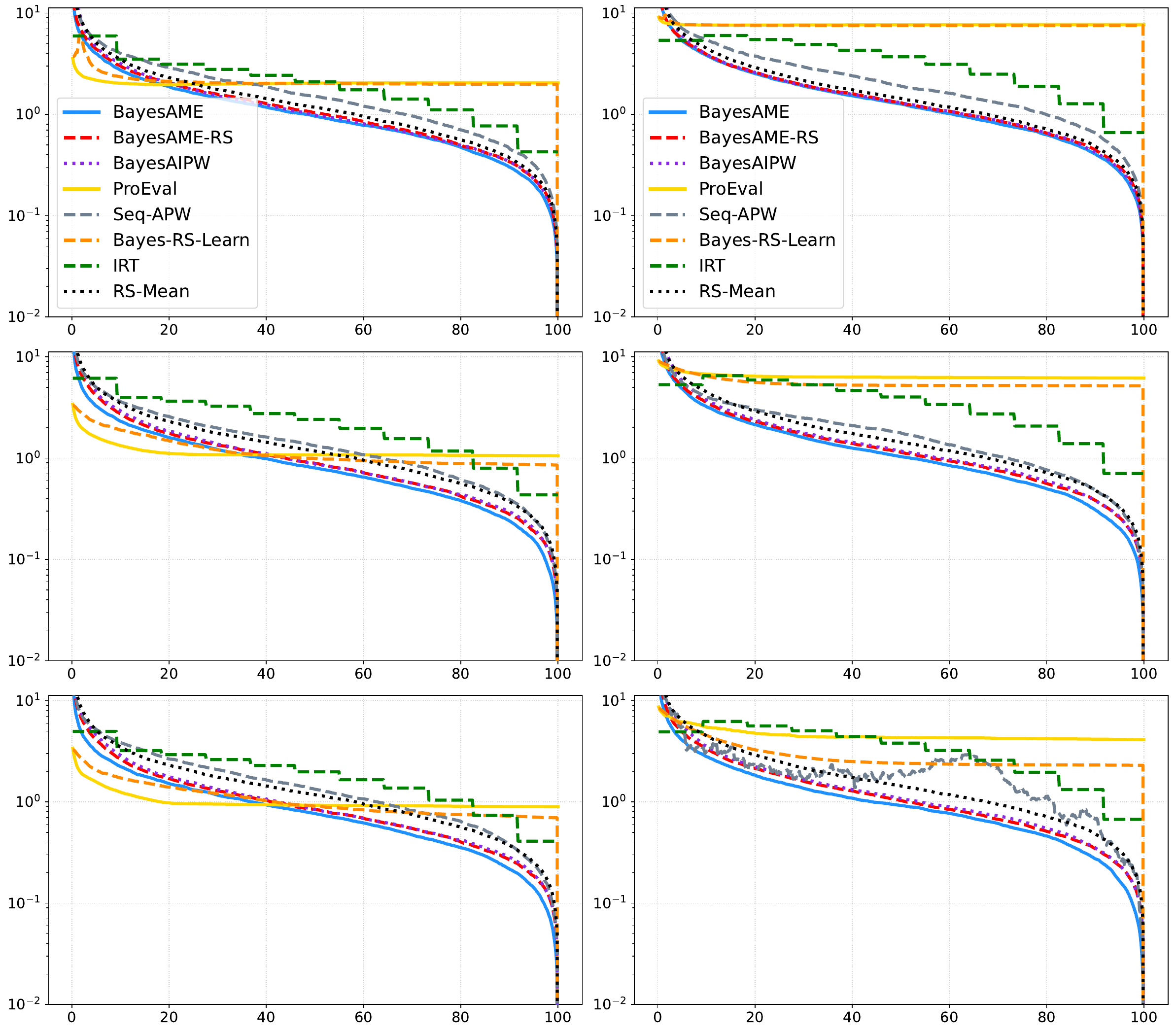} \\
\end{tabular}
\begin{tabular}{c @{\hspace{3pt}} c @{} c}
% --- X-Axis Column Headers (Top) ---
& 
% Headers for the FIRST 2x3 PDF
\makebox[0.22\textwidth][c]{\scriptsize{Interpolation}} \makebox[0.22\textwidth][c]{\scriptsize{Extrapolation}} &
% Headers for the SECOND 2x3 PDF 
\makebox[0.22\textwidth][c]{\scriptsize{Interpolation}} \makebox[0.22\textwidth][c]{\scriptsize{Extrapolation}} \\

\begin{tabular}{@{}c@{}}
    \rotatebox{90}{\scriptsize{10\%}} \\[1.5cm] %
    \rotatebox{90}{\scriptsize{50\%}} \\[1.5cm] %
    \rotatebox{90}{\scriptsize{90\%}}
\end{tabular} &
% First 2x3 PDF
\includegraphics[width=0.48\textwidth, valign=c]
{figures/gpqa_sampling/plots_comparison_openllm_gpqa_extended_acc_extra_546_271985553_gains.pdf} &
% Second 2x3 PDF
\includegraphics[width=0.48\textwidth, valign=c]{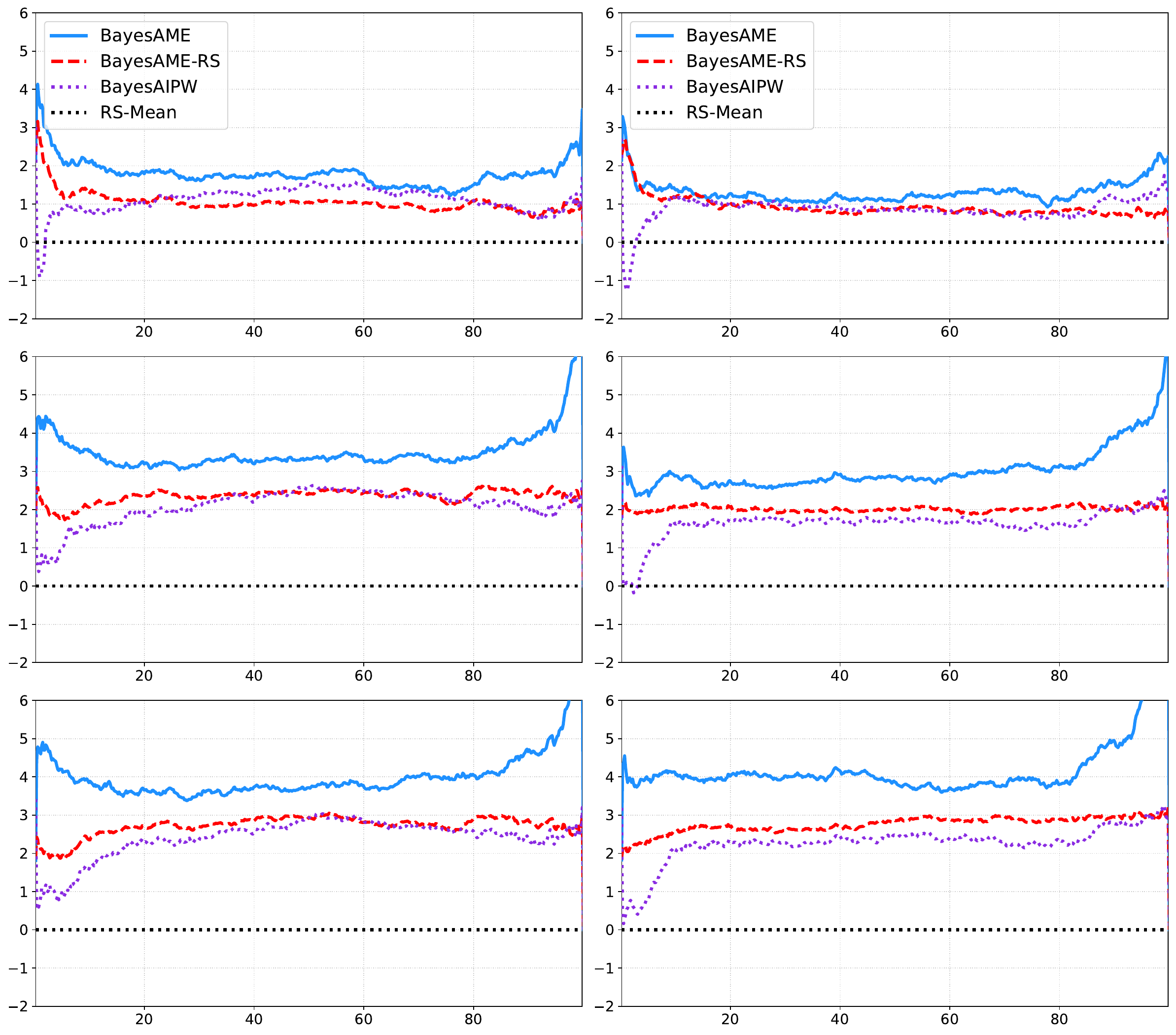} \\
\end{tabular}

\begin{tabular}{c @{\hspace{3pt}} c @{} c}
% --- X-Axis Column Headers (Top) ---
& 
% Headers for the FIRST 2x3 PDF
\makebox[0.22\textwidth][c]{\scriptsize{Interpolation}} \makebox[0.22\textwidth][c]{\scriptsize{Extrapolation}} &
% Headers for the SECOND 2x3 PDF 
\makebox[0.22\textwidth][c]{\scriptsize{Interpolation}} \makebox[0.22\textwidth][c]{\scriptsize{Extrapolation}} \\

\begin{tabular}{@{}c@{}}
    \rotatebox{90}{\scriptsize{10\%}} \\[1.5cm] %
    \rotatebox{90}{\scriptsize{50\%}} \\[1.5cm] %
    \rotatebox{90}{\scriptsize{90\%}}
\end{tabular} &
% First 2x3 PDF
\includegraphics[width=0.48\textwidth, valign=c]
{figures/gpqa_sampling/plots_comparison_openllm_gpqa_extended_acc_extra_546_271985553_trade_off.pdf} &
% Second 2x3 PDF
\includegraphics[width=0.48\textwidth, valign=c]{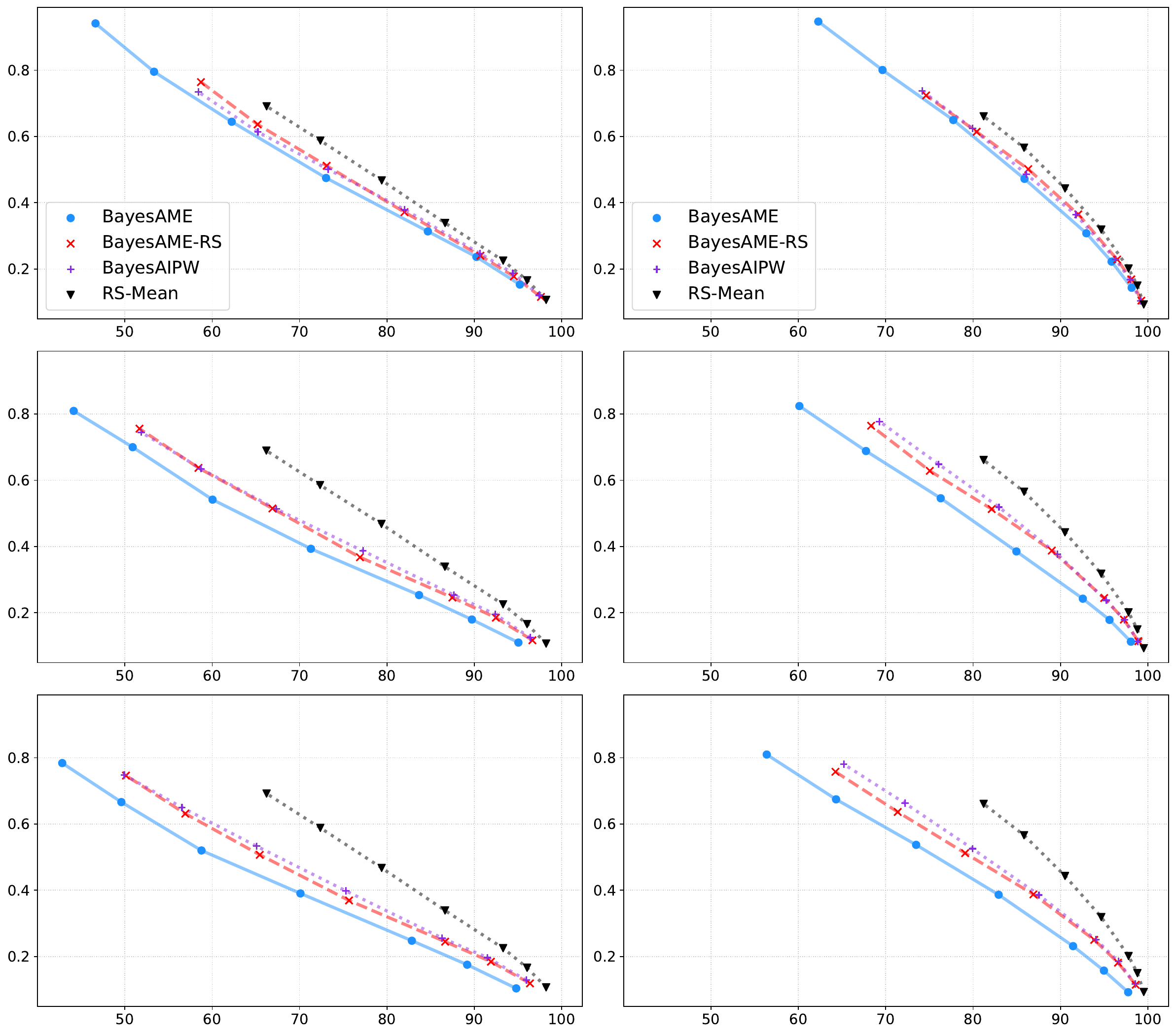} \\
\end{tabular}
\caption{\textbf{Single-target Setting.}  \textbf{GPQA} with binary scores (left two columns) and continuous scores (right two columns). \RMSELog (top three rows), \RMSEGain (middle three rows), and \CAT (bottom three rows) across varying proportions of reference models.}
\label{fig:GPQA-SAMPLING-SCORING-RMSELog-GAIN-CAT}
\end{figure}

\begin{figure}[!h]
\centering
\renewcommand{\arraystretch}{1.2} 

\begin{tabular}{c @{\hspace{3pt}} c @{} c}
% --- X-Axis Column Headers (Top) ---
& 
% Headers for the FIRST 2x3 PDF
\makebox[0.22\textwidth][c]{\scriptsize{Interpolation}} \makebox[0.22\textwidth][c]{\scriptsize{Extrapolation}} &
% Headers for the SECOND 2x3 PDF 
\makebox[0.22\textwidth][c]{\scriptsize{Interpolation}} \makebox[0.22\textwidth][c]{\scriptsize{Extrapolation}} \\

\begin{tabular}{@{}c@{}}
    \rotatebox{90}{\scriptsize{10\%}} \\[1.5cm] %
    \rotatebox{90}{\scriptsize{50\%}} \\[1.5cm] %
    \rotatebox{90}{\scriptsize{90\%}}
\end{tabular} &
% First 2x3 PDF
\includegraphics[width=0.48\textwidth, valign=c]
{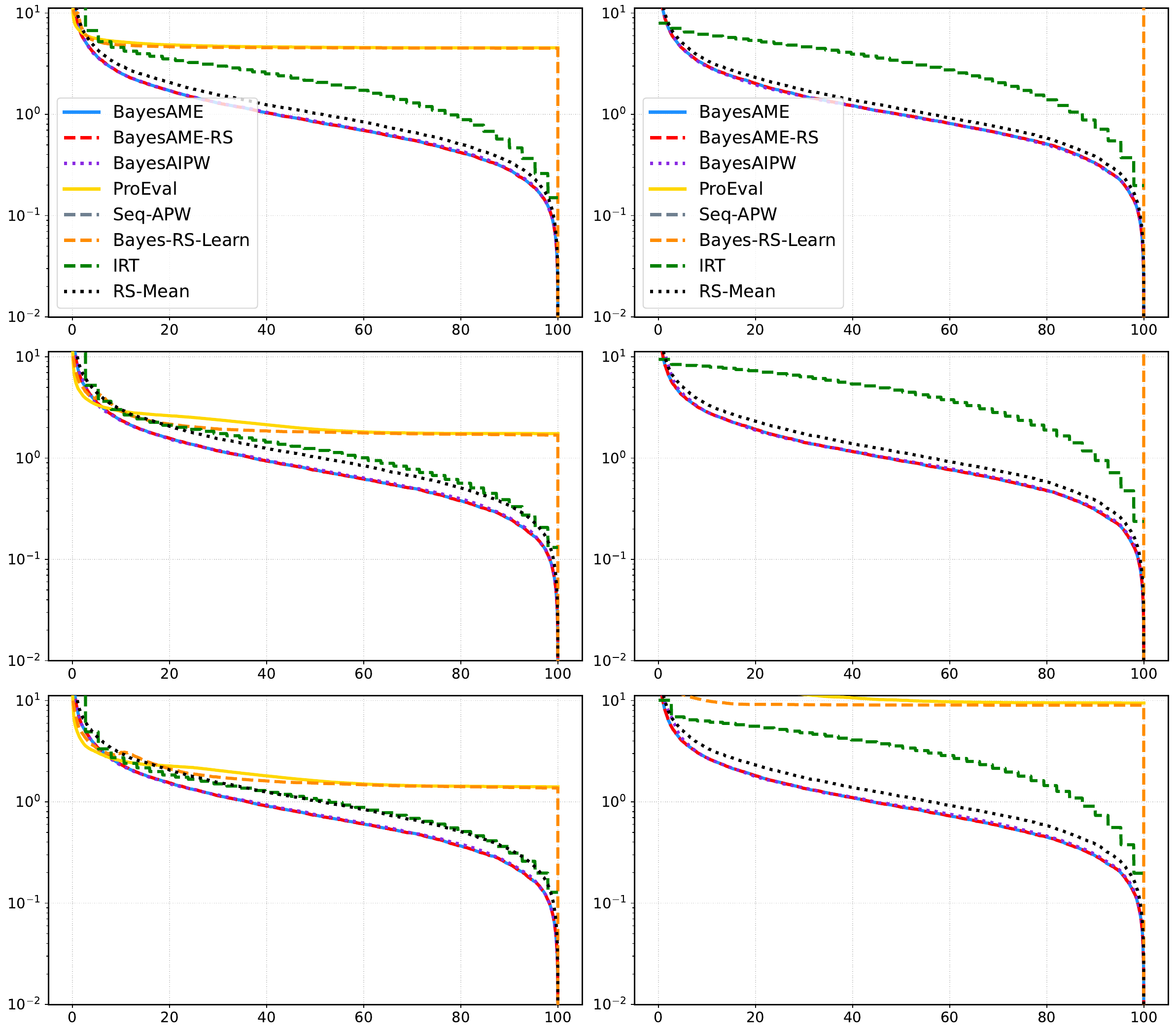} &
% Second 2x3 PDF
\includegraphics[width=0.48\textwidth, valign=c]
{figures/mmlu_pro_scoring/plots_comparison_openllm_mmlu_pro_subset_extra_1890_266940212_trajectories.pdf} \\
\end{tabular}
\begin{tabular}{c @{\hspace{3pt}} c @{} c}
% --- X-Axis Column Headers (Top) ---
& 
% Headers for the FIRST 2x3 PDF
\makebox[0.22\textwidth][c]{\scriptsize{Interpolation}} \makebox[0.22\textwidth][c]{\scriptsize{Extrapolation}} &
% Headers for the SECOND 2x3 PDF 
\makebox[0.22\textwidth][c]{\scriptsize{Interpolation}} \makebox[0.22\textwidth][c]{\scriptsize{Extrapolation}} \\

\begin{tabular}{@{}c@{}}
    \rotatebox{90}{\scriptsize{10\%}} \\[1.5cm] %
    \rotatebox{90}{\scriptsize{50\%}} \\[1.5cm] %
    \rotatebox{90}{\scriptsize{90\%}}
\end{tabular} &
% First 2x3 PDF
\includegraphics[width=0.48\textwidth, valign=c]
{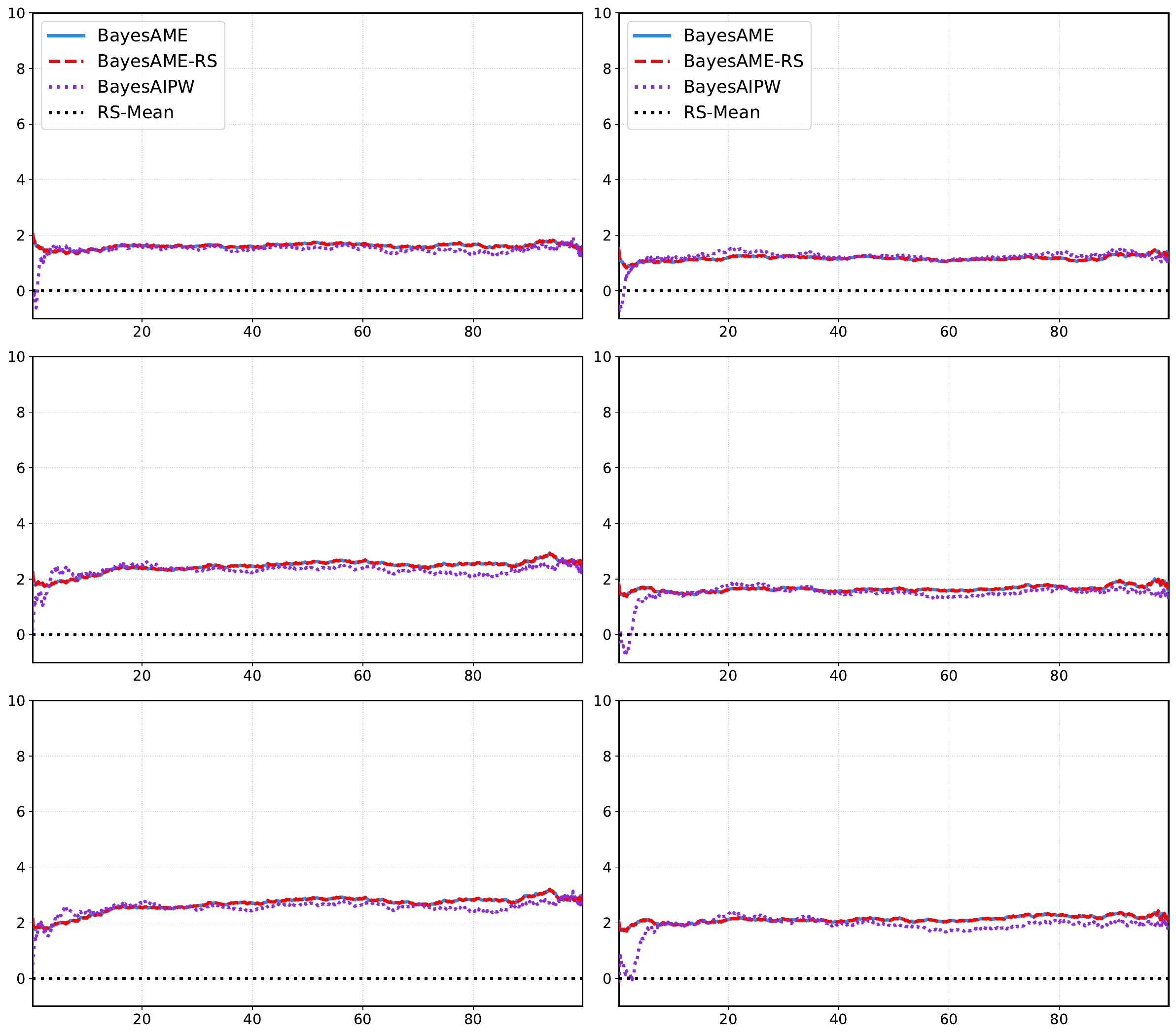} &
% Second 2x3 PDF
\includegraphics[width=0.48\textwidth, valign=c]{figures/mmlu_pro_scoring/plots_comparison_openllm_mmlu_pro_subset_extra_1890_266940212_gain.pdf} \\
\end{tabular}

\begin{tabular}{c @{\hspace{3pt}} c @{} c}
% --- X-Axis Column Headers (Top) ---
& 
% Headers for the FIRST 2x3 PDF
\makebox[0.22\textwidth][c]{\scriptsize{Interpolation}} \makebox[0.22\textwidth][c]{\scriptsize{Extrapolation}} &
% Headers for the SECOND 2x3 PDF 
\makebox[0.22\textwidth][c]{\scriptsize{Interpolation}} \makebox[0.22\textwidth][c]{\scriptsize{Extrapolation}} \\

\begin{tabular}{@{}c@{}}
    \rotatebox{90}{\scriptsize{10\%}} \\[1.5cm] %
    \rotatebox{90}{\scriptsize{50\%}} \\[1.5cm] %
    \rotatebox{90}{\scriptsize{90\%}}
\end{tabular} &
% First 2x3 PDF
\includegraphics[width=0.48\textwidth, valign=c]
{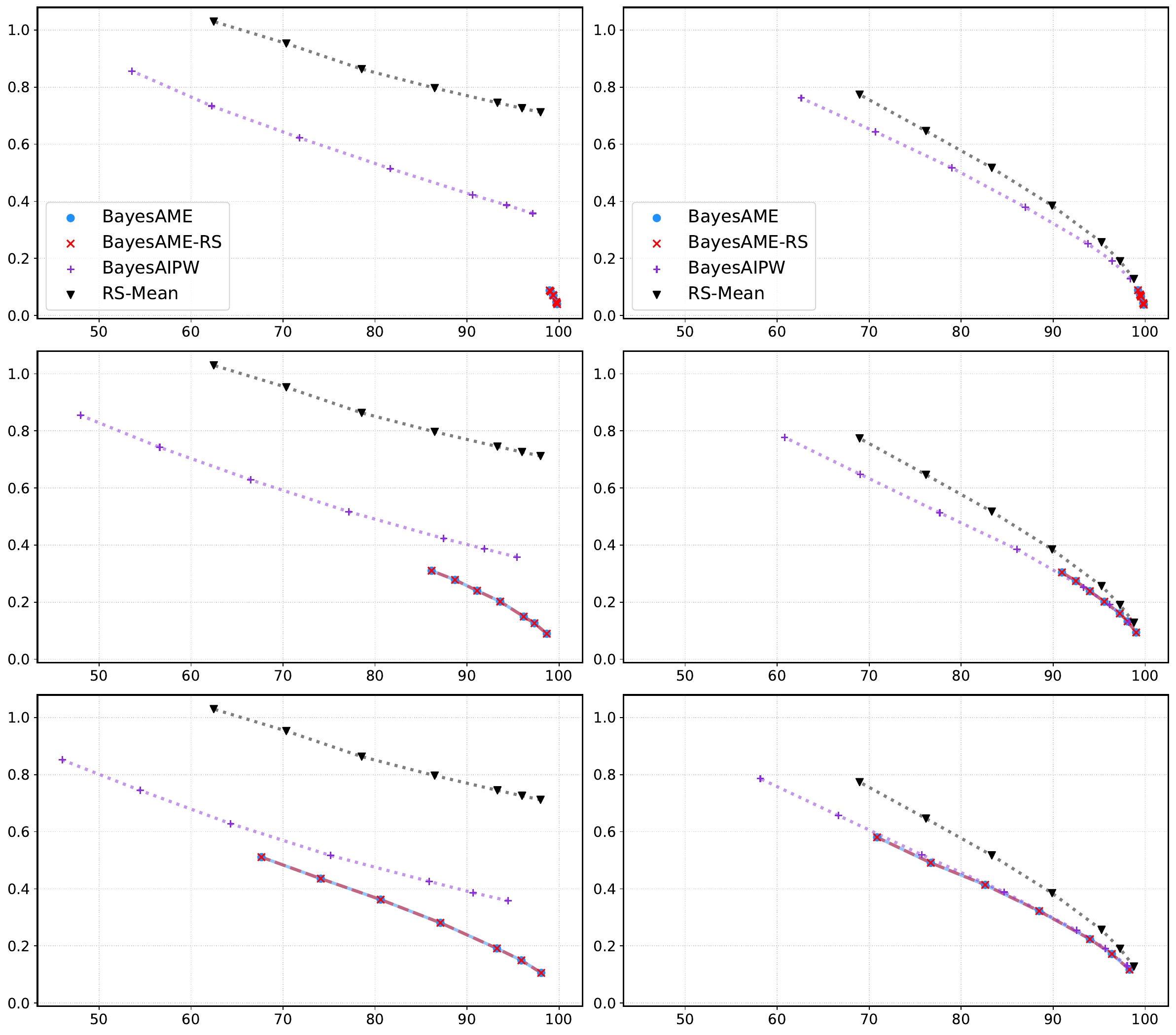} &
% Second 2x3 PDF
\includegraphics[width=0.48\textwidth, valign=c]{figures/mmlu_pro_scoring/plots_comparison_openllm_mmlu_pro_subset_extra_1890_266940212_tradeoff.pdf} \\
\end{tabular}
\caption{\textbf{Single-target Setting.}  \textbf{MMLU-Pro} with binary scores (left two columns) and continuous scores (right two columns). \RMSELog (top three rows), \RMSEGain (middle three rows), and \CAT (bottom three rows) across varying proportions of reference models.}
\label{fig:MMLU-SAMPLING-SCORING-RMSELog-GAIN-CAT}
\end{figure}

\begin{figure}[!h]
\centering
\renewcommand{\arraystretch}{1.2} 

\begin{tabular}{c @{\hspace{3pt}} c @{} c}
% --- X-Axis Column Headers (Top) ---
& 
% Headers for the FIRST 2x3 PDF
\makebox[0.22\textwidth][c]{\scriptsize{Interpolation}} \makebox[0.22\textwidth][c]{\scriptsize{Extrapolation}} &
% Headers for the SECOND 2x3 PDF 
\makebox[0.22\textwidth][c]{\scriptsize{Interpolation}} \makebox[0.22\textwidth][c]{\scriptsize{Extrapolation}} \\

\begin{tabular}{@{}c@{}}
    \rotatebox{90}{\scriptsize{10\%}} \\[1.5cm] %
    \rotatebox{90}{\scriptsize{50\%}} \\[1.5cm] %
    \rotatebox{90}{\scriptsize{90\%}}
\end{tabular} &
% First 2x3 PDF
\includegraphics[width=0.48\textwidth, valign=c]
{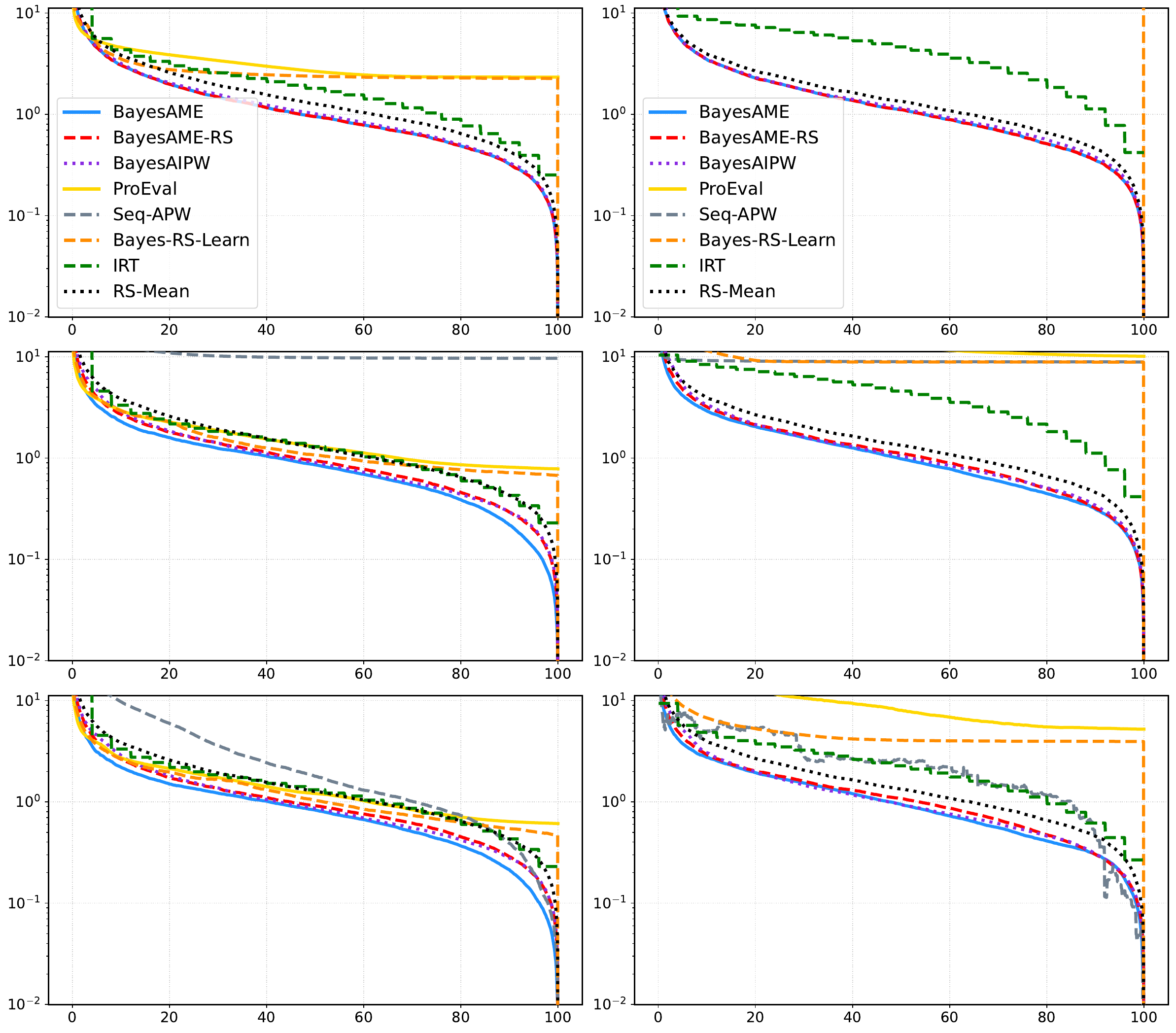} &
% Second 2x3 PDF
\includegraphics[width=0.48\textwidth, valign=c]
{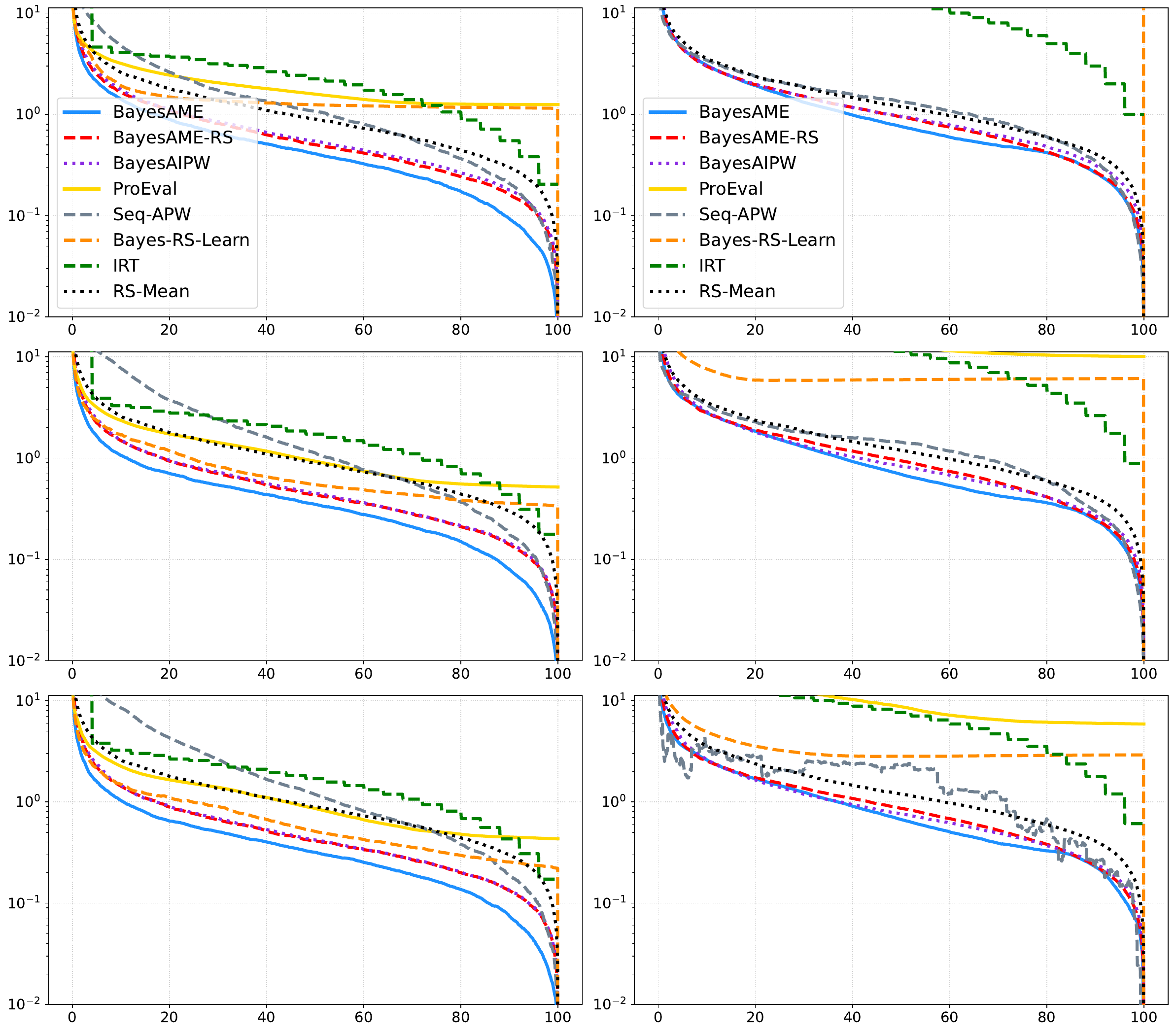} \\
\end{tabular}
\begin{tabular}{c @{\hspace{3pt}} c @{} c}
% --- X-Axis Column Headers (Top) ---
& 
% Headers for the FIRST 2x3 PDF
\makebox[0.22\textwidth][c]{\scriptsize{Interpolation}} \makebox[0.22\textwidth][c]{\scriptsize{Extrapolation}} &
% Headers for the SECOND 2x3 PDF 
\makebox[0.22\textwidth][c]{\scriptsize{Interpolation}} \makebox[0.22\textwidth][c]{\scriptsize{Extrapolation}} \\

\begin{tabular}{@{}c@{}}
    \rotatebox{90}{\scriptsize{10\%}} \\[1.5cm] %
    \rotatebox{90}{\scriptsize{50\%}} \\[1.5cm] %
    \rotatebox{90}{\scriptsize{90\%}}
\end{tabular} &
% First 2x3 PDF
\includegraphics[width=0.48\textwidth, valign=c]
{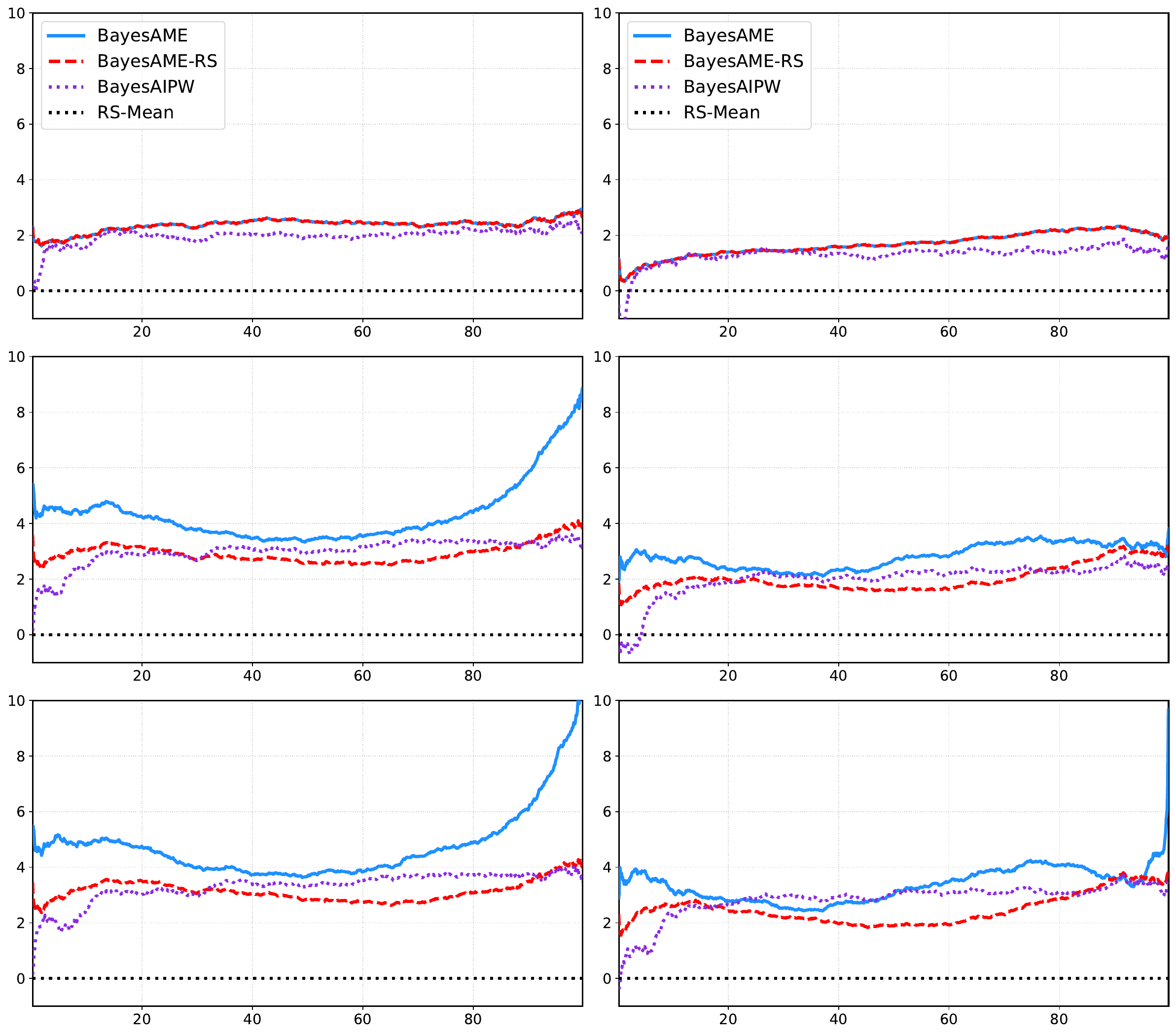} &
% Second 2x3 PDF
\includegraphics[width=0.48\textwidth, valign=c]{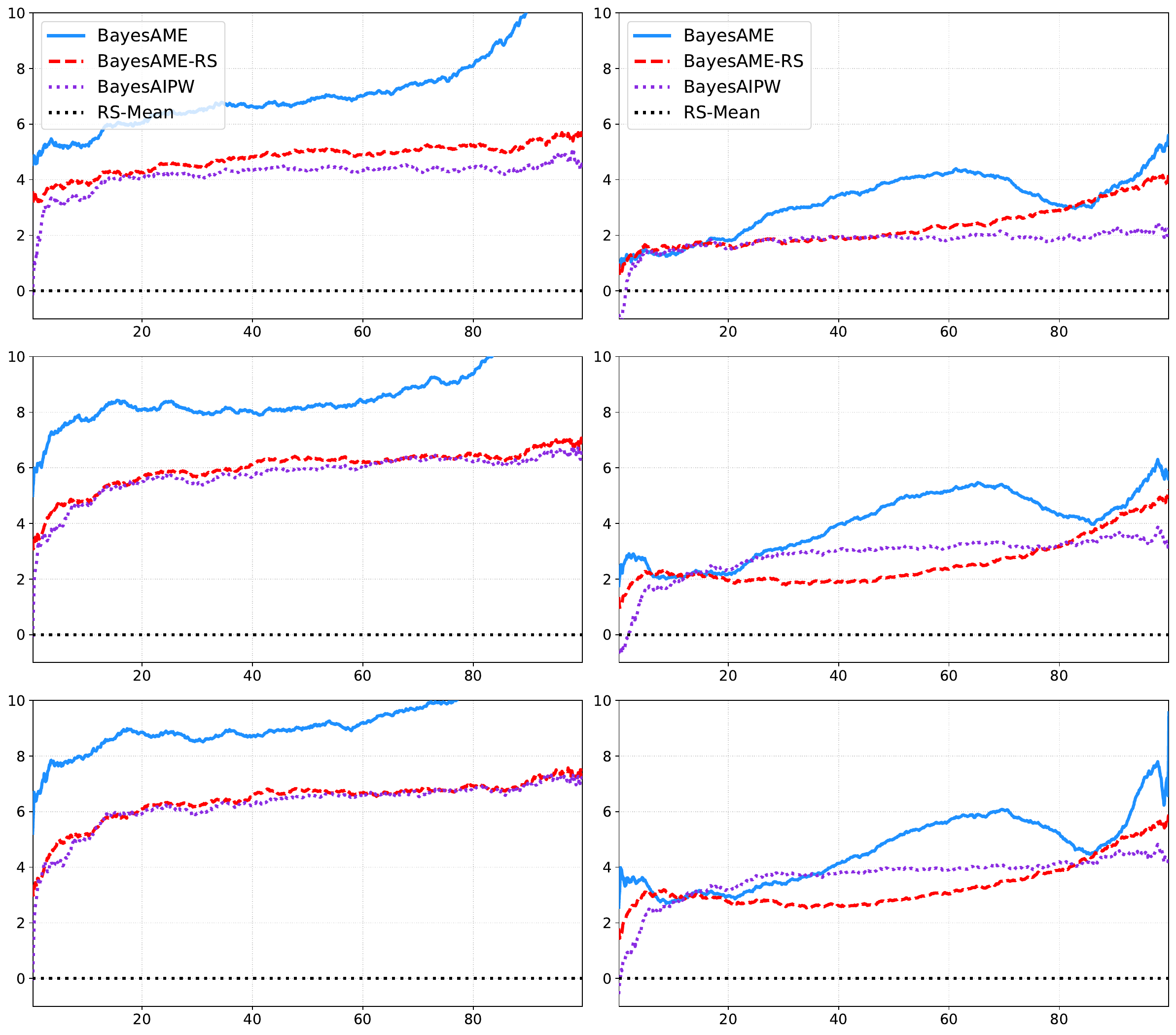} \\
\end{tabular}

\begin{tabular}{c @{\hspace{3pt}} c @{} c}
% --- X-Axis Column Headers (Top) ---
& 
% Headers for the FIRST 2x3 PDF
\makebox[0.22\textwidth][c]{\scriptsize{Interpolation}} \makebox[0.22\textwidth][c]{\scriptsize{Extrapolation}} &
% Headers for the SECOND 2x3 PDF 
\makebox[0.22\textwidth][c]{\scriptsize{Interpolation}} \makebox[0.22\textwidth][c]{\scriptsize{Extrapolation}} \\

\begin{tabular}{@{}c@{}}
    \rotatebox{90}{\scriptsize{10\%}} \\[1.5cm] %
    \rotatebox{90}{\scriptsize{50\%}} \\[1.5cm] %
    \rotatebox{90}{\scriptsize{90\%}}
\end{tabular} &
% First 2x3 PDF
\includegraphics[width=0.48\textwidth, valign=c]
{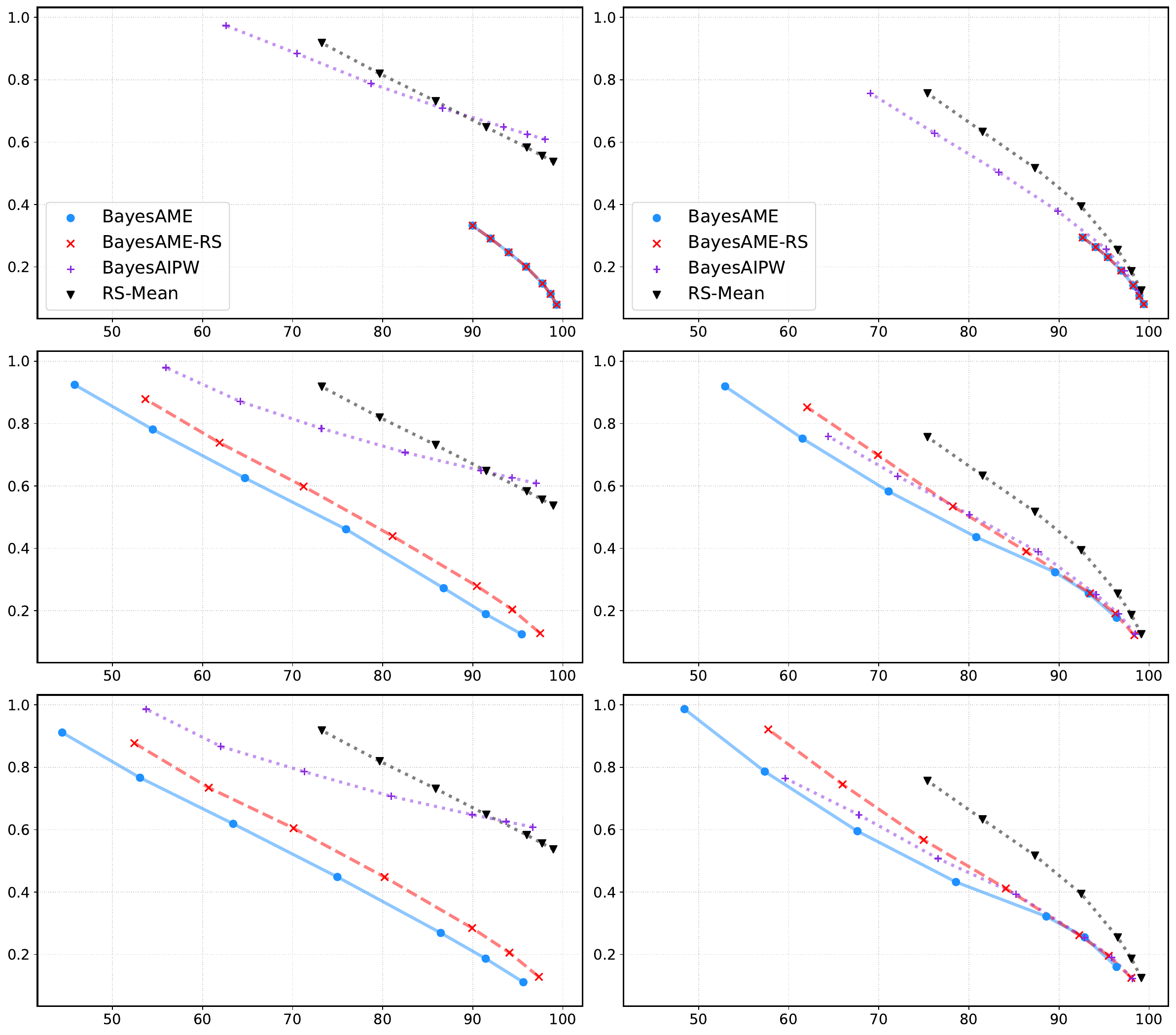} &
% Second 2x3 PDF
\includegraphics[width=0.48\textwidth, valign=c]{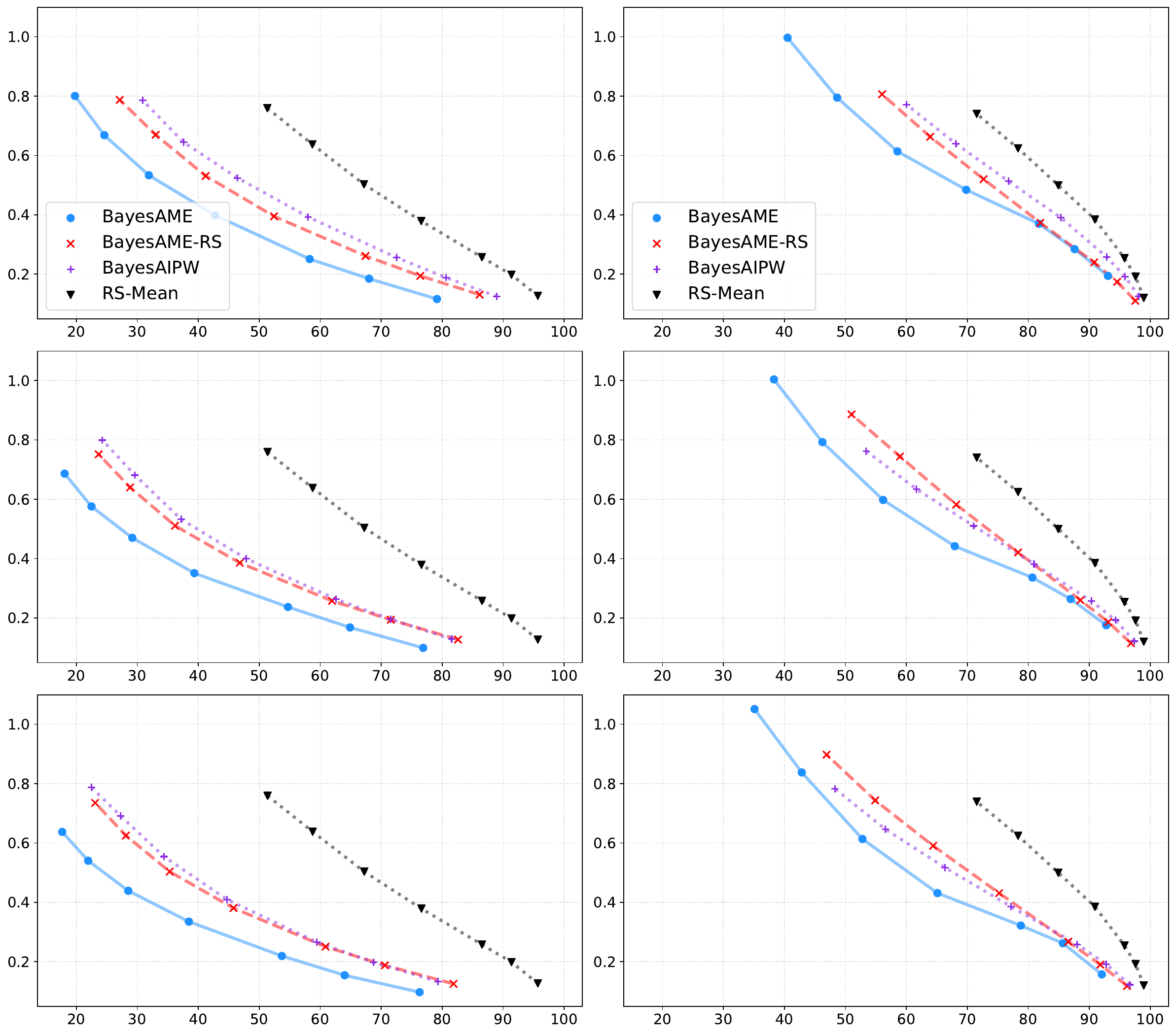} \\
\end{tabular}
\caption{\textbf{Single-target Setting.} \textbf{BBH} with binary scores (left two columns) and continuous scores (right two columns). \RMSELog (top three rows), \RMSEGain (middle three rows), and \CAT (bottom three rows) across varying proportions of reference models.}
\label{fig:BBH-SAMPLING-SCORING-RMSELog-GAIN-CAT}
\end{figure}

\begin{figure}[!h]
\centering
\renewcommand{\arraystretch}{1.2} 

\begin{tabular}{c @{\hspace{3pt}} c @{} c}
% --- X-Axis Column Headers (Top) ---
& 
% Headers for the FIRST 2x3 PDF
\makebox[0.22\textwidth][c]{\scriptsize{Interpolation}} \makebox[0.22\textwidth][c]{\scriptsize{Extrapolation}} &
% Headers for the SECOND 2x3 PDF 
\makebox[0.22\textwidth][c]{\scriptsize{Interpolation}} \makebox[0.22\textwidth][c]{\scriptsize{Extrapolation}} \\

\begin{tabular}{@{}c@{}}
    \rotatebox{90}{\scriptsize{10\%}} \\[1.5cm] %
    \rotatebox{90}{\scriptsize{50\%}} \\[1.5cm] %
    \rotatebox{90}{\scriptsize{90\%}}
\end{tabular} &
% First 2x3 PDF
\includegraphics[width=0.48\textwidth, valign=c]
{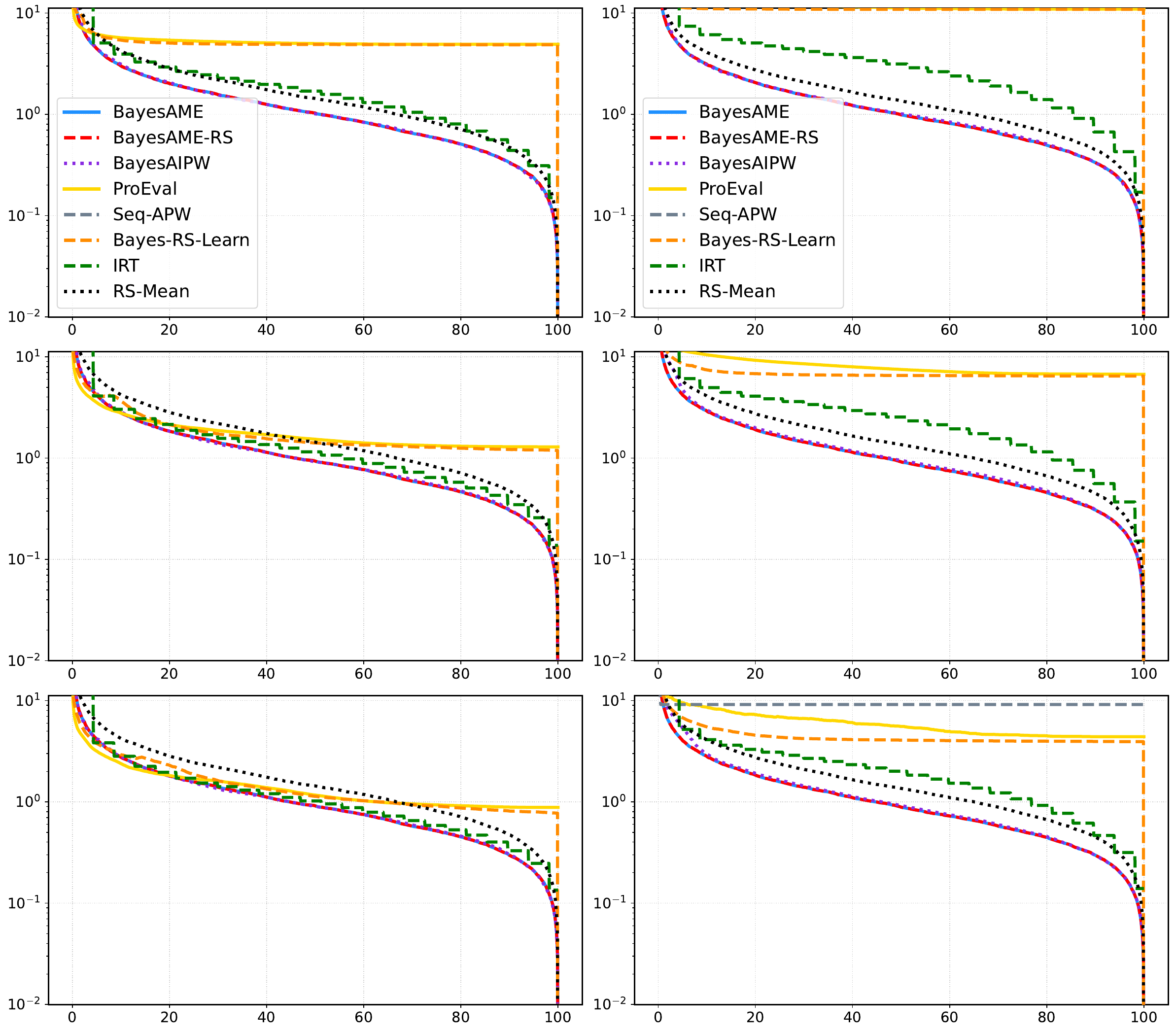} &
% Second 2x3 PDF
\includegraphics[width=0.48\textwidth, valign=c]
{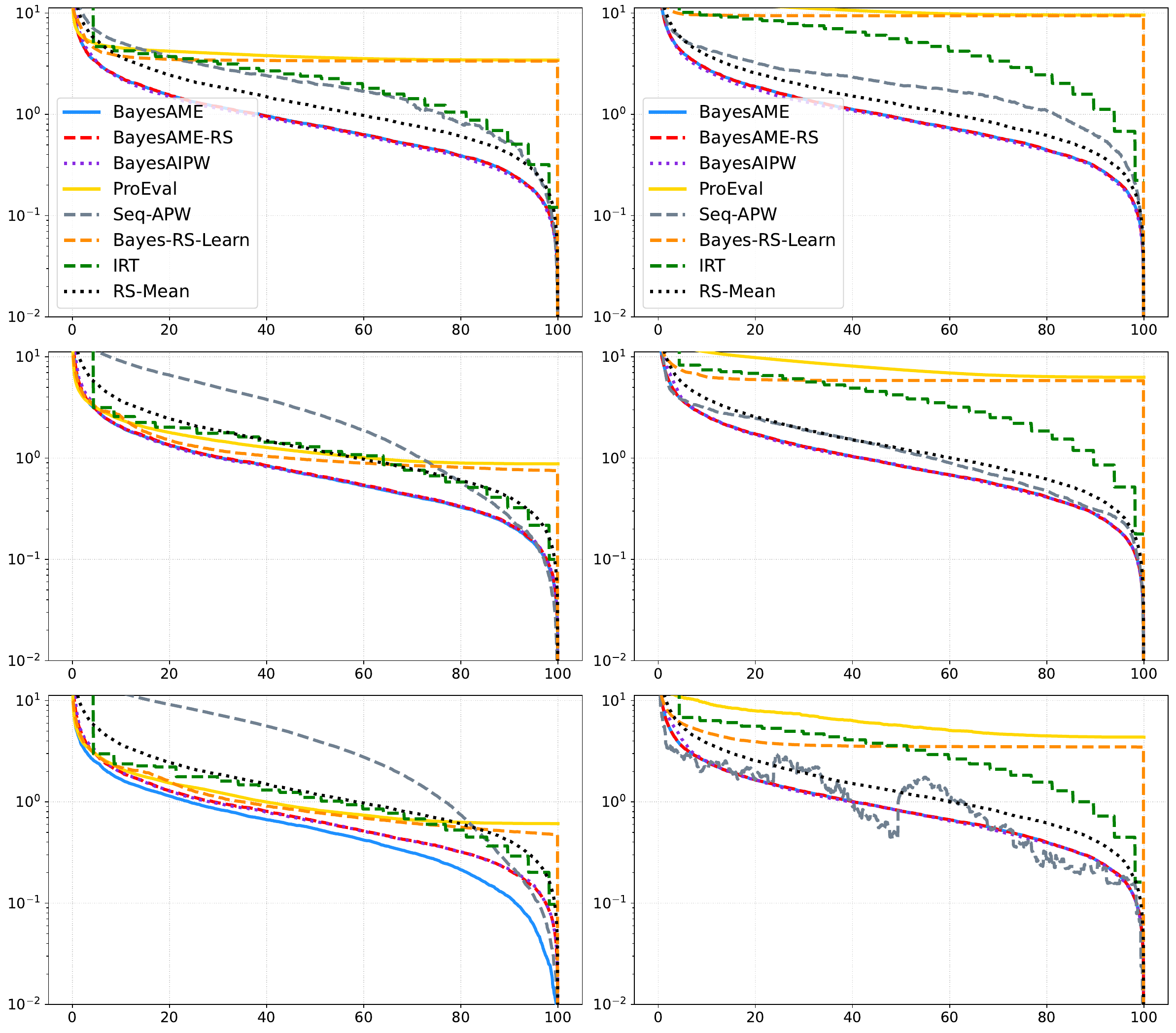} \\
\end{tabular}
\begin{tabular}{c @{\hspace{3pt}} c @{} c}
% --- X-Axis Column Headers (Top) ---
& 
% Headers for the FIRST 2x3 PDF
\makebox[0.22\textwidth][c]{\scriptsize{Interpolation}} \makebox[0.22\textwidth][c]{\scriptsize{Extrapolation}} &
% Headers for the SECOND 2x3 PDF 
\makebox[0.22\textwidth][c]{\scriptsize{Interpolation}} \makebox[0.22\textwidth][c]{\scriptsize{Extrapolation}} \\

\begin{tabular}{@{}c@{}}
    \rotatebox{90}{\scriptsize{10\%}} \\[1.5cm] %
    \rotatebox{90}{\scriptsize{50\%}} \\[1.5cm] %
    \rotatebox{90}{\scriptsize{90\%}}
\end{tabular} &
% First 2x3 PDF
\includegraphics[width=0.48\textwidth, valign=c]
{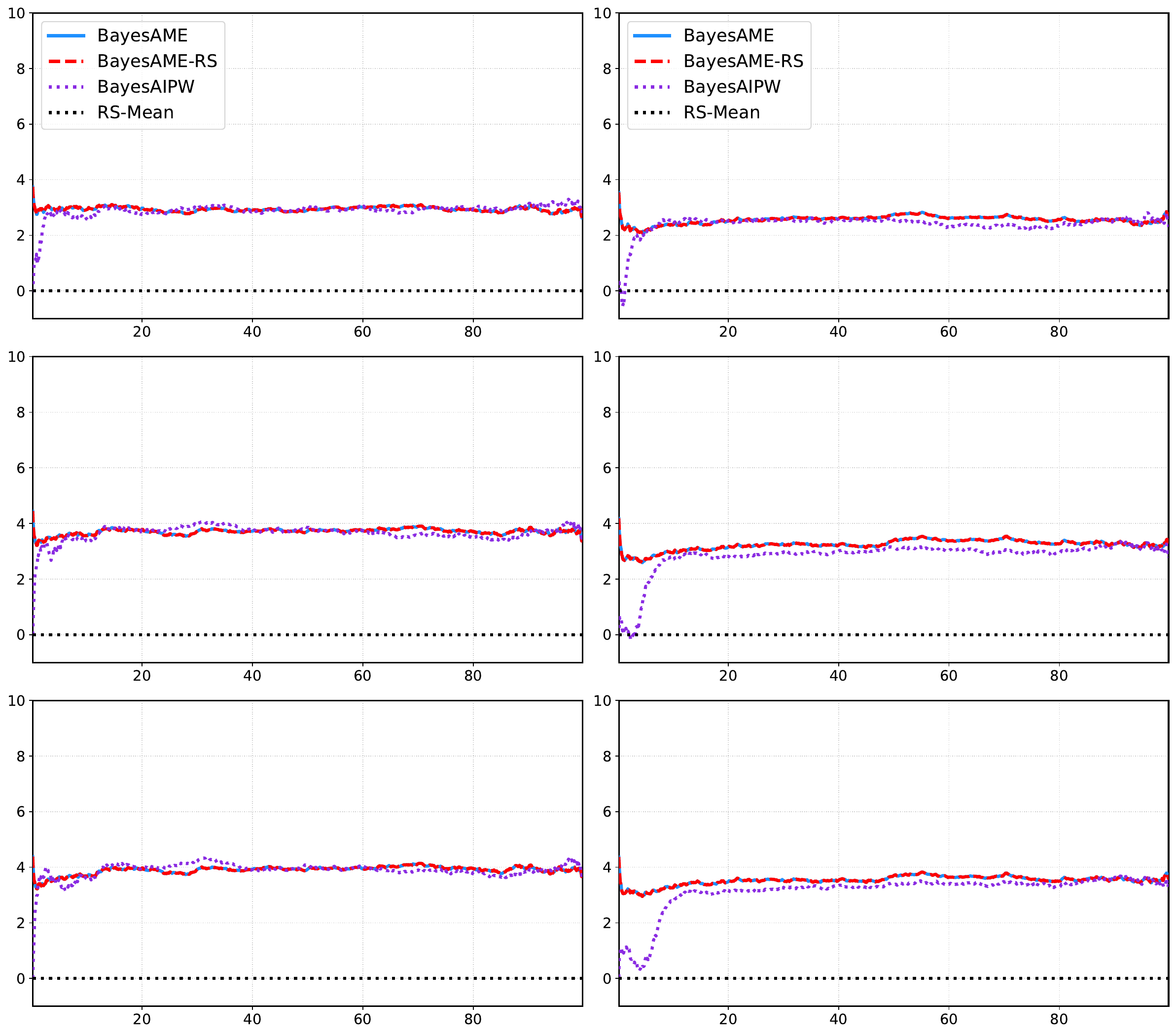} &
% Second 2x3 PDF
\includegraphics[width=0.48\textwidth, valign=c]{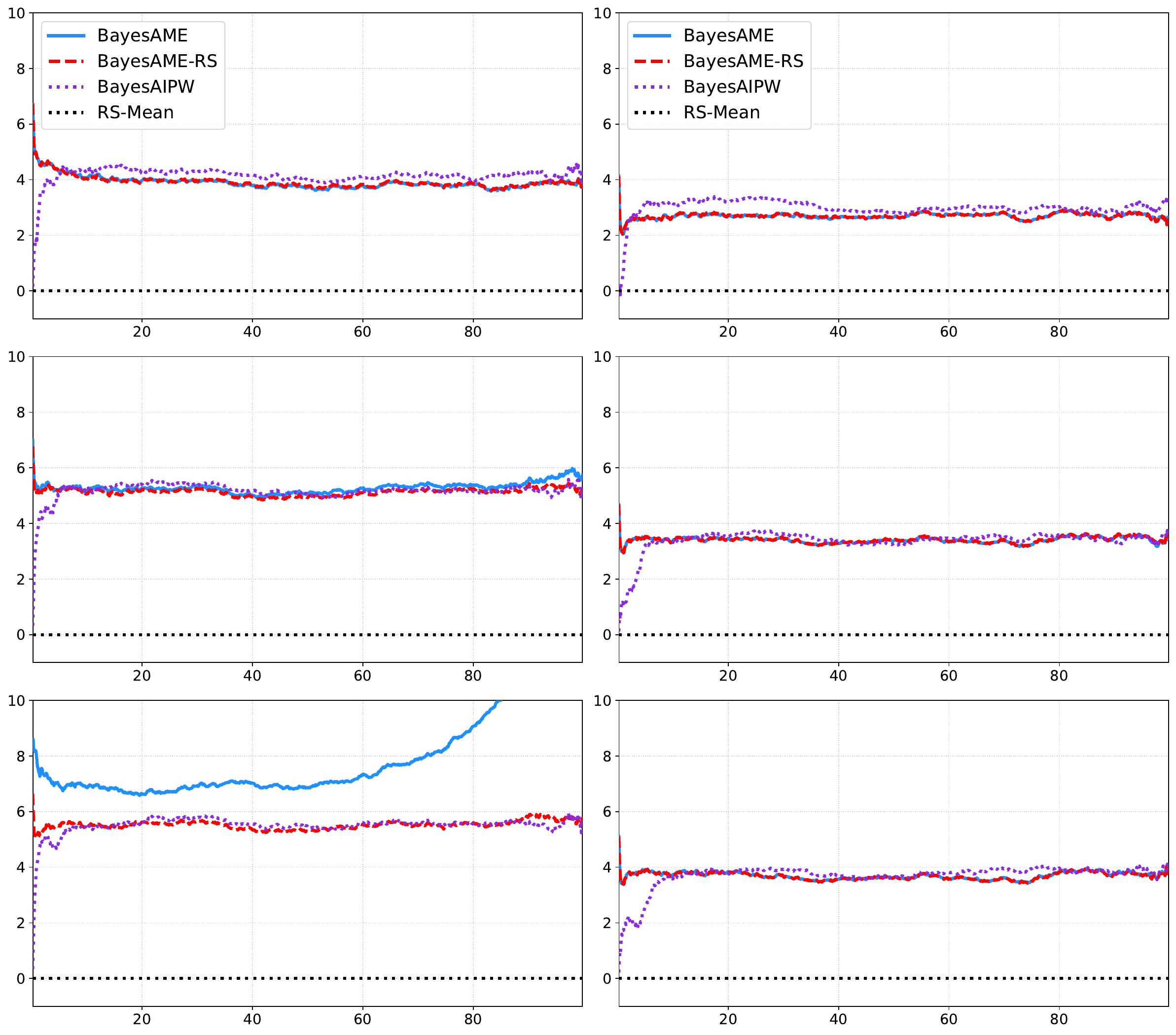} \\
\end{tabular}

\begin{tabular}{c @{\hspace{3pt}} c @{} c}
% --- X-Axis Column Headers (Top) ---
& 
% Headers for the FIRST 2x3 PDF
\makebox[0.22\textwidth][c]{\scriptsize{Interpolation}} \makebox[0.22\textwidth][c]{\scriptsize{Extrapolation}} &
% Headers for the SECOND 2x3 PDF 
\makebox[0.22\textwidth][c]{\scriptsize{Interpolation}} \makebox[0.22\textwidth][c]{\scriptsize{Extrapolation}} \\

\begin{tabular}{@{}c@{}}
    \rotatebox{90}{\scriptsize{10\%}} \\[1.5cm] %
    \rotatebox{90}{\scriptsize{50\%}} \\[1.5cm] %
    \rotatebox{90}{\scriptsize{90\%}}
\end{tabular} &
% First 2x3 PDF
\includegraphics[width=0.48\textwidth, valign=c]
{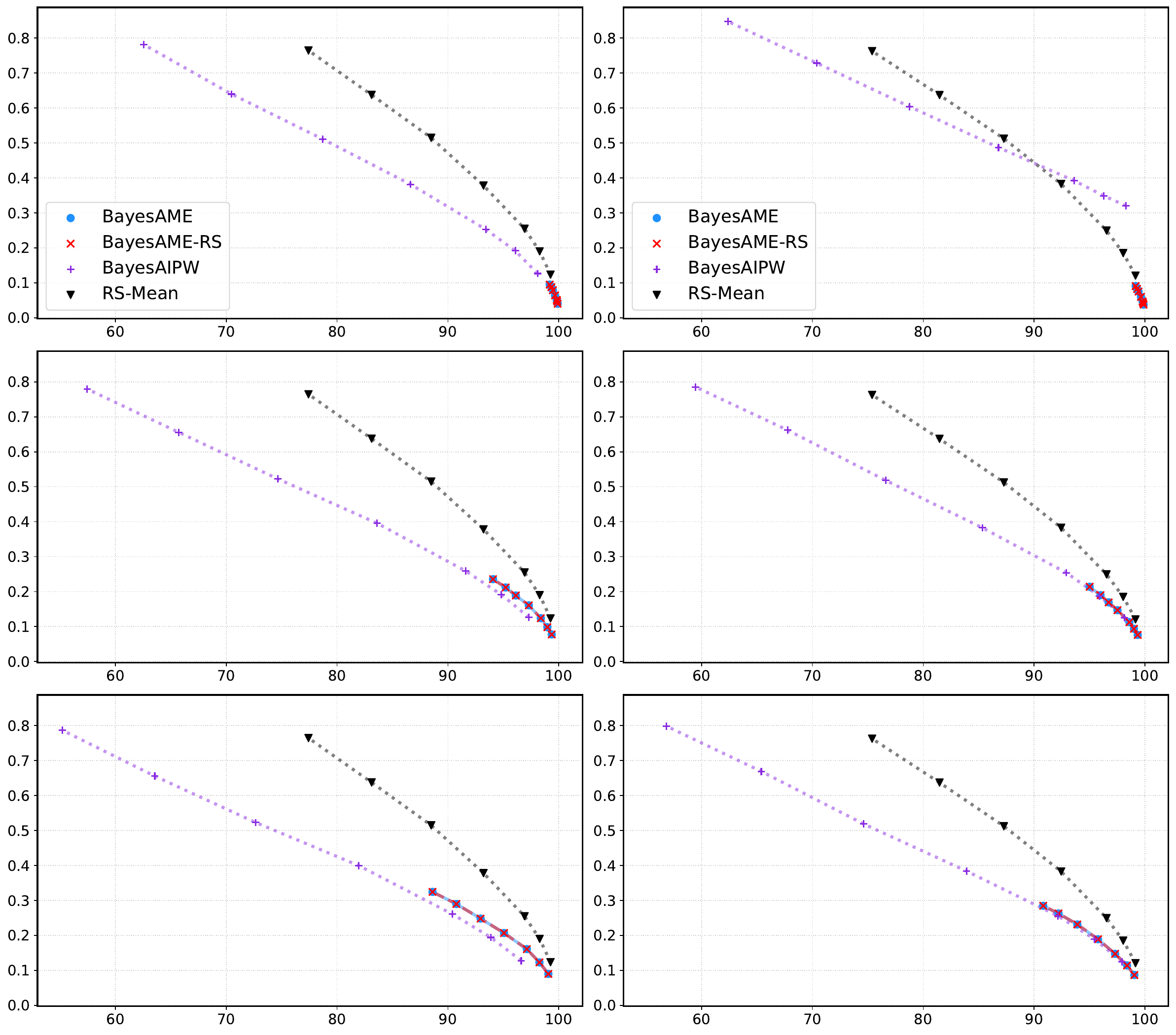} &
% Second 2x3 PDF
\includegraphics[width=0.48\textwidth, valign=c]{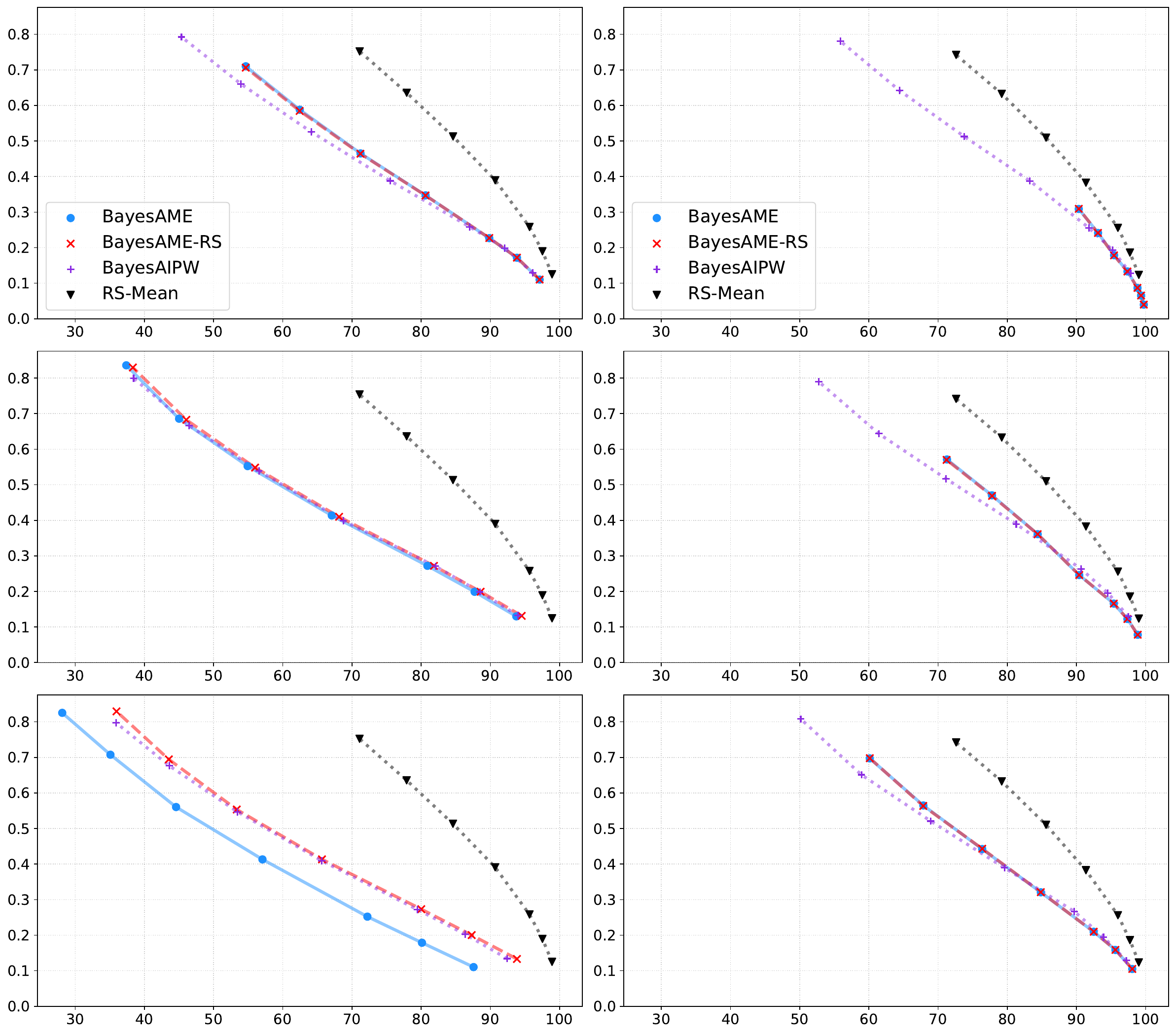} \\
\end{tabular}
\caption{\textbf{Single-target Setting.} \textbf{ARC-Challenge} with binary scores (left two columns) and continuous scores (right two columns). \RMSELog (top three rows), \RMSEGain (middle three rows), and \CAT (bottom three rows) across varying proportions of reference models.}
\label{fig:ARC-CHALLENGE-SAMPLING-SCORING-RMSELog-GAIN-CAT}
\end{figure}

\begin{figure}[!h]
\centering
\renewcommand{\arraystretch}{1.2} 

\begin{tabular}{c @{\hspace{3pt}} c @{} c}
% --- X-Axis Column Headers (Top) ---
& 
% Headers for the FIRST 2x3 PDF
\makebox[0.22\textwidth][c]{\scriptsize{Interpolation}} \makebox[0.22\textwidth][c]{\scriptsize{Extrapolation}} &
% Headers for the SECOND 2x3 PDF 
\makebox[0.22\textwidth][c]{\scriptsize{Interpolation}} \makebox[0.22\textwidth][c]{\scriptsize{Extrapolation}} \\

\begin{tabular}{@{}c@{}}
    \rotatebox{90}{\scriptsize{10\%}} \\[1.5cm] %
    \rotatebox{90}{\scriptsize{50\%}} \\[1.5cm] %
    \rotatebox{90}{\scriptsize{90\%}}
\end{tabular} &
% First 2x3 PDF
\includegraphics[width=0.48\textwidth, valign=c]
{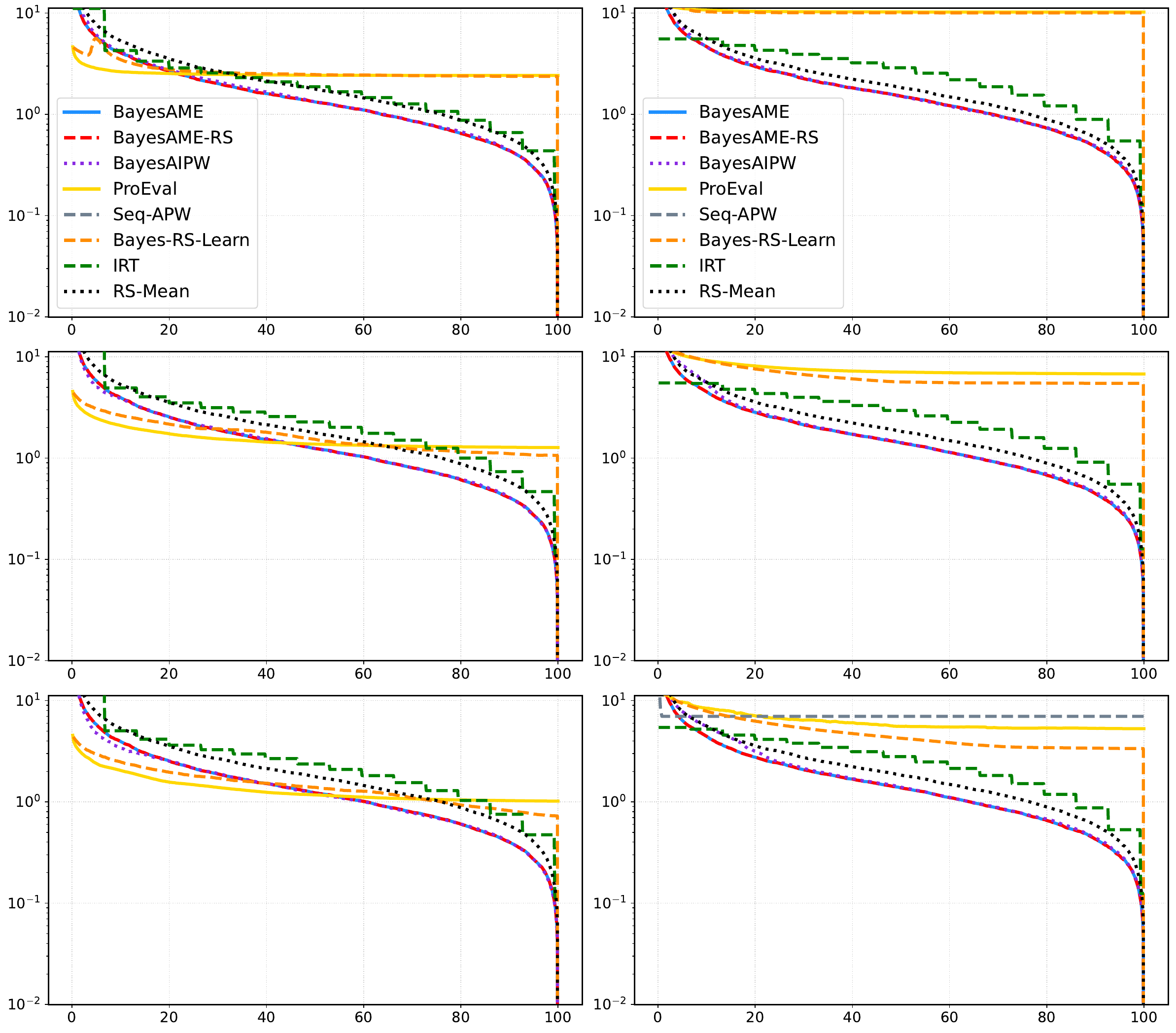} &
% Second 2x3 PDF
\includegraphics[width=0.48\textwidth, valign=c]
{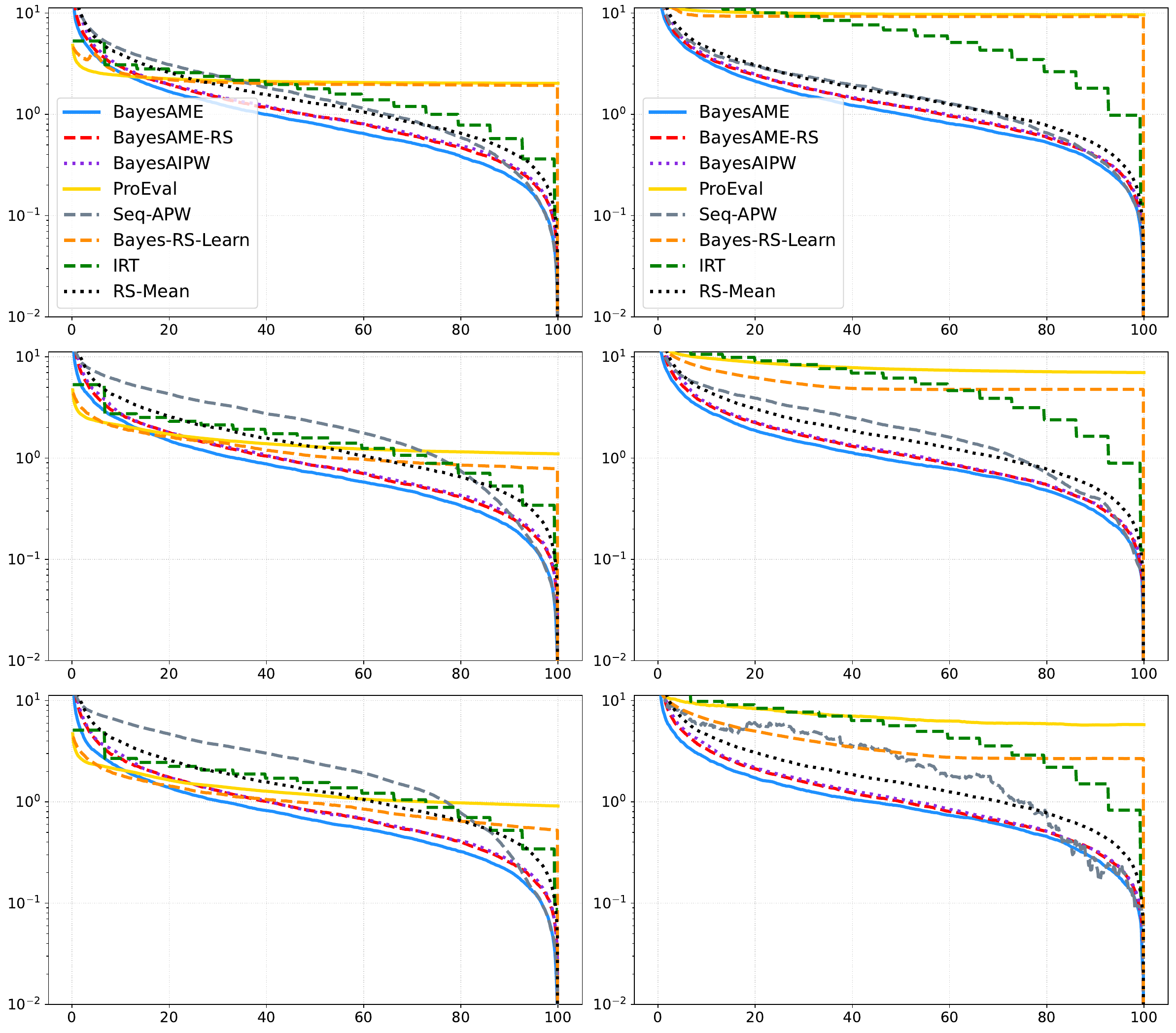} \\
\end{tabular}
\begin{tabular}{c @{\hspace{3pt}} c @{} c}
% --- X-Axis Column Headers (Top) ---
& 
% Headers for the FIRST 2x3 PDF
\makebox[0.22\textwidth][c]{\scriptsize{Interpolation}} \makebox[0.22\textwidth][c]{\scriptsize{Extrapolation}} &
% Headers for the SECOND 2x3 PDF 
\makebox[0.22\textwidth][c]{\scriptsize{Interpolation}} \makebox[0.22\textwidth][c]{\scriptsize{Extrapolation}} \\

\begin{tabular}{@{}c@{}}
    \rotatebox{90}{\scriptsize{10\%}} \\[1.5cm] %
    \rotatebox{90}{\scriptsize{50\%}} \\[1.5cm] %
    \rotatebox{90}{\scriptsize{90\%}}
\end{tabular} &
% First 2x3 PDF
\includegraphics[width=0.48\textwidth, valign=c]
{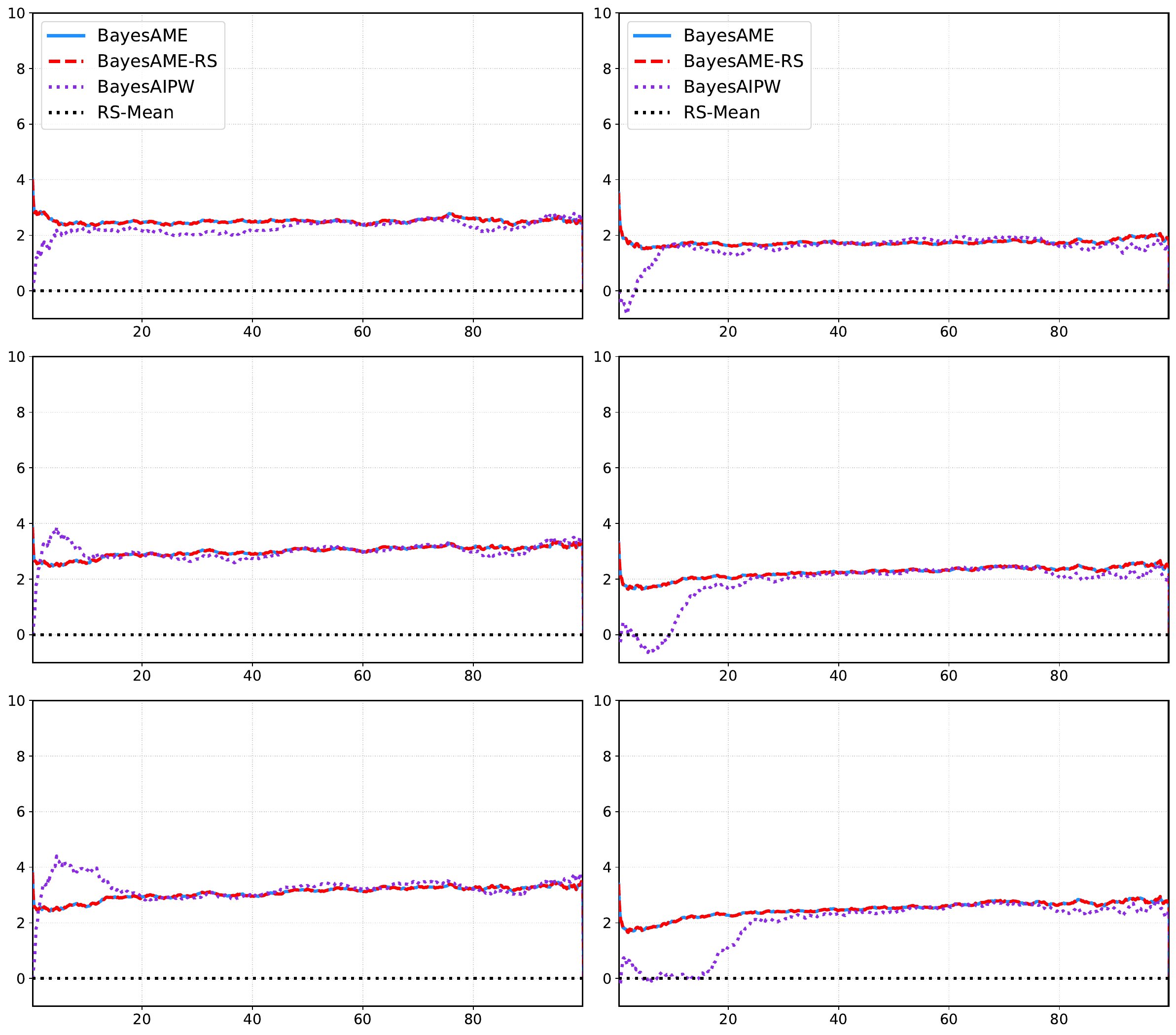} &
% Second 2x3 PDF
\includegraphics[width=0.48\textwidth, valign=c]{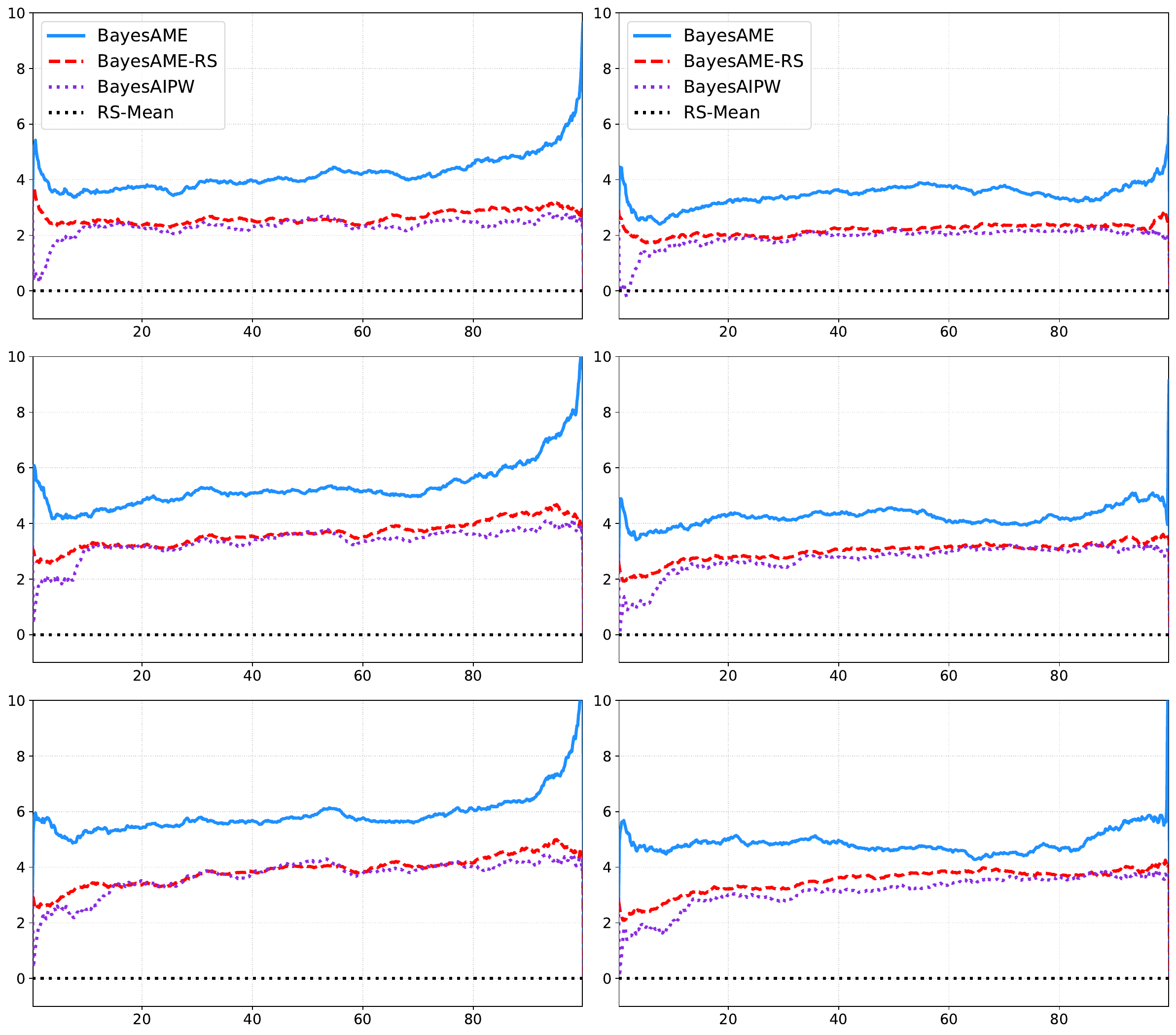} \\
\end{tabular}

\begin{tabular}{c @{\hspace{3pt}} c @{} c}
% --- X-Axis Column Headers (Top) ---
& 
% Headers for the FIRST 2x3 PDF
\makebox[0.22\textwidth][c]{\scriptsize{Interpolation}} \makebox[0.22\textwidth][c]{\scriptsize{Extrapolation}} &
% Headers for the SECOND 2x3 PDF 
\makebox[0.22\textwidth][c]{\scriptsize{Interpolation}} \makebox[0.22\textwidth][c]{\scriptsize{Extrapolation}} \\

\begin{tabular}{@{}c@{}}
    \rotatebox{90}{\scriptsize{10\%}} \\[1.5cm] %
    \rotatebox{90}{\scriptsize{50\%}} \\[1.5cm] %
    \rotatebox{90}{\scriptsize{90\%}}
\end{tabular} &
% First 2x3 PDF
\includegraphics[width=0.48\textwidth, valign=c]
{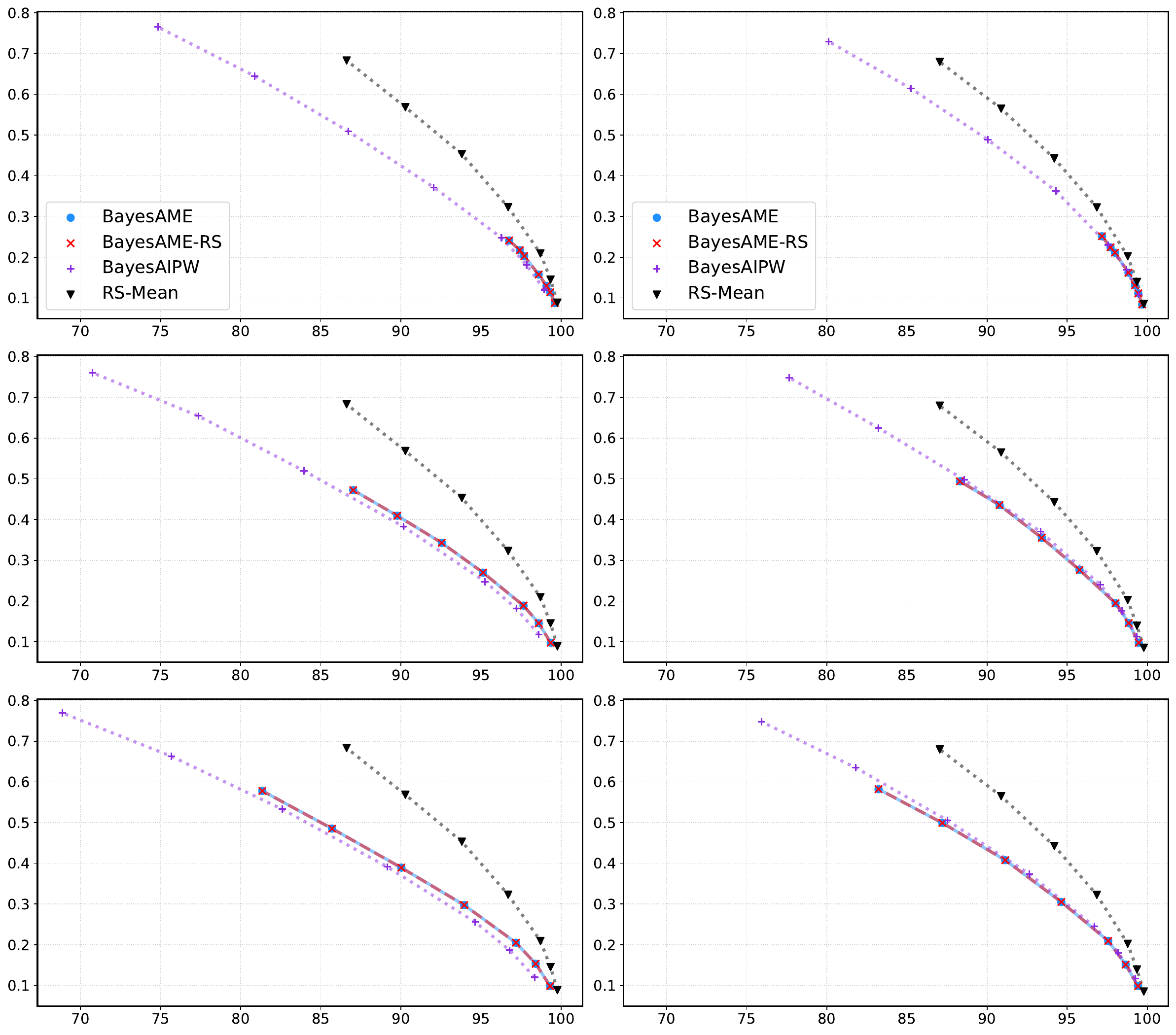} &
% Second 2x3 PDF
\includegraphics[width=0.48\textwidth, valign=c]{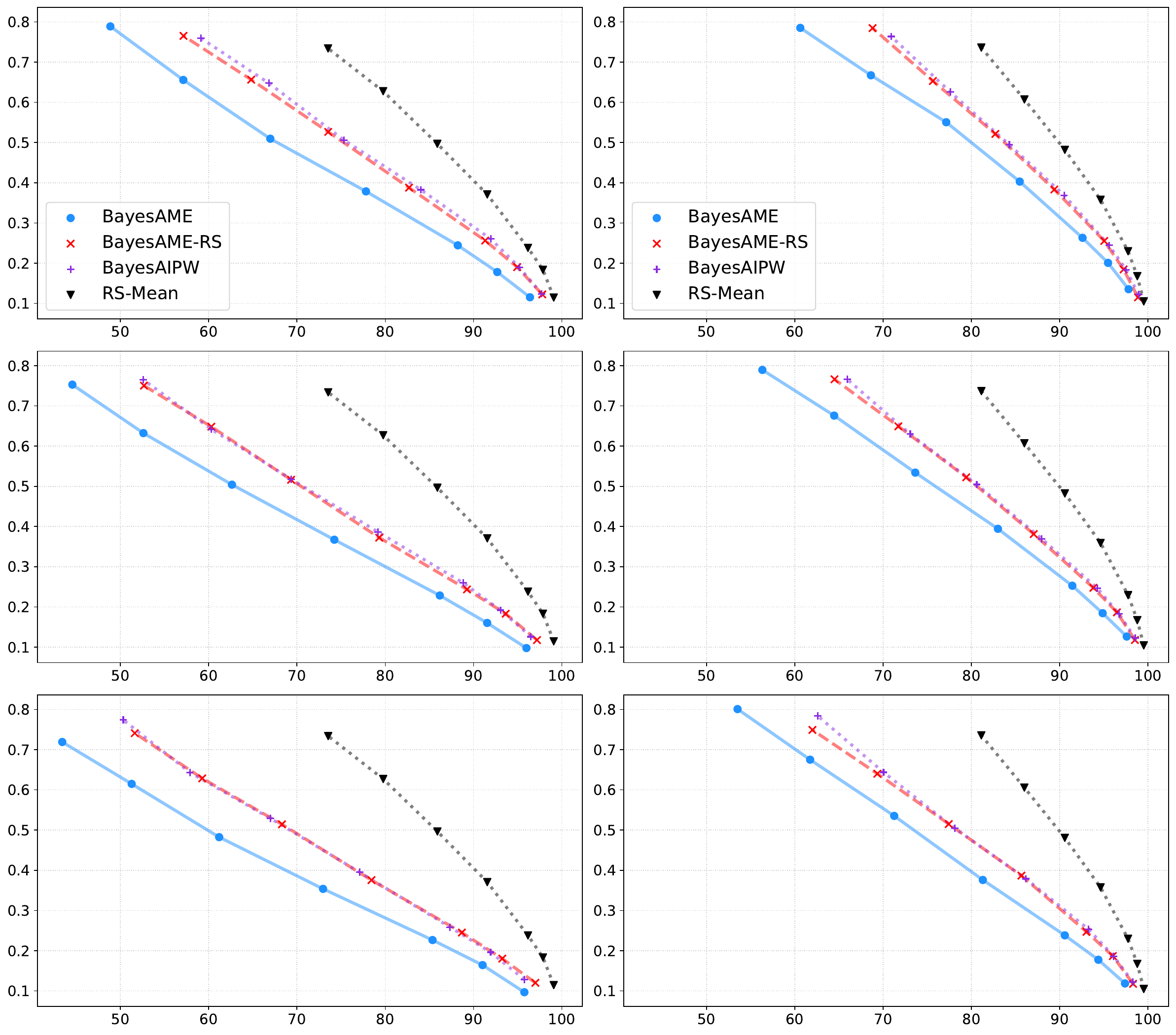} \\
\end{tabular}
\caption{\textbf{Single-target Setting.} \textbf{MuSR} with binary scores (left two columns) and continuous scores (right two columns). \RMSELog (top three rows), \RMSEGain (middle three rows), and \CAT (bottom three rows) across varying proportions of reference models.}
\label{fig:MuSR-SAMPLING-SCORING-RMSELog-GAIN-CAT}
\end{figure}

\begin{figure}[!h]
\centering
\renewcommand{\arraystretch}{1.2} 

\begin{tabular}{c @{\hspace{3pt}} c @{} c}
% --- X-Axis Column Headers (Top) ---
& 
% Headers for the FIRST 2x3 PDF
\makebox[0.22\textwidth][c]{\scriptsize{Interpolation}} \makebox[0.22\textwidth][c]{\scriptsize{Extrapolation}} &
% Headers for the SECOND 2x3 PDF 
\makebox[0.22\textwidth][c]{\scriptsize{Interpolation}} \makebox[0.22\textwidth][c]{\scriptsize{Extrapolation}} \\

\begin{tabular}{@{}c@{}}
    \rotatebox{90}{\scriptsize{10\%}} \\[1.5cm] %
    \rotatebox{90}{\scriptsize{50\%}} \\[1.5cm] %
    \rotatebox{90}{\scriptsize{90\%}}
\end{tabular} &
% First 2x3 PDF
\includegraphics[width=0.48\textwidth, valign=c]
{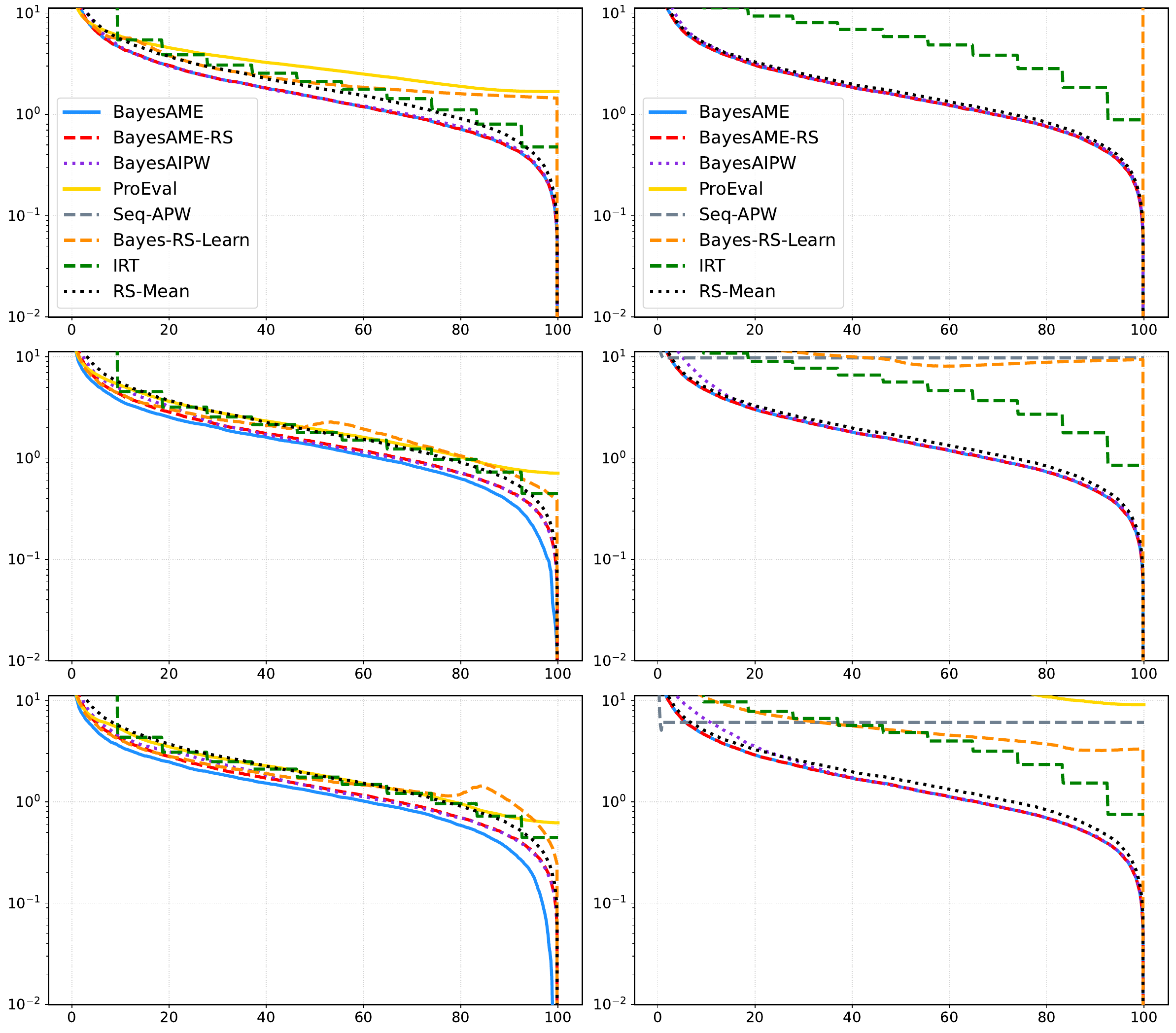} &
% Second 2x3 PDF
\includegraphics[width=0.48\textwidth, valign=c]
{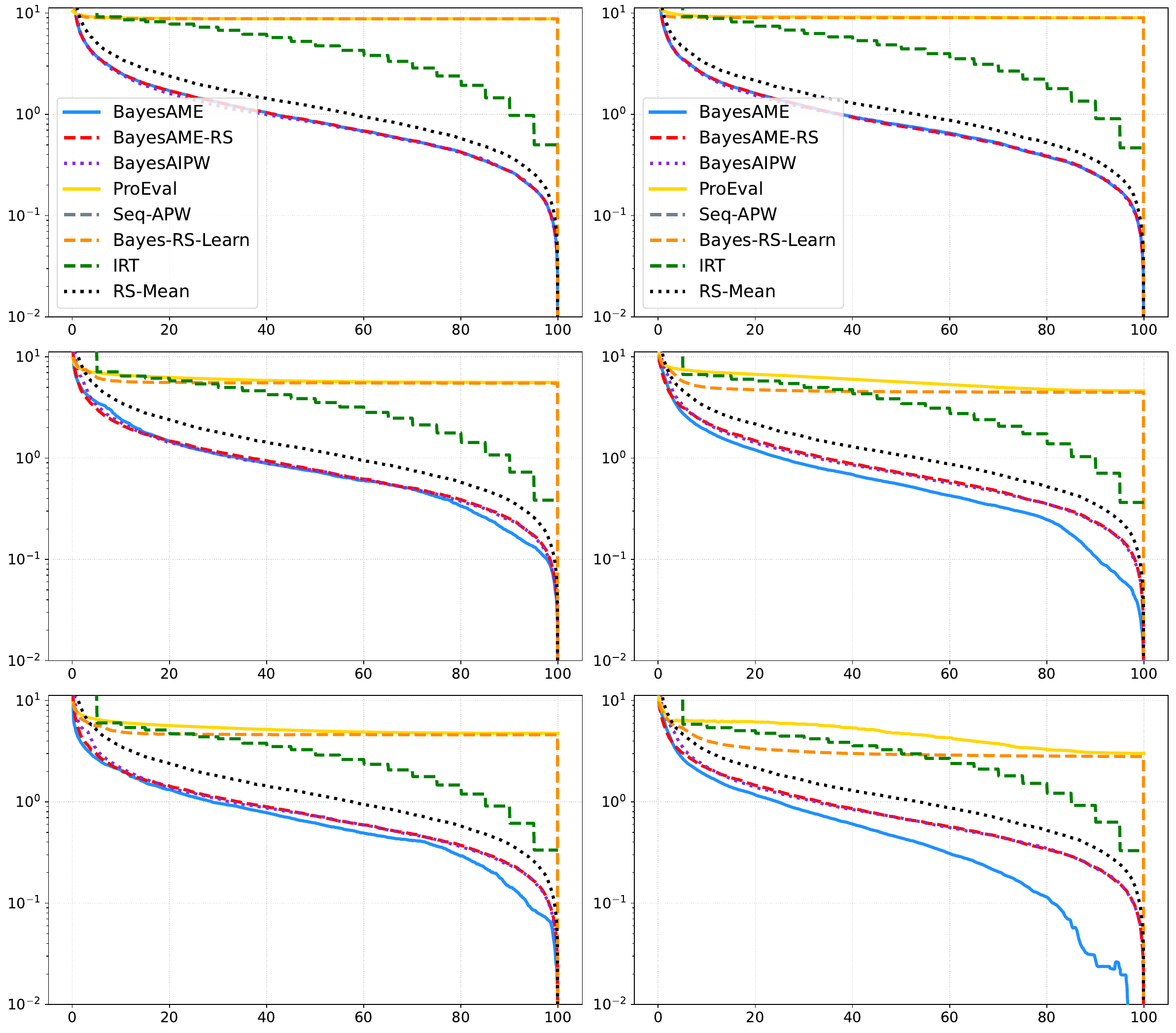} \\
\end{tabular}
\begin{tabular}{c @{\hspace{3pt}} c @{} c}
% --- X-Axis Column Headers (Top) ---
& 
% Headers for the FIRST 2x3 PDF
\makebox[0.22\textwidth][c]{\scriptsize{Interpolation}} \makebox[0.22\textwidth][c]{\scriptsize{Extrapolation}} &
% Headers for the SECOND 2x3 PDF 
\makebox[0.22\textwidth][c]{\scriptsize{Interpolation}} \makebox[0.22\textwidth][c]{\scriptsize{Extrapolation}} \\

\begin{tabular}{@{}c@{}}
    \rotatebox{90}{\scriptsize{10\%}} \\[1.5cm] %
    \rotatebox{90}{\scriptsize{50\%}} \\[1.5cm] %
    \rotatebox{90}{\scriptsize{90\%}}
\end{tabular} &
% First 2x3 PDF
\includegraphics[width=0.48\textwidth, valign=c]
{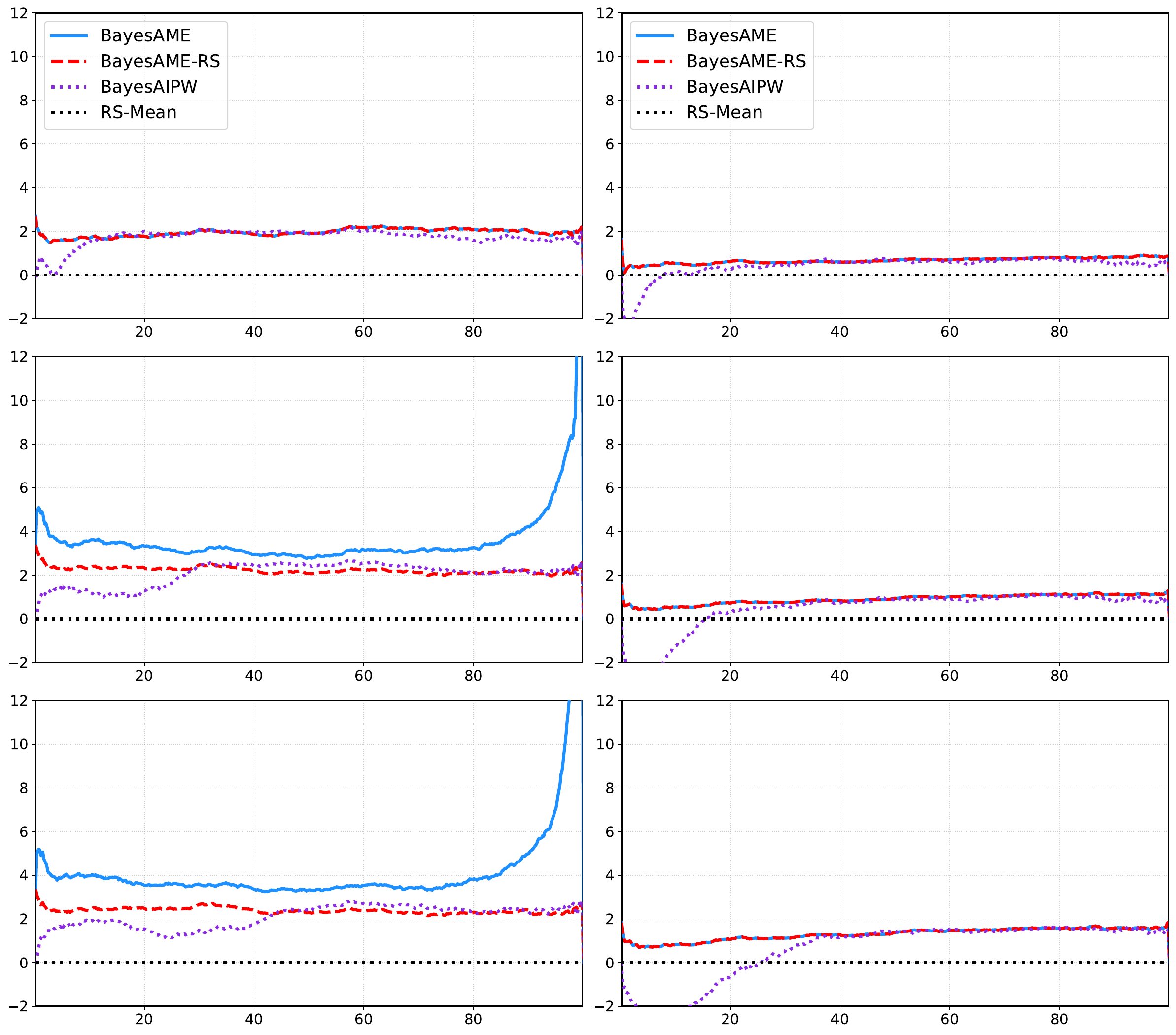} &
% Second 2x3 PDF
\includegraphics[width=0.48\textwidth, valign=c]{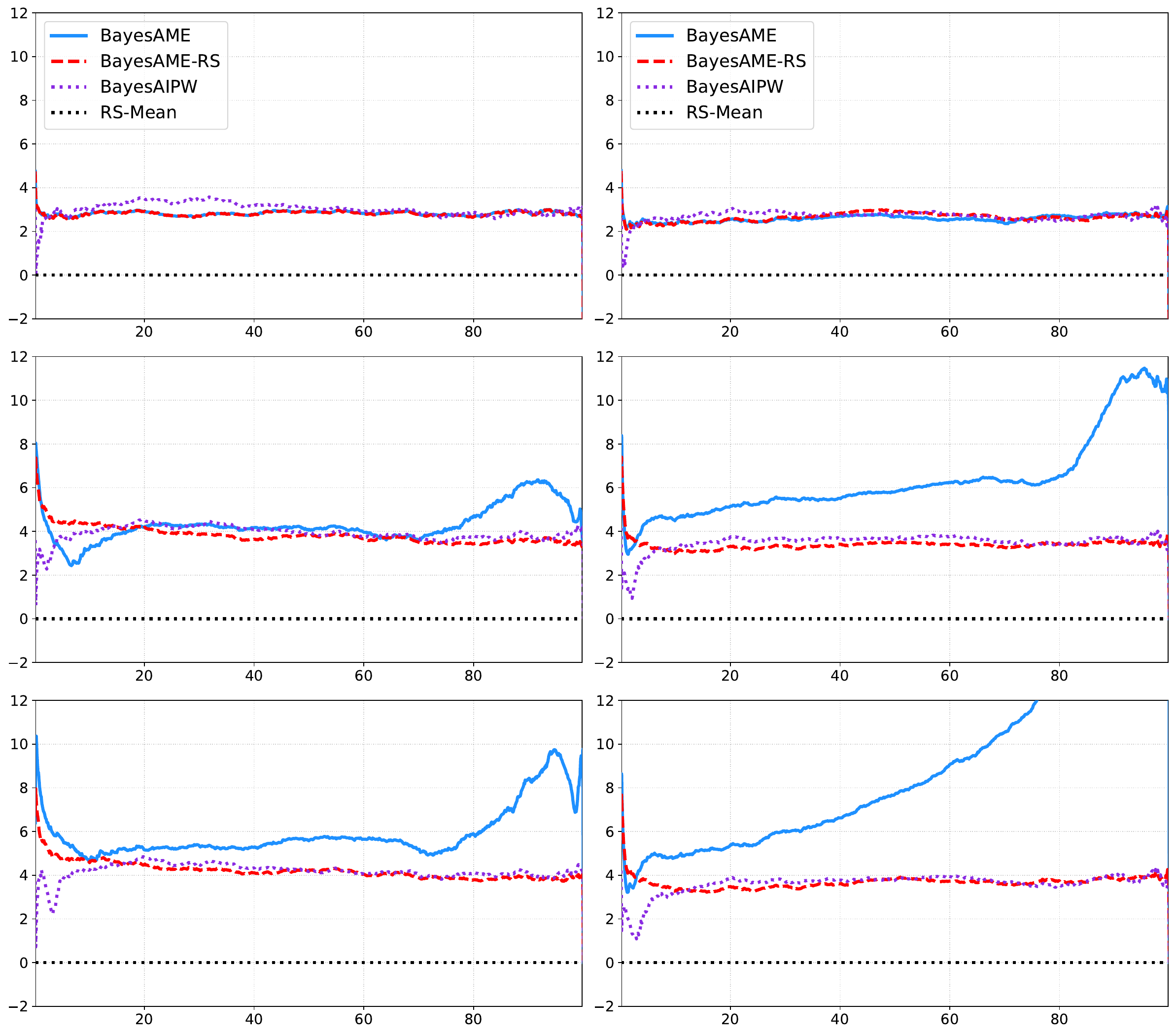} \\
\end{tabular}

\begin{tabular}{c @{\hspace{3pt}} c @{} c}
% --- X-Axis Column Headers (Top) ---
& 
% Headers for the FIRST 2x3 PDF
\makebox[0.22\textwidth][c]{\scriptsize{Interpolation}} \makebox[0.22\textwidth][c]{\scriptsize{Extrapolation}} &
% Headers for the SECOND 2x3 PDF 
\makebox[0.22\textwidth][c]{\scriptsize{Interpolation}} \makebox[0.22\textwidth][c]{\scriptsize{Extrapolation}} \\

\begin{tabular}{@{}c@{}}
    \rotatebox{90}{\scriptsize{10\%}} \\[1.5cm] %
    \rotatebox{90}{\scriptsize{50\%}} \\[1.5cm] %
    \rotatebox{90}{\scriptsize{90\%}}
\end{tabular} &
% First 2x3 PDF
\includegraphics[width=0.48\textwidth, valign=c]
{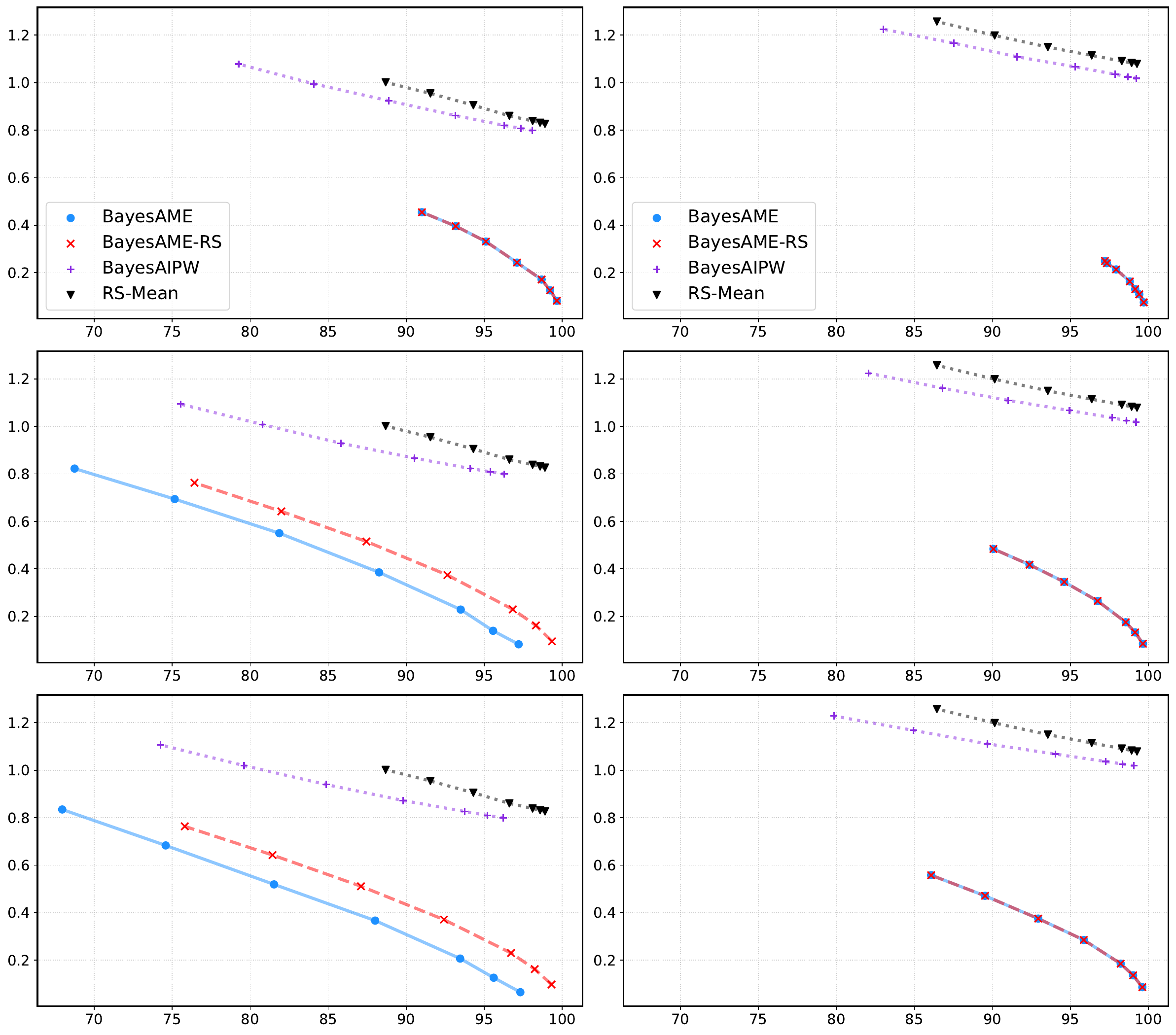} &
% Second 2x3 PDF
\includegraphics[width=0.48\textwidth, valign=c]{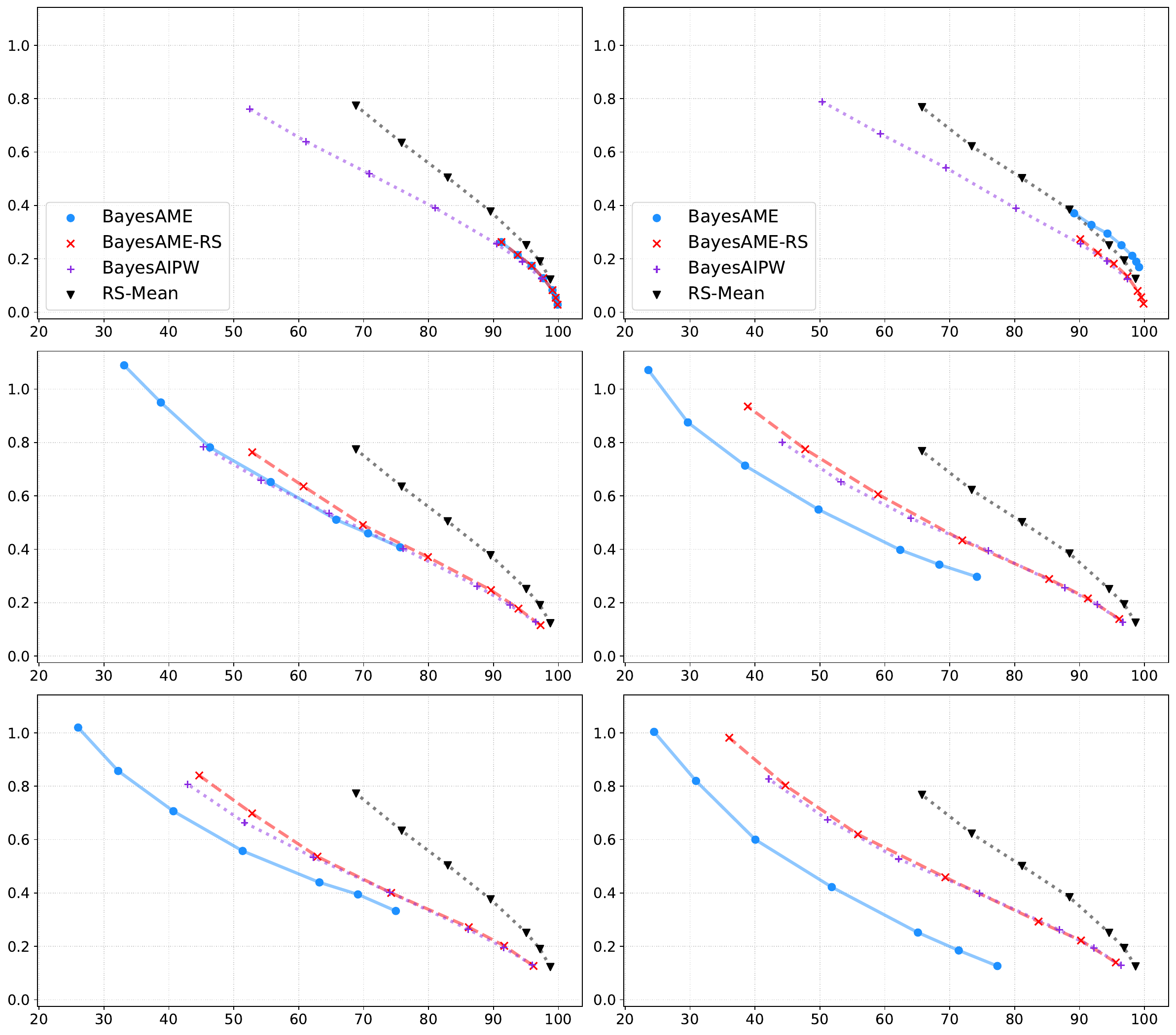} \\
\end{tabular}
\caption{\textbf{Single-target Setting.} \textbf{IFEval} with binary scores (left two columns) and \textbf{Natural QA Openbook} with continuous scores (right two columns). \RMSELog (top three rows), \RMSEGain (middle three rows), and \CAT (bottom three rows) across varying proportions of reference models.}
\label{fig:IFEVAL-NaturalQAOpenbook-RMSELog-GAIN-CAT}
\end{figure}

\begin{figure}[!h]
\centering
\renewcommand{\arraystretch}{1.2} 

\begin{tabular}{c @{\hspace{3pt}} c @{} c}
% --- X-Axis Column Headers (Top) ---
& 
% Headers for the FIRST 2x3 PDF
\makebox[0.22\textwidth][c]{\scriptsize{Interpolation}} \makebox[0.22\textwidth][c]{\scriptsize{Extrapolation}} &
% Headers for the SECOND 2x3 PDF 
\makebox[0.22\textwidth][c]{\scriptsize{Interpolation}} \makebox[0.22\textwidth][c]{\scriptsize{Extrapolation}} \\

\begin{tabular}{@{}c@{}}
    \rotatebox{90}{\scriptsize{10\%}} \\[1.5cm] %
    \rotatebox{90}{\scriptsize{50\%}} \\[1.5cm] %
    \rotatebox{90}{\scriptsize{90\%}}
\end{tabular} &
% First 2x3 PDF
\includegraphics[width=0.48\textwidth, valign=c]{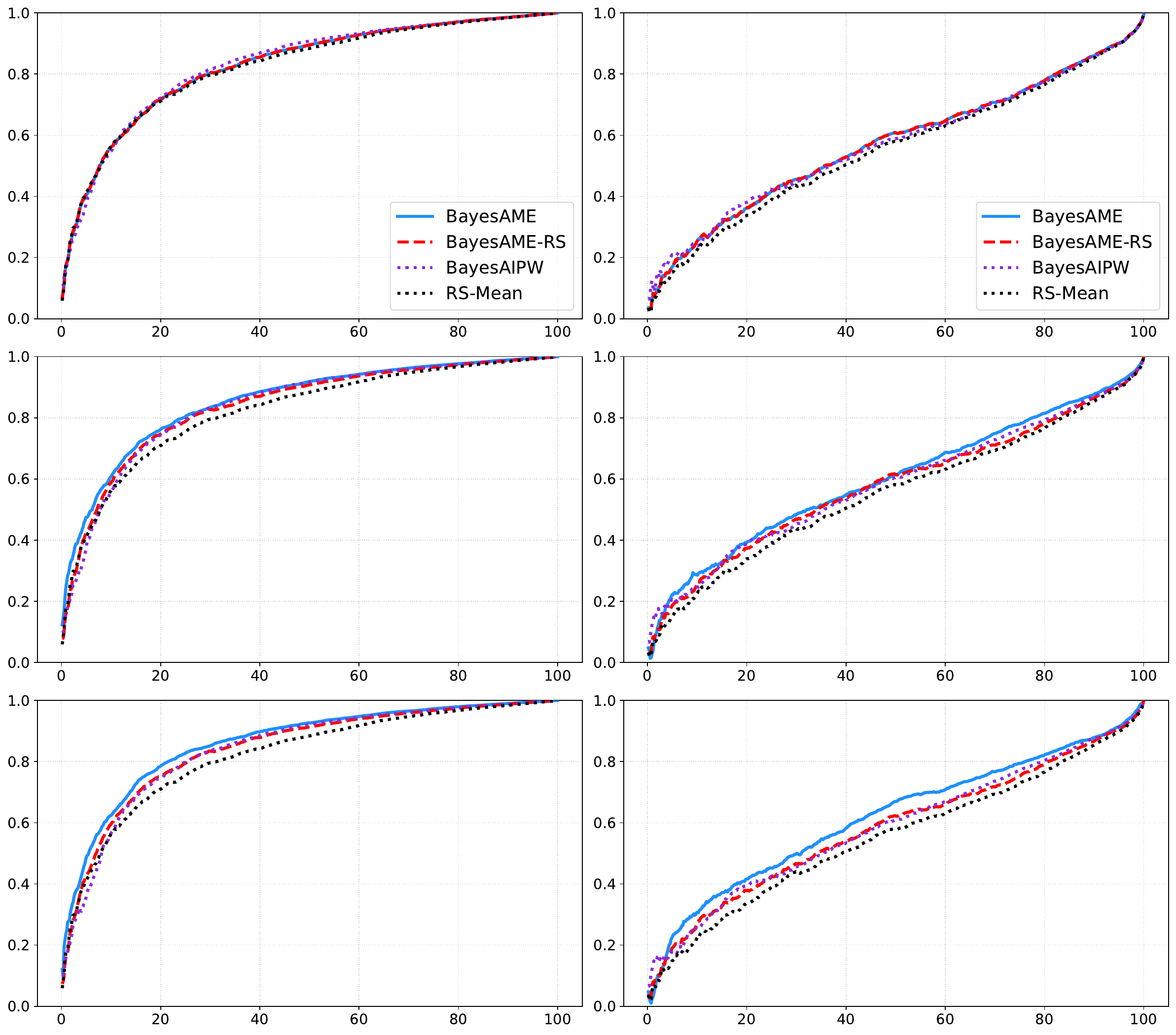} &
% Second 2x3 PDF
\includegraphics[width=0.48\textwidth, valign=c]{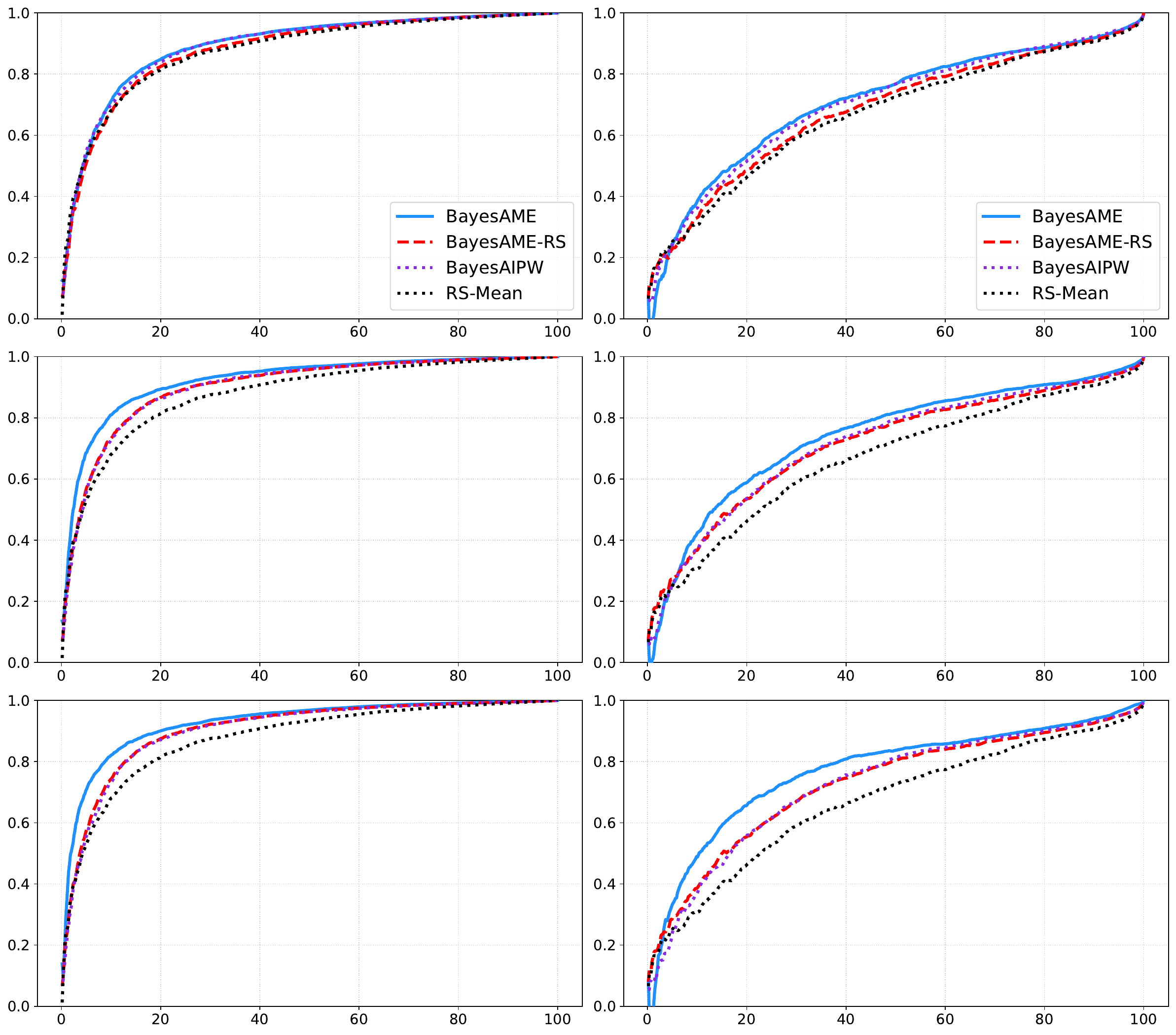} \\
\end{tabular}

\begin{tabular}{c @{\hspace{3pt}} c @{} c}
% --- X-Axis Column Headers (Top) ---
& 
% Headers for the FIRST 2x3 PDF
\makebox[0.22\textwidth][c]{\scriptsize{Interpolation}} \makebox[0.22\textwidth][c]{\scriptsize{Extrapolation}} &
% Headers for the SECOND 2x3 PDF 
\makebox[0.22\textwidth][c]{\scriptsize{Interpolation}} \makebox[0.22\textwidth][c]{\scriptsize{Extrapolation}} \\

\begin{tabular}{@{}c@{}}
    \rotatebox{90}{\scriptsize{10\%}} \\[1.5cm] %
    \rotatebox{90}{\scriptsize{50\%}} \\[1.5cm] %
    \rotatebox{90}{\scriptsize{90\%}}
\end{tabular} &
% First 2x3 PDF
\includegraphics[width=0.48\textwidth, valign=c]{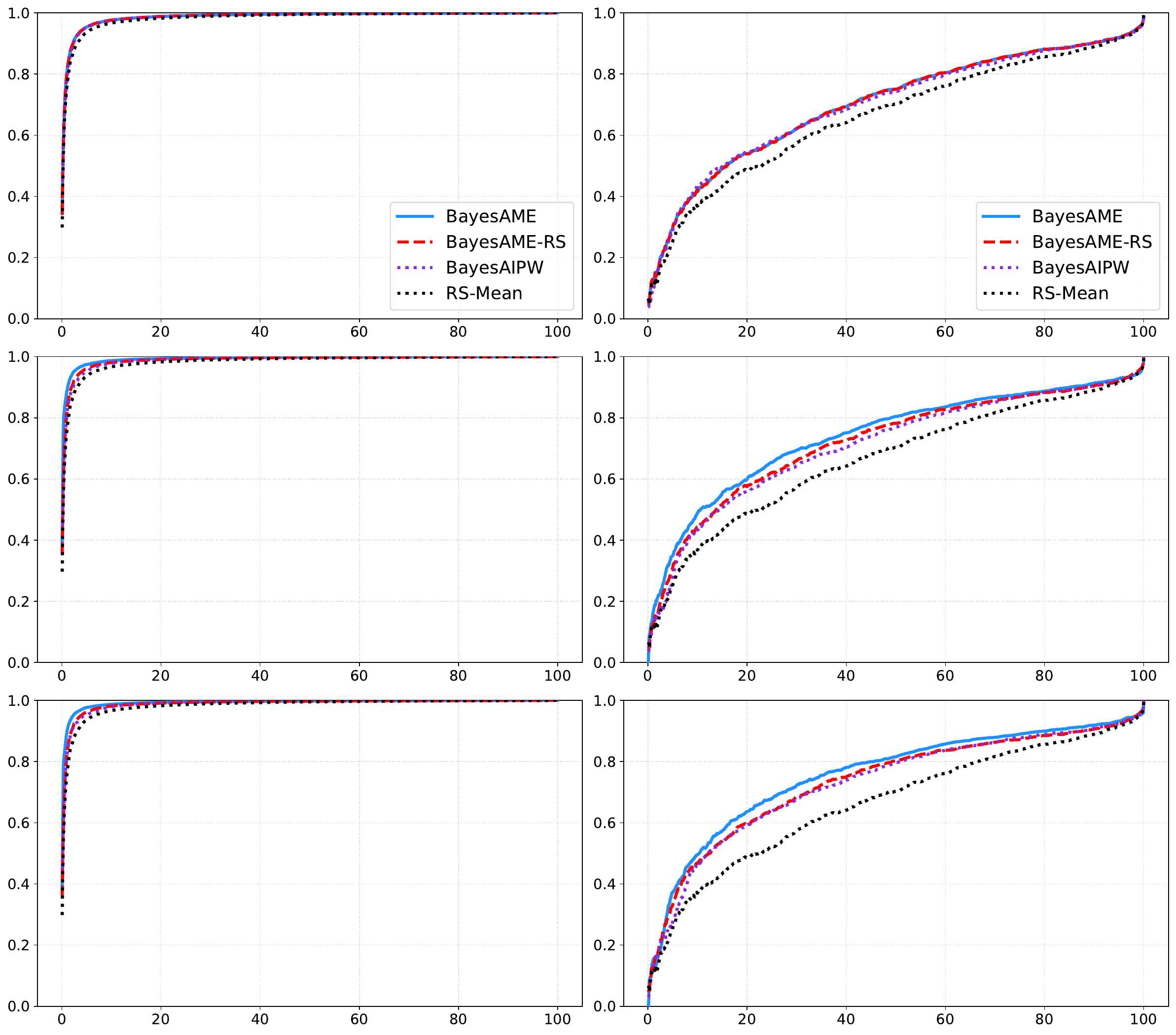} &
% Second 2x3 PDF
\includegraphics[width=0.48\textwidth, valign=c]{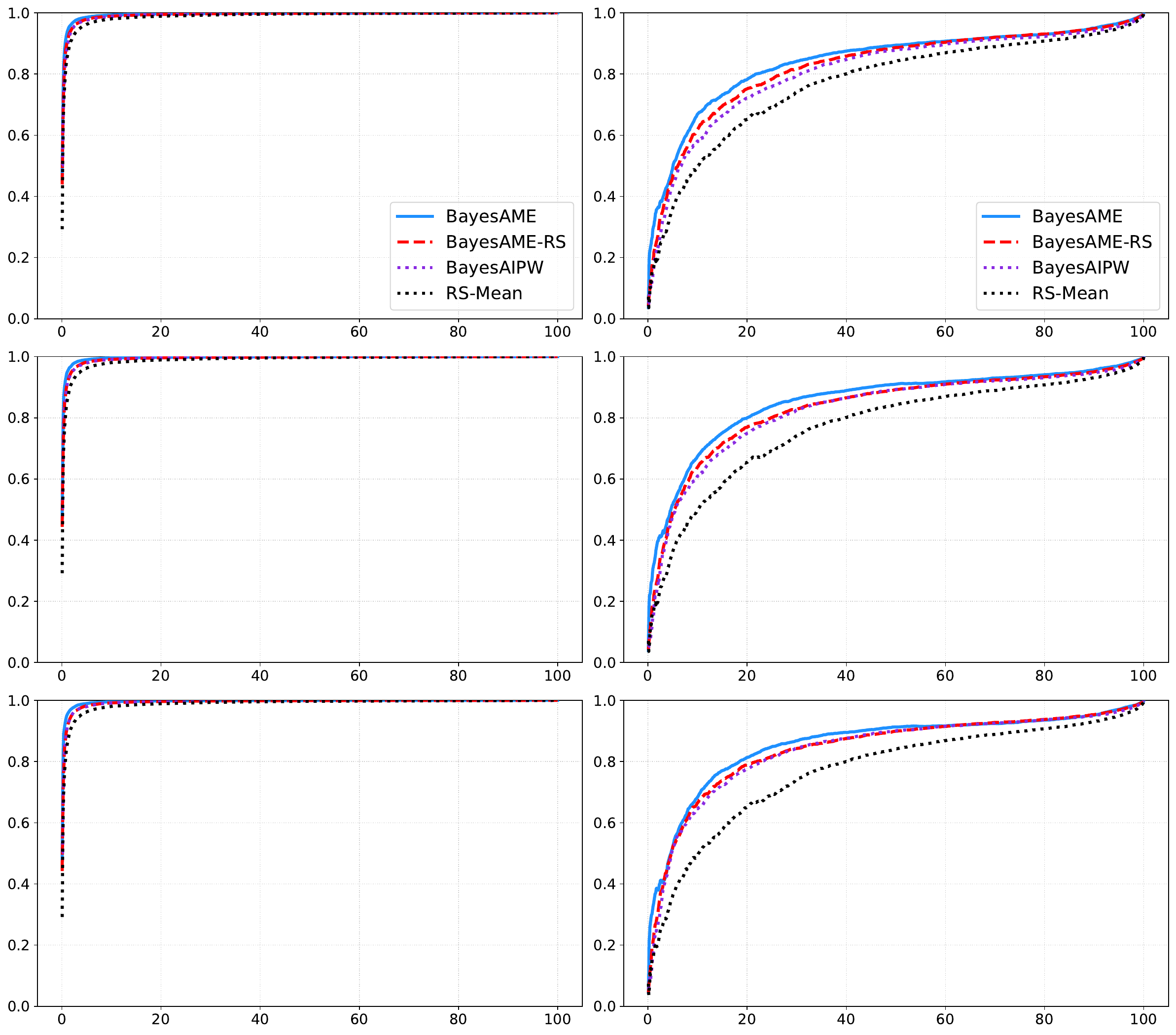} \\
\end{tabular}
\caption{\textbf{Single-target Setting.} \textbf{GPQA} (top three rows) and \textbf{BBH} (bottom three rows) with binary scores (left two columns) and continuous scores (right two columns). Spearman's rank correlation across varying proportions of reference models.}
\label{fig:GPQA-BBH-SAMPLING-SCORING-SPEARMAN}
\end{figure}

\end{document}